\documentclass[11pt]{article}

\usepackage[final]{acl}

\usepackage{times}
\usepackage{latexsym}

\usepackage[T1]{fontenc}

\usepackage[utf8]{inputenc}

\usepackage{microtype}

\usepackage{inconsolata}

\usepackage{graphicx}

\usepackage{times}
\usepackage{soul}
\usepackage{url}
\usepackage[utf8]{inputenc}
\usepackage[font=small]{caption}
\usepackage{amsmath}
\usepackage{amsthm}
\usepackage{booktabs}
\usepackage{algorithm}
\usepackage{algorithmic}
\usepackage[switch]{lineno}

\usepackage{graphicx}
\usepackage{subcaption}
\usepackage{tcolorbox}
\usepackage{booktabs}
\usepackage{amsmath}
\usepackage{enumitem}
\usepackage{array}
\usepackage{makecell}
\usepackage{pifont}
\usepackage{multirow}

\usepackage{float}

\usepackage{caption}
\usepackage{placeins}

\newcommand{\cmark}{\ding{51}} 
\newcommand{\xmark}{\ding{55}}

\title{Culturally Grounded Personas in Large Language Models:  Characterization and Alignment with Socio‑Psychological Value Frameworks}

\author{Candida M. Greco, Lucio La Cava, Andrea Tagarelli \\
         DIMES, University of Calabria, Italy \\ 
         \{candida.greco, lucio.lacava, tagarelli\}\@dimes.unical.it}

\begin{document}
\maketitle
\begin{abstract}
Despite the growing utility of Large Language Models (LLMs) for simulating human behavior, the extent to which these synthetic personas accurately reflect world and moral value systems across different cultural conditionings remains uncertain.  This paper investigates the alignment of synthetic, culturally-grounded personas with established frameworks, specifically the World Values Survey (WVS), the Inglehart–Welzel Cultural Map, and Moral Foundations Theory.   
We conceptualize and produce LLM-generated  personas based on a set of interpretable WVS-derived variables, and we  examine the generated personas through three complementary lenses:  positioning on the Inglehart–Welzel map, which unveils their interpretation  reflecting stable differences across cultural conditionings;  demographic-level consistency with the World Values Survey, where response distributions broadly track human group patterns; and  moral profiles derived from a Moral Foundations questionnaire, which we analyze through a culture-to-morality mapping to characterize how moral responses vary across different cultural configurations. Our approach of culturally-grounded persona generation and analysis  enables evaluation of cross-cultural structure and moral variation.
\end{abstract}

\section{Introduction}

Large Language Models are increasingly used to generate human personas and model preferences and behavior, but it remains unclear how well these personas reflect real cultural, demographic, and moral structures across societies.

In this work, we investigate the extent to which LLM-generated culturally-grounded persona profiles align with established  socio-psychological frameworks:  the World Values Survey (WVS), the Inglehart–Welzel (IW) Cultural Map, and Moral Foundations Theory (MFT). 
More specifically, we aim  to address the following research questions: 

\vspace{1.5mm}
\noindent$\bullet$ \textbf{RQ1}: 
How are synthetically generated culturally-grounded personas positioned along the   principal dimensions of cross-cultural variation?  
 
\noindent$\bullet$ \textbf{RQ2}:  
Can an LLM mimicking a culturally-grounded persona have a WVS cultural profile that aligns with that of human demographic groups?

\noindent$\bullet$ \textbf{RQ3}:   How are the WVS-derived cultural variables mapped to the moral foundations?  And how do   the culturally-grounded personas map onto scores of moral foundations questionnaire administered to a culturally-conditioned LLM?  

\vspace{1mm}
\noindent
\textbf{Contributions.} To answer the above RQs, our proposed methodology starts with the identification of  a targeted set of WVS-derived  cultural variables with explicit IW axis interpretation, which are then used to prompt an LLM  to generate culturally-grounded personas. 
Each of the  generated personas is  used to condition another LLM before administering two different  WVS-based benchmark resources, which respectively   provide  indicators underlying the IW dimensions and WorldValueBench cultural values associated with demographic groups. 
In addition, we define two  moral-foundation representation models for each generated culturally-grounded  persona, to assess responses to the moral foundations questionnaire and align them to LLM-guided mappings of moral foundation scores to our defined cultural variables.

{\color{black}

\section{Background} \label{sec:preliminaries}

\subsection{Theoretical Models}
\label{sec:background-models}
\paragraph{World Values Survey.} 
The World Values Survey (WVS) is a widely used international survey of social, political, economic, religious, and cultural values~\cite{worldvaluessurvey}. It covers a broad range of countries across multiple survey waves since the 1980s, providing a rigorous and publicly available resource for studying cultural values and their evolution over time.

\paragraph{Inglehart--Welzel Cultural Map.}
The Inglehart--Welzel (IW) Cultural Map positions societies in a two-dimensional space defined by two principal dimensions of cross-cultural variation:  
\textit{Traditional} (bottom) vs.\ \textit{Secular-rational} (top) values 
and \textit{Survival} (left) vs.\ \textit{Self-expression} (right) values \cite{inglehart_welzel_2005}. 
According to the WVS 
\cite{worldvaluessurvey}, 
each point in the map is built from two factor scores obtained by applying factor analysis over a set of 10 WVS indicators:  \textit{happiness}, 
\textit{interpersonal trust}, 
\textit{respect for authority}, 
\textit{petition signing}, 
\textit{importance of God}, 
\textit{justifiability of homosexuality}, 
\textit{abortion}, 
\textit{national pride}, 
\textit{post-materialism}, 
and   \textit{autonomy index}.  
Country positions are   derived by computing within-country survey-weighted averages of these two factor scores, yielding each country’s coordinates on a 2D cultural map.

\paragraph{Moral Foundations Theory.} 
 Moral Foundations Theory (MFT) posits that moral judgments arise from   psychological foundations that cultures develop into distinct moral values, making it a descriptive theory of human morality.\footnote{\url{https://moralfoundations.org/}} 
The original MFT   identifies five moral foundations---\textit{care}, \textit{fairness}, \textit{loyalty}, \textit{authority}, and \textit{purity}---with subsequent revisions 
splitting \textit{fairness} into \textit{equality} and \textit{proportionality} \cite{atari2023morality}.

\subsection{Data and Resources}\label{sec:data}

\paragraph{World Values Survey Wave 7 (WVS-7).}
To access WVS resources, we refer to its Wave 7, which corresponds to a country-pooled datafile (Version~5.0) \cite{wvs7v5}.  
The WVS-7 master questionnaire, administered between 2017 and 2022, is organized into thematic blocks covering a broad range of cultural and social domains, including \emph{Social values, attitudes \& stereotypes}; \emph{Happiness and well-being}; \emph{Social capital, trust \& organizational membership}; \emph{Postmaterialist index}; \emph{Religious values}; \emph{Ethical values and norms}; and \emph{Political interest \& political participation}.

\paragraph{WorldValuesBench.} 
We also use WorldValuesBench (WVB) \cite{zhao-etal-2024-worldvaluesbench}, a benchmark derived from WVS-7 that proposes a task of \emph{multi-cultural value prediction} as producing a rating response to a value question conditioned on demographic attributes.
 Specifically, we use WVB-PROBE, a compact probe split (subset of the test set)   focusing on 36 value questions and three demographic variables (\textit{continent}, \textit{residential area}, \textit{education level}), with stratified sampling to ensure demographic diversity.  
We will  exploit  WVB-PROBE to evaluate whether the personas induce culturally plausible value profiles, without the computational cost of processing the full  WVS-7.

\paragraph{Moral Foundations Questionnaire-2 (MFQ-2).} 
This was  designed to measure moral aspects  generalizing across cultural contexts \cite{atari2023morality}. Building on evidence from 25 populations, MFQ-2  consists of 36 declarative items (six per moral foundation) rated on a 5-point Likert scale from \emph{Does not describe me at all} (1) to \emph{Describes me extremely well} (5). The authors report evidence for cross-cultural reliability and validity, with  predictive performance improvement over  the previous version (MFQ-1) on a broad set of external criteria.

}

\section{Cultural Variables  and Persona Generation}
\label{sec:resource}

\noindent\textbf{Cultural Variables.\ }
We identify   a set of 10 cultural variables, which encompass the main indicators  from WVS-7 that are used in constructing the Inglehart–Welzel axes and related dimensions, namely:  \textit{religiosity}, \textit{child-rearing values}, \textit{moral acceptability}, \textit{trust and social capital}, \textit{political participation}, \textit{national pride}, \textit{happiness}, \textit{gender equality}, \textit{materialism orientation}, \textit{attitude toward outgroups}. 
Each of the variables is defined as a categorical variable with admissible values that    are coded ordinally  
but conceptually categorical for interpretation in terms of the Inglehart–Welzel axes.  
Due to space limits, we report in  \textit{Appendix~\ref{app:wvs-conditioning}} all details  regarding the conceptual definition of our defined cultural variables, their admissible values, 
and links to the WVS-7 questionnaire IDs,  interpretation and relation with Inglehart–Welzel axes.

Let us denote with $\mathcal{V}=\{v_i\}_{i=1}^{10}$ our defined set  of cultural variables, where each variable $v_i$ has admissible values, or levels,  $l(v_i)=\{\ell_{i1}, \ldots, \ell_{ik_i}\}$.  
We compute the \textit{cultural configuration space} $\mathcal{C}(\mathcal{V})$ of $\mathcal{V}$  by taking the Cartesian product over all variable levels, i.e.,  $\mathcal{C}(\mathcal{V})=\prod_{i=1}^{10} l(v_i)$. This leads to a total of \textbf{93\,312} unique \textbf{cultural configurations}.

\vspace{1mm}
\noindent\textbf{Culturally-grounded Personas.\ }
 Each cultural variable configuration is to be used to condition the generation of  a persona profile. 
 To this purpose, we design a \textit{persona generation prompt template}---shown in
 \textit{Appendix~\ref{app:prompts}}
---where a cultural configuration is provided as the  conditioning parameter in input,     
and each   persona profile is generated to contain 
 (i) profile metadata (name, age, gender, occupation, and a plausible country/region), (ii) a short bio, and (iii) an explicit cultural variable mapping that grounds attitudes and behaviors in each of the cultural variables in $\mathcal{V}$.

By using  this template to prompt an LLM, we generate a set $\mathcal{P}$ of persona profiles, each corresponding to one of the  93\,312 cultural configurations. 
To do this, we choose   \textsf{gpt-oss-20B}   \cite{gpt-oss} 
which offers a favorable trade-off between model size and competitive reasoning, generation, and instruction-following capabilities, while remaining open-weight, thus suitable for model transparency and reproducibility.  
 We also repeated the three analyses with \textsf{Qwen3.5-9B} and \textsf{Llama3.1-8B-Instruct}, using the same cultural conditionings and evaluation settings (\textit{Appendix}~\ref{sec:qwen_llama}).

Looking at the distribution of demographics (age, gender, and occupations) extracted from the  metadata fields of the generated personas, we found a gender balance with  48.96\% `man', 47.98\% `woman' and  the remainder comprising non-binary or other/unspecified labels; age is strongly concentrated in the 30--39 and 40--49 ranges, and the most frequent occupation categories are `education \& academia' (21.37\%), `social services \& nonprofit' (13.28\%), `sales \& marketing' (12.69\%), and `tech / IT / data' (8.63\%).  
We also examined the \textit{country frequency distribution} of the generated personas, finding it as a  long-tailed already within the top 100 most frequent countries ($<$1\% for many countries), with  North America and Europe dominating (led by the United States), followed by Asia, while other continents appear with smaller shares. 
Details on the demographic and  country distributions of the generated personas, together with lexical statistics of their profiles,  are reported in \textit{Appendix~\ref{app:demographics}}.

\textit{Note that our defined cultural variable framework as well as a stratified sample of generated personas (extremes, axis-aligned cases, central cases, and regional representatives) was reviewed with the support of  
 a team of anthropologists at the} \textit{Institute for Studies on the Mediterranean (ISMed)} (ismed.cnr.it),  
 whose 
 feedback was   useful  to   assess   the plausibility of the     persona descriptions.

{\color{black}

\section{Methodology} 

 \begin{figure*}[t!]
 \centering
\includegraphics[width=0.95\linewidth]{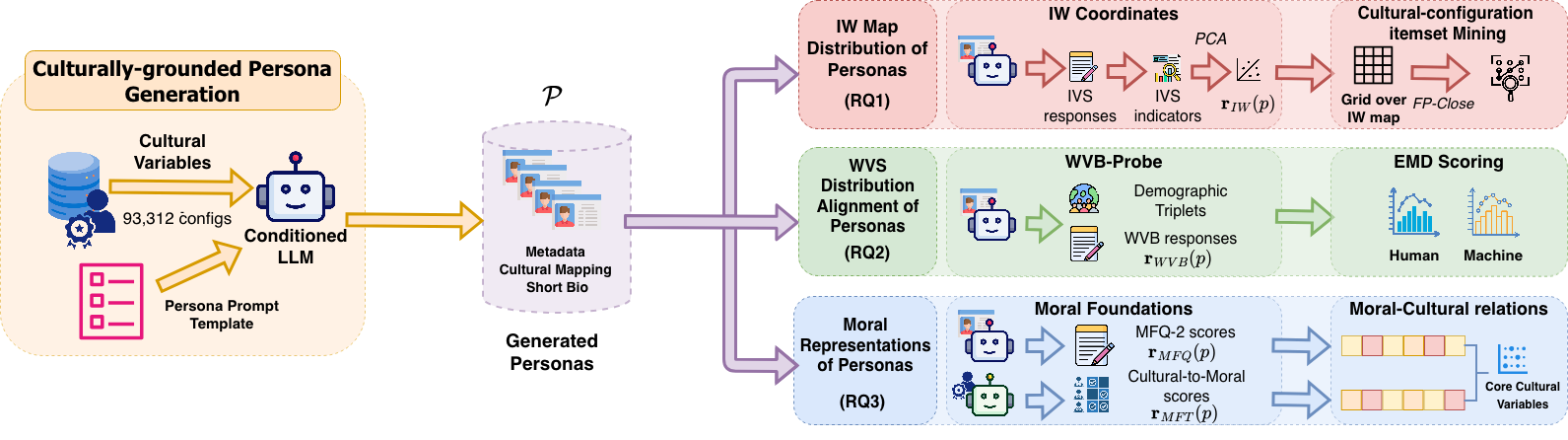}
 \caption{Overview of the main flows and data modules in our proposed framework}
 \label{fig:overview}
\end{figure*}

 We examine our  generated culturally-grounded  personas $\mathcal{P}$ through complementary analytical lenses leveraging LLM conditioning on these personas.  
We define four persona representation models 
 to  address  the stated RQs: 
 (i) IW-based embeddings of personas  for mining cultural patterns in the IW map   (\textbf{RQ1}), 
 (ii) WVS cultural profiles based on persona-conditioned LLM responses to the WVB-Probe questions (\textbf{RQ2}),  
 and 
 (iii) two  moral-foundation representations of personas based on persona-conditioned LLM responses to   MFQ-2 and on   aggregation of LLM-judged moral foundations scores   mapped to our cultural variables (\textbf{RQ3}).

Figure~\ref{fig:overview} provides an illustrative overview of the proposed methodological framework, with the three branches on the right corresponding to the RQs outlined above. These branches are discussed in detail in the following sections.

\subsection{IW Map Distribution of Personas (RQ1)}
\label{subsec:geometric-iw}

\noindent\textbf{Computing the IW map coordinates.\ } 
We first  project each persona into the IW  map by extracting coordinates $\mathbf{r}_{IW}(p)=(z_1, z_2)$ that represent the persona's position relative to the  IW  axes. 

To this aim, we resorted to the methodology in \cite{tao2024cultural} to reconstruct country-level coordinates from responses collected in the WVS and European Values Study (EVS), using the longitudinal EVS/WVS trend resources (\S Data Download in \cite{worldvaluessurvey}), 
 i.e., the \emph{Integrated Values Surveys (IVS)} obtained by merging the EVS Trend File and the WVS Trend File, which aggregate nationally representative survey waves and provide harmonized indicators underlying the IW dimensions. 
In our setting, we condition the  LLM  with a persona $p$ and ask it to respond  the IVS questions corresponding to the ten WVS indicators 
(cf.  Sect.~\ref{sec:background-models}). 
%
Then, following the guidance provided by the WVS Association (\S Data and Documentation in \cite{worldvaluessurvey}),  Principal Component Analysis 
with varimax rotation is applied to the standardized individual-level responses. Accordingly, the first two components explain 39\% of the variation in the data, with PC1 identifying the IW axis of \textit{Survival vs. Self-Expression} and PC2 the IW axis of \textit{Traditional vs. Secular Values}. 
Finally, a rescaling is applied to produce the IW coordinates $z_1$ and $z_2$, respectively,   used for the mapping:   
$z_1 = 1.81 \times PC1 + 0.38$ and $z_2 = 1.61 \times PC2 - 0.01$ \cite{hall2025aivalues}.

\vspace{1mm}
\noindent\textbf{Mining cultural patterns in the IW space.\ }
Upon the calculation of all points $\mathbf{r}_{IW}(p)$, we then carry out an \textit{exploratory mining} process which is comprised of two steps: \textit{Step 1} -- induce a partitioning of the set $\{\mathbf{r}_{IW}(p)\}$ on the IW plane into spatially contiguous regions, and \textit{Step 2} -- characterize the most relevant persona-conditioning cultural-configurations. We describe each step    next.

\noindent
$\bullet$ \textit{Step 1}. We partition   the IW map based on a regular grid, as we expect no evidence of cluster structure over the  $\mathbf{r}_{IW}(p)$ representations (cf. Sect. \ref{sec:rq1results}).   
  Each grid cell is set  with side length 1 for a sufficiently fine-grained discretization of the space.

\noindent
$\bullet$ \textit{Step 2}. 
Once computed the cultural regions in the IW map, 
each persona  is represented in a \textit{transactional format} based on the associated conditioning cultural configuration. 
 For each  grid cell, a transactional mini-dataset is built where each persona  $p$ (corresponding to its point  $\mathbf{r}_{IW}(p)$ located in that cell) is modeled as a set of items,  where an item is an element (i.e., variable-value) of the cultural configuration originally associated with $p$. 

Each transactional mini-dataset specific of a  grid cell is then fed to a \textit{frequent-pattern-mining method} to induce the most frequent   sets of cultural configuration items, in that cell. 
To this aim, we used   FPClose~\cite{GrahneZ05} to retain only the \textit{closed-frequent itemsets} (i.e., frequent itemsets having no superset with identical support). 
%
We start with minimum support sufficiently low (0.2 in our experiments) to include itemsets covering all cultural variables across all cells. 
Then, for the sake of presentation of each cell, we select   the most representative cultural-configuration itemsets:   given a cultural-configuration itemset $T$ and  denoted as $s_c(T)$ its support   within a cell $c$ (i.e., the fraction of personas in $c$ whose cultural-configuration itemsets contain $T$),  
we retain only itemsets whose support is a  fraction $\rho \in [0,1]$ of the maximum support $s_c^{\max}$ observed in $c$.

{\color{black}

\vspace{-1mm}
\subsection{WVS Distribution Alignment of Personas~(RQ2)} 
\label{sec:meth-wvb}
\vspace{0.5mm} 
We model each   persona $p$ as a vector $\mathbf{r}_{WVB}(p)$ representing, for each of the 36 probe questions from the WVB-Probe, the categorical response,  
obtained by conditioning the LLM on   $p$---prompt template    in \textit{Appendix~\ref{app:prompts}}.  
Since the WVB links the questions to the demographic variables of continent, residential area and education level (cf. Sect. \ref{sec:data}), the prompt also assigns a persona $p$  to a  demographic triple.   
Subsequently, we aggregate the WVB responses corresponding to personas  into groups each sharing the same demographic triple. For each   group  $g$ and probe question $q$,   we obtain a   distribution $P_{g,q}$ of  the response categories provided by the LLM. 

WVB-Probe also provides reference human-based distributions for each group and question. Therefore, we   measure the distributional alignment  by first comparing  the distance between $P_{g,q}$ and the corresponding  distribution $H_{g,q}$  using the \textit{Earth Mover's distance} (EMD),  
defined as 
$EMD(P,H)=\sum_{k=1}^{K}|F_P(k)-F_H(k)|$,  
with $K$ as the  number of response-categories for the WVB-Probe,   $F_P(k)$ and $F_H(k)$ denoting the cumulative distribution function  of $P_{g,q}$ and $H_{g,q}$, resp., up to category $k$.
An  overall, distributional alignment  score is   computed as  $1-\overline{EMD}$, where $\overline{EMD}$ is the average EMD  over all questions and groups.
}

{\color{black}

\vspace{-0.5mm}
\subsection{Moral Representations of Personas (RQ3)} 
\label{subsec:alignment}

Based on the moral foundations theory referring to 6 moral foundations (cf. Sect. \ref{sec:background-models}), 
we define two moral-foundation representation models for any given  culturally-grounded generated persona. The two representation  models are devised to be  alternative in that the one is based on the response to the MFQ-2 questionnaire items after administering them to an LLM prompted with a culturally-grounded persona, while the other is produced by aggregating moral foundation scores directly mapped to our defined cultural variables. 
As discussed later in the experimental part, the ultimate goal is to assess   the alignment of the two moral representations of personas.  

\vspace{1mm}
\noindent\textbf{MFQ-based Representation.\ }
For each persona $p \in \mathcal{P}$, we prompt an  LLM with  the full description of $p$ and elicit numeric responses to all 36 MFQ-2 items---prompt templates shown in \textit{Appendix~\ref{app:prompts}} (Fig.~\ref{fig:mfq2-prompt}). As a result,  $p$ is associated with a 6-dimensional representation $\mathbf{r}_{MFQ}(p)=[y_{p,m}]_{m\in\mathcal{M}}$, 
with $\mathcal{M}$ denoting the set of 6 moral foundations and  each  $y_{p,m}$ being  computed as the average over the   5-point Likert  items in the MFQ-2 questionnaire associated with the particular moral foundation $m$  \cite{atari2023morality}.  

\vspace{1mm}
\noindent\textbf{MFT-inferred Representation.\ }
We   define a scoring function as an \textit{oracle} to assign each value of a cultural variable with a 5-point Likert score for each of the moral foundations. 
We provide the oracle with full definitions of the moral foundations and the cultural variables in the prompt used to ask the oracle generate the mapping scores. 
To choose the oracle, we experimented with   recently released powerful LLMs---Google Gemini 3 Pro \cite{gemini3}, OpenAI GPT-5.2 \cite{chatgpt-52}, Anthropic Claude Sonnet 4.5 \cite{claude-sonnet}, in addition to the default LLM used for the other tasks. 
\textit{Note that our supporting team at ISMed finally validated the oracle.}

The resulting  moral-foundation scoring model is 
a matrix of shape 32 (values of the cultural variables) $\times$ 6 (moral foundations), with 5-point Likert-scale values. 
For any $p \in \mathcal{P}$,  a 6-dimensional representation $\mathbf{r}_{MFT}(p)$ 
is derived by aggregating the row vectors from the model-matrix  corresponding to the cultural configuration items of $p$.

}

\section{Experimental Evaluation} 
\label{sec:results}

\begin{figure}[t!]
    \centering
     \includegraphics[width=0.32\textwidth]{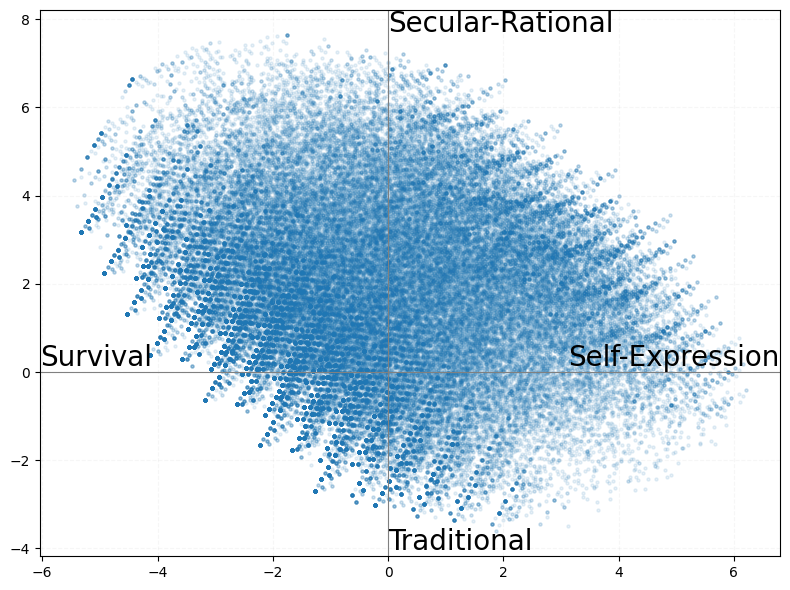} 
     \caption{{\color{black}
     Inglehart--Welzel map of all our synthetically-generated culturally-grounded personas.   Axes correspond to \emph{Traditional} (downward), \emph{Secular-rational} (upward),   \emph{Survival} (leftward), and \emph{Self-expression} (rightward) IW dimensions.  }  }
  \label{fig:iw-voronoi}
  \vspace{-3mm}
\end{figure}

{\color{black}

\subsection{Results for RQ1}
\label{sec:rq1results}

 Figure~\ref{fig:iw-voronoi} shows the distribution of the   $\mathbf{r}_{IW}(p)$ representations of  personas (cf. Sect.~\ref{subsec:geometric-iw}), 
 which form  a slightly tilted, elliptical cloud, indicating a mild upward shift and diagonal structure. 
 This reflects a tendency to emphasize the Secular dimension over the Traditional one, together with a negative correlation between Self-expression and Secular values. The highest density of points lies in the mid-to-low ranges of both Secular and Self-expression, with sparser distribution toward the extremes.

 To further assess the impact of culturally grounded persona conditioning, we   evaluated the LLM on IVS questions without any persona conditioning. As shown in \textit{Appendix~\ref{app:iw-map}} (Fig.~\ref{fig:iw-countries}), the baseline model’s IW coordinates lie in the upper-right quadrant, indicating a tendency toward Secular and Self-expression values.
 Moreover, the computed $\mathbf{r}_{IW}(p)$ points span a much broader region of the IW map compared to typical country-level IW coordinates, which are more tightly clustered around the center (cf. \textit{Appendix~\ref{app:iw-patterns}}, Fig.~\ref{fig:iw-overlap}). This wider spread reflects the greater heterogeneity expected at the individual level due to cultural variation.

\begin{table}[t]
\centering
\small 
\setlength{\tabcolsep}{2.5pt}
\renewcommand{\arraystretch}{1.05}
\begin{tabular}{
p{0.49\columnwidth}
r
c
c
r
}
\toprule
\textbf{Closed frequent}
& \textbf{\#}
& \textbf{avg.}
& \textbf{range}
& \textbf{avg./cell}
\\
\textbf{Cultural config.} & & & & \\
\midrule

\textit{happy = very happy;}\\
\textit{trust = most people trusted}
& 21 & .735 & [.47, .99] & 353 \\

\midrule

\textit{happy = not at all happy;}\\
\textit{trust = cannot trust people}
& 15 & .806 & [.51, 1] & 370 \\

\midrule

\textit{happy = very happy;}\\
\textit{pride = very proud;}\\
\textit{trust = most people trusted}
& 11 & .796 & [.55, 1] & 131 \\

\midrule

\textit{happy = not at all happy;}\\
\textit{pride = not proud;}\\
\textit{trust = cannot trust people}
& 10 & .808 & [.57, 1] & 205 \\

\midrule

\textit{child rear. = obedience faith;}\\
\textit{trust = cannot trust people}
& 6 & .670 & [.47, .84] & 577 \\

\midrule

\textit{moral accept. = always justif.;}\\
\textit{trust = most people trusted}
& 4 & .658 & [.59, .81] & 132 \\

\bottomrule
\end{tabular} 
\caption{{\color{black}Sample of  closed frequent cultural-configuration itemsets discovered in the Inglehart--Welzel space of our synthetically-generated culturally-grounded  personas (min. support = 0.2,  $\rho$ = 0.5,  cell side length = 1, min. number of covered cells = 2).  \textbf{\# }indicates the number of cells; \textbf{avg.}, resp. \textbf{range}, denote the mean, resp. full range, of support values; \textbf{avg./cell} indicates the per-cell mean no. of personas. 
}}
\label{tab:iw-voronoi-patterns}
\vspace{-4mm}
\end{table}

\begin{figure*}[ht!]
    \centering
    \setlength{\tabcolsep}{1pt}

    \begin{subfigure}{0.16\textwidth}
        \centering
        \includegraphics[width=\linewidth]{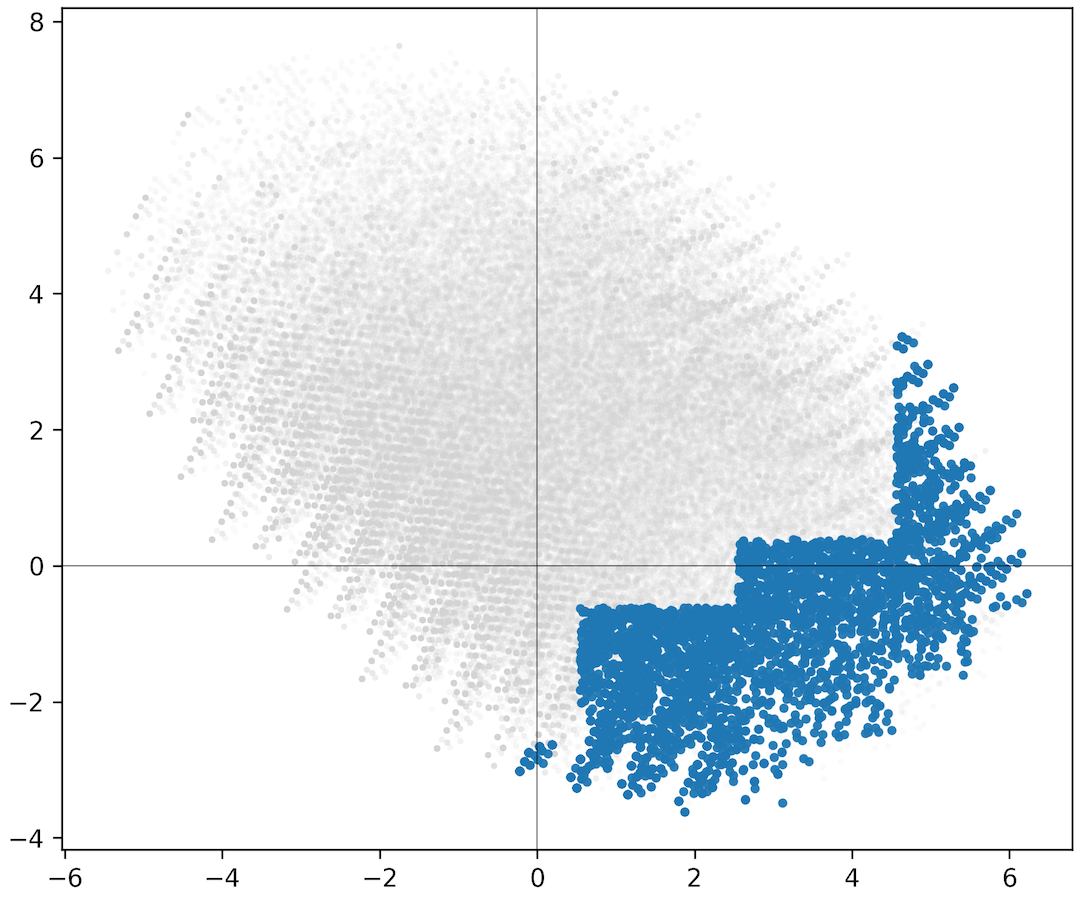}
        \caption{}
    \end{subfigure}\hfill
    \begin{subfigure}{0.16\textwidth}
        \centering
        \includegraphics[width=\linewidth]{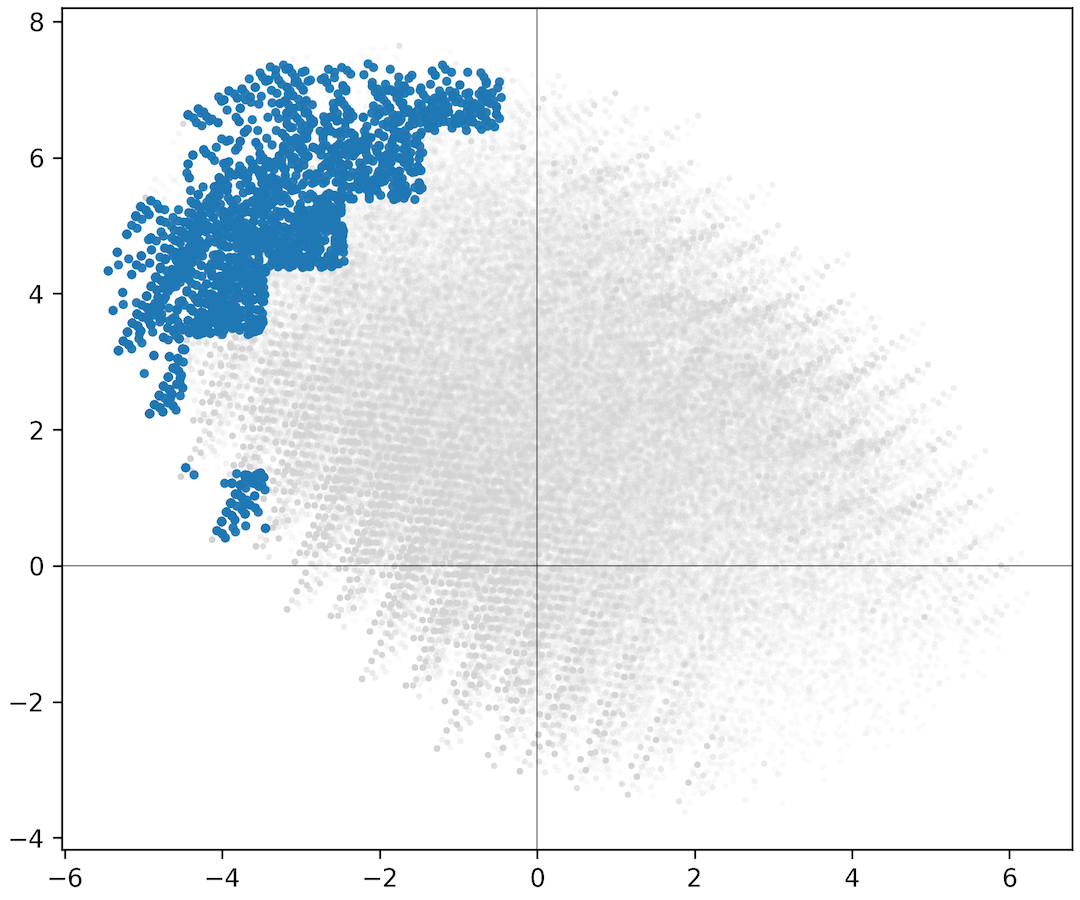}
        \caption{}
    \end{subfigure}\hfill
    \begin{subfigure}{0.16\textwidth}
        \centering
        \includegraphics[width=\linewidth]{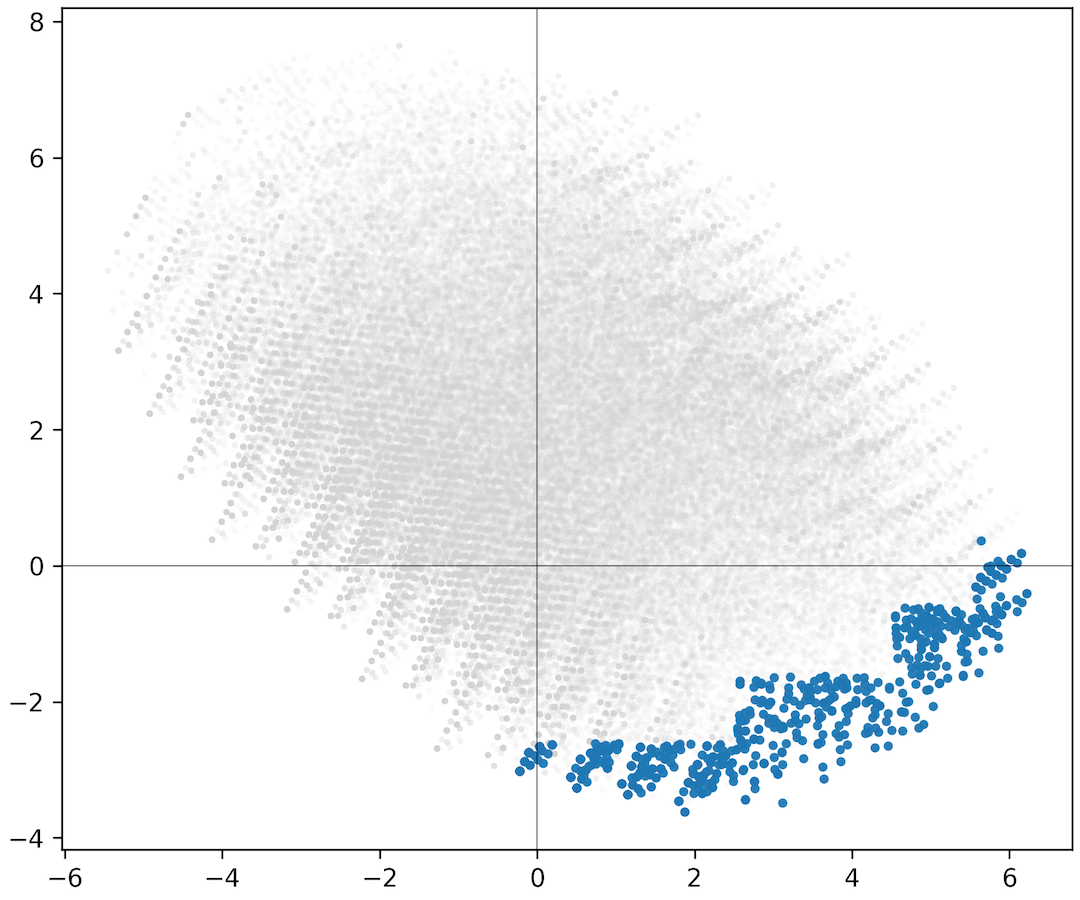}
        \caption{}
    \end{subfigure}\hfill
    \begin{subfigure}{0.16\textwidth}
        \centering
        \includegraphics[width=\linewidth]{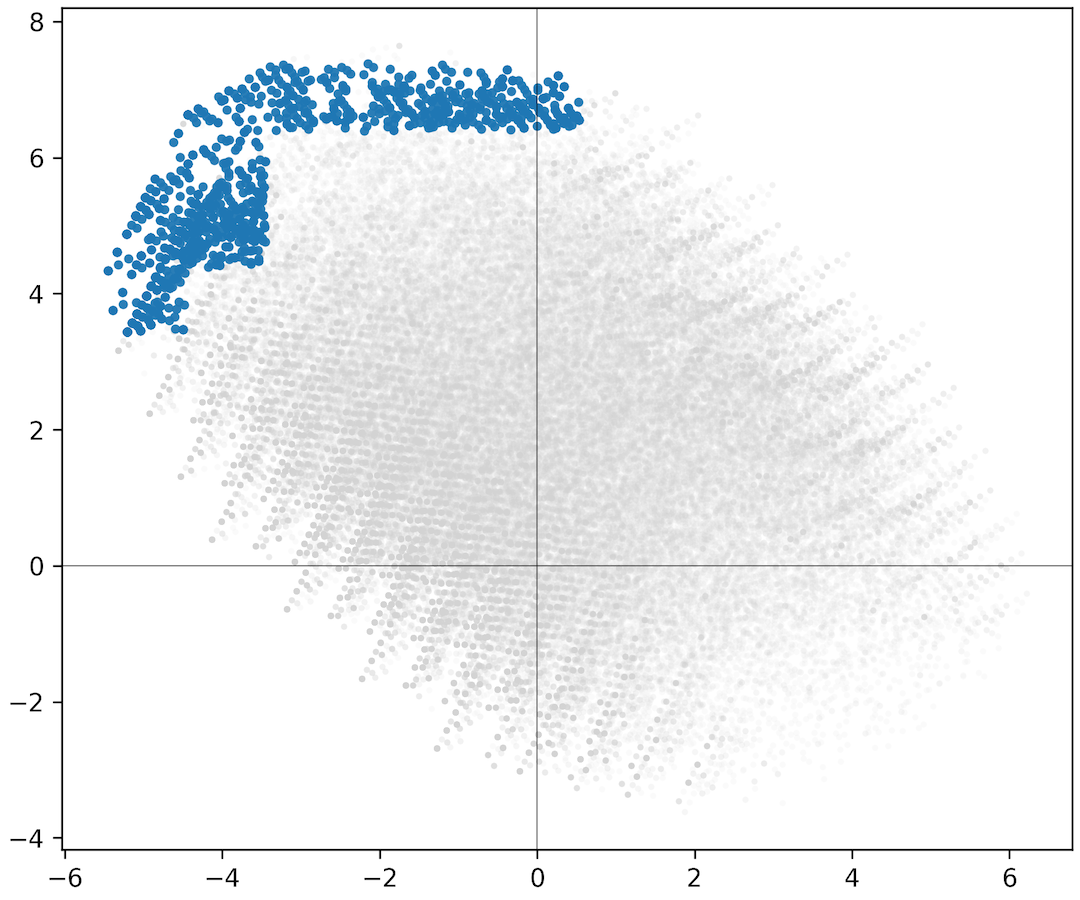}
        \caption{}
    \end{subfigure}\hfill
    \begin{subfigure}{0.16\textwidth}
        \centering
        \includegraphics[width=\linewidth]{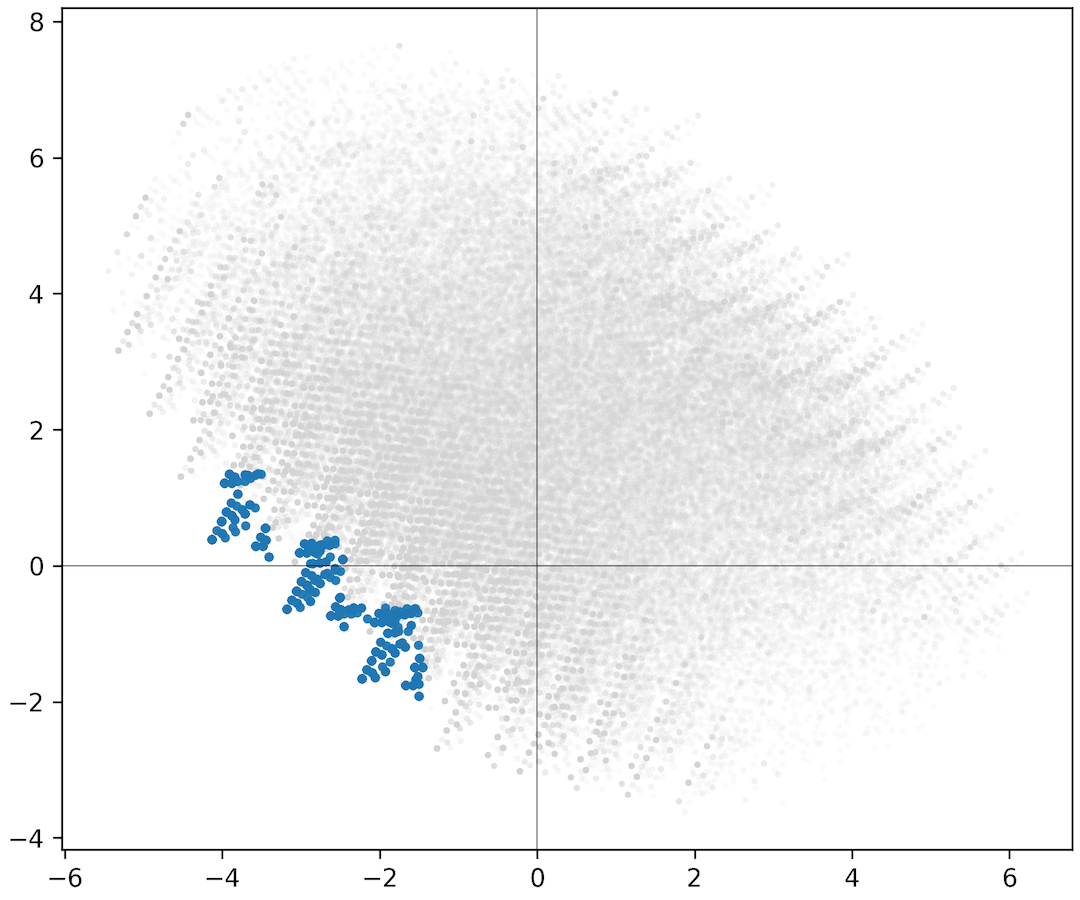}
        \caption{}
    \end{subfigure}\hfill
    \begin{subfigure}{0.16\textwidth}
        \centering
        \includegraphics[width=\linewidth]{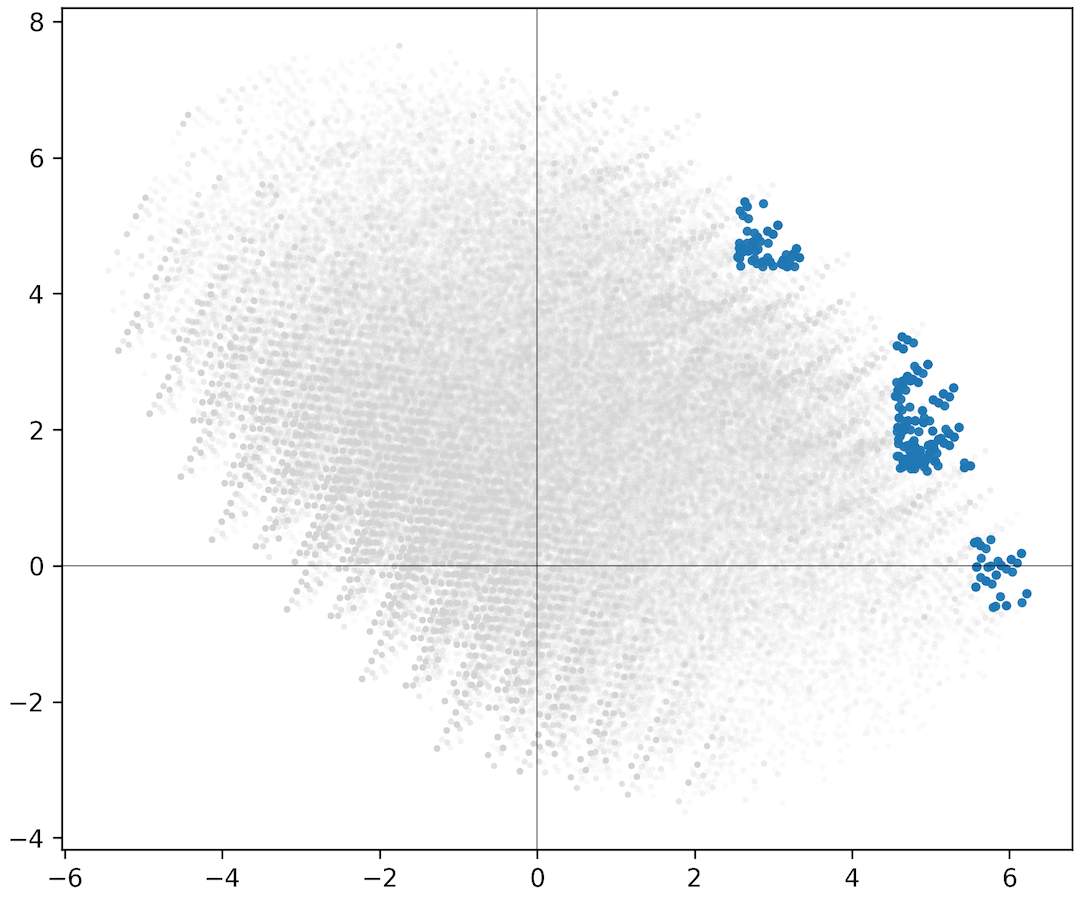}
        \caption{}
    \end{subfigure}

    \caption{{\color{black}Inglehart--Welzel map distribution  of   the personas  (blue points) satisfying the itemsets from rows in  Table~\ref{tab:iw-voronoi-patterns}.   
    }}
\label{fig:iw-voronoi-patterns}

    \label{fig:itemsets}
    \vspace{-2mm}
\end{figure*}

The closed frequent itemset mining over the grid  partitioning of the   $\mathbf{r}_{IW}(p)$ in  Fig.~\ref{fig:iw-voronoi}  identified 
138 distinct closed frequent itemsets. Table~\ref{tab:iw-voronoi-patterns} reports a representative subset of non-singleton patterns. The dominant structure is driven by \textsl{trust} and \textsl{happiness}, indicating a strong coupling between interpersonal trust and subjective well-being, with `high trust / very happy' configuration frequent across 21 cells (mean support 0.735), and `low trust / not at all happy' occurring across 15 cells (mean support 0.806).  
Both \textsl{trust}/\textsl{happiness} items  also co-occur frequently with \textsl{national pride}, where `high-trust/high-happiness' aligns with `very proud' and the opposite configuration with `not at proud at all',   suggesting that national pride may modulate the trust–well-being relationship.
Beyond this core pattern, additional regularities (two bottom  rows of Table~\ref{tab:iw-voronoi-patterns}) show that `cannot trust' is associated with \textsl{child-rearing} values such as `obedience' and `faith', and also co-occurs with stronger endorsement of \textsl{moral acceptability} (`always justifiable').

Figure~\ref{fig:itemsets} shows that the closed frequent cultural-configuration itemsets have clear and interpretable spatial footprints in the IW space. 
The two \textsl{trust}/\textsl{happiness}  patterns occupy opposite regions: `high trust / very happy' concentrates in the high Self-expression/Traditional  quadrant (bottom-right), while `low trust / not at all happy' appears in the opposing Survival, Secular-rational quadrant (top-left).
Adding \textsl{national pride} refines these cells:  
`very proud' contracts the `high-trust/high-happiness' footprint towards the extreme right tail (the strongest self-expression side of the space), whereas `not proud at all' contracts the `low-trust/low-happiness' footprint towards the upper ridge (the most secular-rational end of the vertical axis), yielding smaller but still coherent neighborhoods.
By contrast, the remaining patterns have smaller and less contiguous footprints, consistent with more localized cultural niches, as in the case of Figures~\ref{fig:itemsets} (e)–(f), showing the \textsl{obedience faith} with `low trust' areas in the traditional/survival side of the map, whereas `high moral acceptability' with `high trust' portions  on the opposite, more secular/self-expression side.

  While closed frequent itemsets characterize  the periphery of the IW map, many central areas retain only singleton closed frequent itemsets (see \textit{Appendix~\ref{app:iw-patterns}}, Fig.~\ref{fig:iw-voronoi-singleton} for   a representative sample of singleton patterns).  
In line with the IW ``geometry'',  such areas are  mostly  covered by singleton itemsets associated with intermediate levels of the cultural variables (e.g.,  `not very happy' and `rather happy' for \textsl{happiness}); by contrast, extreme categories tend to be placed  toward peripheral regions.  

}

 }

\subsection{Results for RQ2}
\label{sec:rq2_results}

\begin{table}[t]
\centering
\setlength{\tabcolsep}{5pt}
\scalebox{0.8}{
\begin{tabular}{lrrr}
\toprule
\textbf{Aggregation} & 
$(1-\overline{EMD})$ & 
\%$\mathit{EMD}_{<0.4}$ & 
\%$\mathit{EMD}_{<0.2}$\\
\midrule
Unweighted  & 0.792 & 90.12 & 56.98 \\
Weighted   & 0.809 & 94.25 & 59.06 \\
\bottomrule
\end{tabular}
}
\caption{{\color{black}WVB-Probe alignment between the  persona distributions and   human-reference distributions, aggregated by demographic groups.  
Lower Earth Mover's distance (EMD) indicates closer alignment with WVS reference distributions.
}}
\label{tab:wvb-probe-summary}
\vspace{-2mm}
\end{table}

{\color{black}
\noindent\textbf{WVB-Probe.\ }  
We evaluate WVB-Probe alignment using personas whose demographic triples match groups with available human reference distributions, covering 92,710 personas (>99\% of the dataset) and 45 out of 46 demographic groups (cf. Fig.~\ref{fig:wvb-demotriad} in \textit{Appendix} for details). Table~\ref{tab:wvb-probe-summary} reports EMD-based alignment under two aggregation schemes: unweighted averaging across groups and weighted averaging by group size. The results show strong overall alignment, with scores of 0.792 (unweighted) and 0.809 (weighted).

\vspace{1mm}
\noindent\textbf{Question-level Agreement.\ }
Table~\ref{tab:wvb-probe-summary} also reports the proportion of probe questions achieving \textit{moderate alignment} ($\mathit{EMD}<0.4$) and \textit{high alignment} ($\mathit{EMD}<0.2$). Moderate alignment is nearly saturated across groups, with on average 90.12\% of the 36 questions per group meeting the threshold (94.25\% when weighted), indicating generally low distributional discrepancy.
In contrast, the stricter high-alignment regime is much more selective (less than 60\% of cases satisfying $\mathit{EMD}<0.2$).

\begin{table}[t]
\centering
\setlength{\tabcolsep}{2.3pt}
\scalebox{0.8}{
\begin{tabular}{lrccc}
\toprule
\textbf{Group} & \textbf{\#personas} & $1-\overline{\mathit{EMD}}$ & \%$\mathit{EMD}_{<0.4}$& \%$\mathit{EMD}_{<0.2}$ \\
\midrule
(E \textbar U \textbar T) & 32,716 & 0.803 & 94.4 & 52.8 \\
(N-A \textbar U \textbar T) & 14,984 & 0.832 & 97.2 & 69.4 \\
(AS \textbar U \textbar T) & 12,868 & 0.825 & 97.2 & 66.7 \\
(S-A \textbar U \textbar T) & 4924 & 0.799 & 94.4 & 55.6 \\
(N-A \textbar R \textbar T) & 4586 & 0.817 & 91.7 & 61.1 \\ 
\bottomrule
\end{tabular}
}
\caption{\color{black}WVB-Probe alignment between the  persona   and the  human-reference distributions, 
for each of the five mostly represented groups. 
Each group consists of (Continent \textbar Settlement \textbar Education), where E = Europe, N/S-A = North/South America,   AS = Asia, 
U = Urban, R = Rural, T = Tertiary. 
}
\label{tab:wvb-probe-top10}
\vspace{-2mm}
\end{table}

\vspace{1mm}
\noindent\textbf{Group-level Patterns.\ }
Table~\ref{tab:wvb-probe-top10} reports  alignment for the five most represented demographic profiles (\textit{Appendix~\ref{app:wvb}}, Fig.~\ref{fig:wvb-demotriad} for  all groups).
Alignment scores ($1-\overline{EMD}$) are consistently high ($0.799$--$0.832$), with `urban' and `tertiary'    dominating the group profiles.  
(`Europe' \textbar `urban' \textbar `tertiary') is the largest group by count but exhibits a substantially lower fraction of strongly aligned questions ($52.8\%$ with $\mathit{EMD}<$0.2) compared to groups with the same residential area and education level but different continent (e.g., `North America', at $69.4\%$, and `Asia', at $66.7\%$), despite comparable overall scores. 
A complete breakdown over all 45 demographic groups is reported in \textit{Appendix~\ref{app:wvb}} (Table~\ref{tab:wvb-probe-all-groups}). Global alignment scores  span  $0.63$ to  $0.84$.

\subsection{Results for RQ3}
To address RQ3, we   evaluated  the two moral representations for each   culturally-grounded persona $p$: those   based on  MFQ-2 ($\mathbf{r}_{MFQ}(p)$) and  those  inferred from the mapping of the cultural variables ($\mathbf{r}_{MFT}(p)$) (cf. Sect.~\ref{subsec:alignment}). 
For $\mathbf{r}_{MFT}(p)$, we use outputs from Gemini 3 Pro, as it produced   more informative score distributions than  other models which mostly assigned   neutral scores   
across inputs.

For each moral foundation, we compared $\mathbf{r}_{MFQ}(p)$ with  the default $\mathbf{r}_{MFT}(p)$, i.e.,  obtained by averaging over the full set of ten cultural configuration items of $p$;  
however, given the substantial differences in trends and scale between the two representations (\textit{Appendix~\ref{app:mfq}}, Fig.~\ref{fig:mfq2-all10}), we further investigate which \textit{cultural variables are most relevant to each foundation} when constructing the $\mathbf{r}_{MFT}(p)$ representations.
To this end, we perform a \textit{foundation-specific cultural-variable selection} over the $\mathbf{r}_{MFT}(p)$ representations. After splitting the data into training and validation sets, we apply a greedy forward selection procedure that iteratively adds one cultural variable at a time from an initially empty set. Each candidate is evaluated by measuring the validation $R^2$ after fitting a simple linear correction on the training data to better align predictions with the reference  $\mathbf{r}_{MFQ}(p)$ values.
To produce statistically robust results, the foundation-specific greedy procedure was repeated for 50 independent runs. 
For each moral foundation, we finally kept the set of variables, hereinafter called \textit{core-set}, that were selected in at least 80\% of the 50 runs. 
The discovered \textbf{core-sets} are:  
\textsl{child rearing value},  \textsl{gender equality},  \textsl{happiness},  \textsl{materialism orientation}, \textsl{social trust},  \textsl{tolerance diversity} for {Care}; 
\textsl{materialism orientation} only for 
{Equality} and for {Proportionality}; 
\textsl{national pride} for {Loyalty};
\textsl{child rearing value},  \textsl{gender equality},  \textsl{national pride},  \textsl{religiosity},  \textsl{tolerance diversity} for {Authority};
  \textsl{religiosity} for {Purity}.

Based on these core-sets, we  recomputed the $\mathbf{r}_{MFT}(p)$, finding  that the foundation-specific core-set-based aggregation tracks MFQ-2 trends more closely than the uniform 10-variable mean, with the clearest improvements for Loyalty, Authority, and Purity. 
(Details in   
\textit{Appendix}, Fig.~\ref{fig:mfq2-greedy}) 

 Focusing on each core  
 variable and corresponding  moral foundation (results   in 
\textit{Appendix}-Table~\ref{tab:mfq2-core-detailed}),  the mean moral-foundation scores vary monotonically with the levels of each cultural variable, indicating an approximately linear relationship.
However, the fraction of personas having a certain cultural configuration can vary significantly over the values  of the same core cultural-variable when considering moderate-to-strong association (score $\geq4$) with a moral foundation;  this particularly holds for Authority with \textsl{child rearing value}, \textsl{gender equality}, \textsl{religiosity},   for Purity and Loyalty.

By contrast, 
Care, Equality and, to a lesser extent,  Proportionality appear to be \textit{culturally-invariant}. 
 We interpret the above results through the   individualizing vs. binding distinction \cite{graham2013moral}. Individualizing foundations (Care, and Fairness—here as Equality and Proportionality) emphasize universal concerns about individuals' rights and welfare, whereas binding foundations (Loyalty, Authority, Purity) are more tightly connected to group cohesion, social norms, and tradition. 
Since our personas are conditioned on culturally grounded attributes such as religiosity, national pride, social trust, tolerance, and gender roles, stronger effects are expected for binding foundations. 
Consistently, the greedy procedure selects compact and stable core-sets for Loyalty, Authority, and Purity, while individualizing foundations show weaker dependence on cultural configuration. This  is further supported by the Spearman analyses in \textit{Appendix}-Fig.~\ref{fig:spear}, where cultural variables and IW indicators correlate more strongly with binding than with individualizing foundations.

}

 \section{Related Work}

Research on cultural representation and inclusion in LLMs employ semantic and demographic proxies \cite{adilazuarda-etal-2024-towards}, 
cultural conditioning with  minimal identity indicators (e.g.,  names \cite{pawar-etal-2025-presumed}, language  \cite{alkhamissi-etal-2024-investigating}) or cross-country benchmarks assessing distributional distance between models and human populations on personality traits \cite{dey-etal-2025-llms}. Also,  survey-based alignment evaluations  compare persona-prompted model responses with actual human respondents or survey-derived human profiles \cite{alkhamissi-etal-2024-investigating,tao2024cultural}, or 
%
 emphasize sensitivity to prompt format 
 \cite{kabir-etal-2025-break},   and the need for grounding and transparency  in generating  personas \cite{batzner-etal-2025-whose}.

Studies on moral assessments  associate  
judgements on   norms
across cross-national survey standards \cite{ramezani-xu-2023-knowledge}, analyze the alignment between moral generations and group-labeled human datasets 
\cite{he-etal-2024-whose}, or measure moral beliefs 
\cite{scherrer2023evaluating}. 
Others investigate persona-conditioning in simulated debates \cite{liu-etal-2025-synthetic} and extract moral preferences from daily dilemmas \cite{ChiuJ025}.
MFT/MFQ-based investigations  study political/demographic associations \cite{smith2025investigating} and role-play robustness \cite{abs-2511-08565}.  Moral Compass \cite{ijcai2025p1059} also provide a 
decision benchmark built around normative-ethics factors.

A  comparison with related work  across five aspects---persona prompting, reliance on human survey data and moral frameworks, analysis on the IW cultural map, and     evaluation goal (cf. \textit{Appendix \ref{appx:related}}, Table~\ref{tab:rw-overlap})---highlights that  our research uniquely combines the above aspects in a \textit{controlled audit}: we conceptualize personas based on WVS-derived cultural variables, 
and analyze them through well-established theoretical models and benchmarks to study human-values and moral foundations.

\section{Conclusions}
We presented a WVS-based framework to generate culturally grounded personas and assess how they align with established value and moral theories.

\paragraph{Lessons Learned.}
The conditioned cultural grounding yields personas whose induced responses are culturally positioned, broadly aligned with human demographic patterns, and differentiated in their moral profiles.

Regarding cultural positioning (RQ1), personas are broadly distributed over the   the Inglehart--Welzel space,   
although their configurations produce coherent local footprints: intermediate values mainly characterize central regions, while extreme values extend toward the periphery. 
The outcomes of our analyses, independently repeated   using three LLMS---\textsf{gpt-oss-20B},  \textsf{Qwen3.5-9B} and \textsf{Llama3.1-8B-Instruct}---share this broad spatial distributions and a common core of prominent cultural patterns, although their exact positions and some retained itemsets are model-dependent.

In terms of alignment with human groups (RQ2),  
personas are not only theoretically grounded in WVS-derived cultural variables, but also produce empirically plausible value distributions at the demographic-group level. 
Persona-conditioned responses broadly align with the human-reference distributions provided by WVB-Probe, reaching overall scores of approximately $0.78$--$0.81$ across the three models, with stronger alignment that can vary  across demographic groups, highlighting the importance of subgroup-level analyses. 

Our  culture--morality analysis (RQ3) reveals that cultural variables are selectively associated with moral foundations, where the  binding foundations (Loyalty, Authority, and Purity)   show clearer   variation than Care, Equality, and Proportionality,  
while the   core cultural variables selected for each foundation vary across the evaluated LLMs.

We emphasize that the proposed research questions provide complementary perspectives on the same underlying process:  explicit cultural conditioning   induces structured variation in the model responses; this variation can then be evaluated against human-reference distributions, and finally its implications for moral representations can be examined. The framework therefore connects cultural positioning,
empirical alignment, and moral characterization within a single audit pipeline.

\paragraph{Impact.}  
We believe the proposed framework has a significant impact on the development of  more culturally inclusive  AI systems by serving as a bridge between socio-psychological theory and controlled NLP evaluation. 
By enabling rigorous audits and benchmarking, the generated personas, which reflect multi-perspective annotations and various cultural configurations, support the assessment of fairness and alignment across specific target populations, including varied WVS groups and IW cultural regions. 
Our study facilitates counterfactual probing,  
by varying one cultural or demographic attribute while keeping the task fixed to investigate model sensitivity and the emergence of potential stereotypes. 
Culturally grounded personas can also be integrated into  
multi-agent systems and synthetic audience debates, to study complex social dynamics, negotiation, and persuasion.

\section*{Limitations}

Our  framework is English-centric in both persona descriptions and response elicitation. We do not test whether equivalent cultural configurations yield stable signals under multilingual prompting or culturally specific phrasing.  
Future work could explore multilingual and culturally localized elicitation setups to evaluate whether the same cultural specifications produce consistent signals across languages and prompt formulations.  

The generated personas are the result of  93,312 configurations (i.e., all possibile combinations of the 10 cultural variables and their admissibile values), which may contain  configurations that are     rare or sociologically implausible in real populations.  
Some discovered patterns in the IW space may therefore reflect the designed LLM-based generation procedure, which depends on the LLM's parametrization and training data, combinatorics rather than naturally occurring cultural profiles. Treating each configuration as equally likely describes what is possible under the variable scheme, not what is probable in human populations.   
We will consider 
reweighting strategies to better reflect empirical prevalence and reduce the influence of atypical configurations. 

Our defined cultural variables are derived from WVS-7 constructs and indicators, which comprise  surveys from 2017 to 2022. Since cultural norms evolve over time due to migration, crises, and social change, our alignment with static, survey-era patterns may not fully capture current or emerging cultural dynamics.

\section*{Ethical Considerations}
 
While we operationalize culture through interpretable WVS-derived variables and publicly available survey resources, this remains a proxy-based representation that captures only a subset of cultural phenomena, particularly  survey-measurable values and attitudes. As a result, our findings should not be read as a comprehensive account of culture, but rather as evidence about alignment under this specific operationalization.

We grounded personas in interpretable cultural variables and evaluated them against established socio-psychological frameworks. Nonetheless, the resulting profiles should be interpreted with caution, since they 
cannot capture the full complexity of lived human experiences, socialization processes, or situated moral decision-making. We also emphasize that the generated personas   are not intended to correspond to real individuals and do not have access to real-life personal or identifiable information.

As synthetic artifacts, the generated personas may potentially reinforce stereotyped associations, simplified cultural caricatures, or unrealistic combinations of demographic attributes and value profiles. In this respect, 
the metadata distributions (e.g., country, age, occupation) in our personas are uneven, showing over-representation of North America and Europe and   concentration in the 30–49 age range. This may introduce generation biases that are independent of the intended cultural variables, limiting worldwide representativeness. Also, the culture-to-country association is endogenous to the involved  LLMs and may reflect pretraining priors on relations  between national identity and cultural values.

Furthermore, our framework is conceived to be descriptive rather than prescriptive: it is not meant to rank cultures, define  ``correct'' moral views, or justify decisions about individuals or groups.

\bibliography{custom}

\appendix

\vspace{5mm}
\section{Experimental Environment}

All generations have been executed on a server with 8$\times$ NVIDIA A30 GPUs (24GB VRAM each), 764GB system RAM, 2$\times$ Intel Xeon Gold 6248R (96 CPU cores total), running Ubuntu Linux 20.04.6 LTS. We set \emph{temperature} $=1.0$ and \emph{top\_p} = $1.0$ as recommended by OpenAI.\footnote{\url{github.com/openai/gpt-oss?tab=readme-ov-file\#recommended-sampling-parameters}}

\section{Comparison with Related Work}
\label{appx:related}
Table \ref{tab:rw-overlap} summarizes a comparison between our work and related ones according to five dimensions. 

\begin{table}[h!]
\centering
\setlength{\tabcolsep}{3pt}
\scalebox{0.73}{
\begin{tabular}{lccccl}
\toprule
\textbf{Work} &
\textbf{P} &
\textbf{S} &
\textbf{M} &
\textbf{IW} &
\textbf{Evaluation} \\ 
\midrule
\citet{adilazuarda-etal-2024-towards} &
\cmark & \xmark & \xmark & \xmark & Proxy taxonomy \\

\citet{pawar-etal-2025-presumed} &
\cmark & \xmark & \xmark & \xmark & Names as proxy \\

\citet{dey-etal-2025-llms} &
\cmark & \xmark & \xmark & \xmark & Distributional distance \\

\citet{alkhamissi-etal-2024-investigating} &
\cmark & \cmark & \xmark & \xmark & Anthropol. survey simul. \\

\citet{tao2024cultural} &
\xmark & \cmark & \xmark & \cmark & Survey simulation \\

\citet{kabir-etal-2025-break} &
\cmark & \cmark & \xmark & \cmark & Prompt format \\

\citet{batzner-etal-2025-whose} &
\cmark & \xmark & \xmark & \xmark & Ground. \& transp. anal. \\

\citet{ramezani-xu-2023-knowledge} &
\xmark & \cmark & \xmark & \xmark & Survey-validated probing \\

\citet{he-etal-2024-whose} &
\xmark & \xmark & \cmark & \xmark & Corpus alignment \\

\citet{scherrer2023evaluating} &
\xmark & \xmark & \xmark & \xmark & Prompt marginalization \\

\citet{liu-etal-2025-synthetic} &
\cmark & \xmark & \xmark & \xmark & Multi-turn debate \\

\citet{ChiuJ025}&
\xmark & \cmark & \cmark & \xmark & Dilemma preferences\\

\citet{smith2025investigating} &
\cmark & \xmark & \cmark & \xmark & MFT probing \\

\citet{abs-2511-08565} &
\cmark & \xmark & \cmark & \xmark & MFQ robustness \\

\citet{ijcai2025p1059} & \xmark & \xmark & \cmark & \xmark & Ethics benchmark \\
\midrule
\textbf{Ours} &
\cmark & \cmark & \cmark & \cmark & Controlled audit \\
\bottomrule
\end{tabular}
}
\caption{Comparison  between related work and our setting along five axes: persona use (P), survey grounding (S), moral framework (M), IW-based structure (IW), and evaluation goal.}
\label{tab:rw-overlap} 
\end{table}

\section{Details on the Cultural Variables}
\label{app:wvs-conditioning}

{\color{black}
Table \ref{tab:wvs-conditioning} overviews our defined 10 cultural variables, which are used in our study to culturally condition the generation of persona profiles.  

It should be noted that our defined 10 cultural variables are derived from WVS-7 items  as a   subset     identified     to operationalize the two main Inglehart–Welzel dimensions of cross-national cultural variation. 
 More specifically, the 10 cultural variables are abstractions that often group more WVS-7 questions, as reported next:  
importance of religion (Q6); child-rearing priorities such as obedience, religious faith, and tolerance/respect (Q7--Q17); subjective happiness (Q46); generalized social trust (Q57); religiosity items centered on the importance of God and related beliefs/practices (Q164--Q175); moral acceptability judgments on a 1--10 justifiability scale (e.g., Q177--Q191); gender-role attitudes (e.g., Q29--Q33); postmaterialism priorities (Q152--Q157); and attitudes toward outgroups (Q18--Q26); political participation (Q199-Q234). 
 
 }

\newcolumntype{P}[1]{>{\raggedright\arraybackslash}p{#1}}
\renewcommand\cellalign{tl}   
\renewcommand\cellset{\renewcommand\arraystretch{1.0}} 
\begin{table*}
\centering
\footnotesize
\begin{tabular}{P{2.5cm} P{3.3cm} P{4.8cm} P{3.6cm}}
\toprule
\textbf{Cultural variable} & \textbf{Concept} & \textbf{Admissible values} & \textbf{Relation with IW axes}\\
\midrule

\texttt{religiosity} &
Importance and practice of religion; belief in God. &
\makecell[tl]{\texttt{very important} (10)\\
\texttt{quite important} (7--9)\\
\texttt{not very important} (4--6)\\
\texttt{not at all important} (1--3)} &
\makecell[tl]{Traditional (higher) $\rightarrow$ \\Secular (lower)}\\

\addlinespace
\texttt{child rearing value} &
Qualities desired for children (obedience/faith vs.\ independence/imagination). &
\makecell[tl]{\texttt{obedience-faith}\\
(selects obedience/religious faith)\\
\texttt{independence-imagination}\\
(selects independence/imagination)\\
\texttt{neutral} (no clear preference)} &
\makecell[tl]{Traditional (obedience-faith); \\Secular-Rational / Self-\\Expression (independence-\\imagination)}\\

\addlinespace
\texttt{moral acceptability} &
Justifiability of morally contested acts (e.g., abortion, euthanasia, divorce, suicide, homosexuality). &
\makecell[tl]{\texttt{never justifiable} (1--3)\\
\texttt{sometimes justifiable} (4--7)\\
\texttt{always justifiable} (8--10)} &
\makecell[tl]{Traditional (never) $\rightarrow$ \\ Secular (always), \\for moral items; \\Survival (never) $\rightarrow$ \\  Self-Expression (always), \\for tolerance items}\\

\addlinespace
\texttt{social trust} &
Generalized interpersonal trust (``most people can be trusted''). &
\makecell[tl]{\texttt{most people trusted} (1)\\
\texttt{cannot trust people} (0)} &
\makecell[tl]{Survival (0) \\Self-Expression (1)}\\

\addlinespace
\texttt{political participation} &
Civic engagement and activism. &
\makecell[tl]{\texttt{active participant}\\
(signed petition, joined protest,\\
contacted officials)\\
\texttt{passive participant}\\
(votes only, minimal engagement)\\
\texttt{nonparticipant} (no participation)} &
\makecell[tl]{Survival (nonpartic.) $\rightarrow$ \\ Self-Expression (active)\\ mixed (passive)} \\

\addlinespace
\texttt{national pride} &
Identification with nation and pride in nationality. &
\makecell[tl]{\texttt{very proud} (4)\\
\texttt{somewhat proud} (3)\\
\texttt{not very proud} (2)\\
\texttt{not proud at all} (1)} &
\makecell[tl]{Traditional (higher) $\rightarrow$ \\  Secular (lower)}\\

\addlinespace
\texttt{happiness} &
Subjective well-being / self-reported happiness. &
\makecell[tl]{\texttt{very happy} (4)\\
\texttt{rather happy} (3)\\
\texttt{not very happy} (2)\\
\texttt{not at all happy} (1)} &
\makecell[tl]{Survival (lower) $\rightarrow$ \\Self-Expression (higher)}\\

\addlinespace
\texttt{gender equality} &
Attitudes toward gender roles in education, jobs, and politics. &
\makecell[tl]{\texttt{egalitarian}\\
(disagrees that men are better leaders /\\
deserve jobs more)\\
\texttt{moderate} (neutral)\\
\texttt{traditional}\\
(agrees that men should have priority)} &
\makecell[tl]{Survival / Traditional \\ (tradit.) 
$\rightarrow$ Self-Expression / \\Secular (egalit.)}\\

\addlinespace
\texttt{materialism orientation} &
Materialist vs.\ post-materialist priorities (Inglehart battery). &
\makecell[tl]{\texttt{materialist}\\
(prioritizes economic and physical \\security)\\
\texttt{mixed} (balanced concerns)\\
\texttt{postmaterialist}\\
(prioritizes freedom, self-expression,\\
participation)} &
\makecell[tl]{Survival (material.) $\rightarrow$ \\Self-Expression (post\\material.)}\\

\addlinespace
\texttt{tolerance diversity} &
Attitudes toward outgroups (minorities, immigrants, different lifestyles). &
\makecell[tl]{\texttt{high tolerance}\\
(accepts immigrants, LGBTQ+, other\\ religions)\\
\texttt{moderate tolerance}\\
(accepts some, not others)\\
\texttt{low tolerance}\\
(rejects most outgroups)} &
\makecell[tl]{Survival (low) $\rightarrow$ \\ Self-Expression (high)}\\

\bottomrule
\end{tabular}
\caption{Overview of our defined 10 cultural variables: conceptual definition, admissible categories, and interpretation of the ordinal values in terms of Inglehart--Welzel axes. }
\label{tab:wvs-conditioning}
\end{table*}

\section{Statistics of the Generated Culturally-grounded Personas}
\label{app:demographics}

We summarize the generated personas across lexical and demographic dimensions. Table~\ref{tab:persona-lexical-stats} reports length statistics for bios and cultural mappings, while Tables~\ref{tab:persona-gender} and~\ref{tab:persona-age} show the distributions of gender and age bands extracted from metadata.

\begin{table}[H]
\centering
\small
\begin{tabular}{lrrrr}
\toprule
\textbf{Metric} & \textbf{Avg}. & \textbf{Std}. & \textbf{Min} & \textbf{Max} \\
\midrule
Chars  & 3271.7 & 329.6 & 74 & 6864 \\
Words  & 421.4  & 46    & 9  & 884 \\
Tokens & 657.3  & 61.2  & 24 & 1392 \\
\bottomrule
\end{tabular}
\caption{Lexical statistics of the \textit{short bio} and \textit{cultural variable mapping} texts from the generated personas.}
\label{tab:persona-lexical-stats}
\end{table}

\begin{table}[H]
\centering
\small
\begin{tabular}{l r}
\toprule
\textbf{Gender} & \textbf{\%} \\
\midrule
Woman & 47.984 \\
Man & 48.964 \\
Non-binary / gender-neutral & 1.631 \\
Agender & 0.002 \\
Genderqueer & 0.001 \\
Unspecified / prefer not to say / Other & 1.418 \\
\bottomrule
\end{tabular}
\caption{Distribution of gender labels extracted from the metadata of the generated personas.}
\label{tab:persona-gender}
\end{table}

\begin{table}[H]
\centering
\small
\begin{tabular}{l r}
\toprule
\textbf{Age band} & \textbf{\%} \\
\midrule
20--29 & 0.209 \\
30--39 & 67.743 \\
40--49 & 31.003 \\
50--59 & 1.014 \\
60--69 & 0.030 \\
\bottomrule
\end{tabular}
\caption{Distribution of age bands extracted from the metadata of the generated personas.}
\label{tab:persona-age}
\end{table}

\begin{table*}[h]
\centering
\small
\setlength{\tabcolsep}{6pt}
\begin{tabular}{@{}p{0.40\textwidth} r p{0.40\textwidth} r@{}}
\toprule
\textbf{Occupation macro} & \textbf{\%} & \textbf{Occupation macro} & \textbf{\%} \\
\midrule
Education \& Academia & 21.37 & Arts / Media / Design & 3.10 \\
Social Services \& Nonprofit & 13.28 & Finance / Banking / Insurance & 2.04 \\
Sales \& Marketing & 12.69 & Consulting / Professional Services & 1.88 \\
Tech / IT / Data & 8.63 & Agriculture \& Natural Resources & 1.79 \\
Management / Operations / Admin & 5.99 & Other / Unclear & 1.74 \\
Retail / Hospitality / Small Business & 5.95 & Healthcare \& Medicine & 1.60 \\
Manufacturing \& Skilled Trades & 5.86 & Transportation \& Logistics & 1.43 \\
Government \& Public Sector & 4.14 & Religious / Clergy & 0.81 \\
Construction / Real Estate / Planning & 4.05 & Not in labor force & 0.17 \\
Engineering (non-IT) & 3.32 & Law \& Legal & 0.15 \\
\bottomrule
\end{tabular}
\caption{Distribution of occupations after mapping free-form model generations to 20 macro-categories.}
\label{tab:persona-occupations}
\end{table*}

\begin{figure*}[h]
    \centering
    \includegraphics[width=1\linewidth]{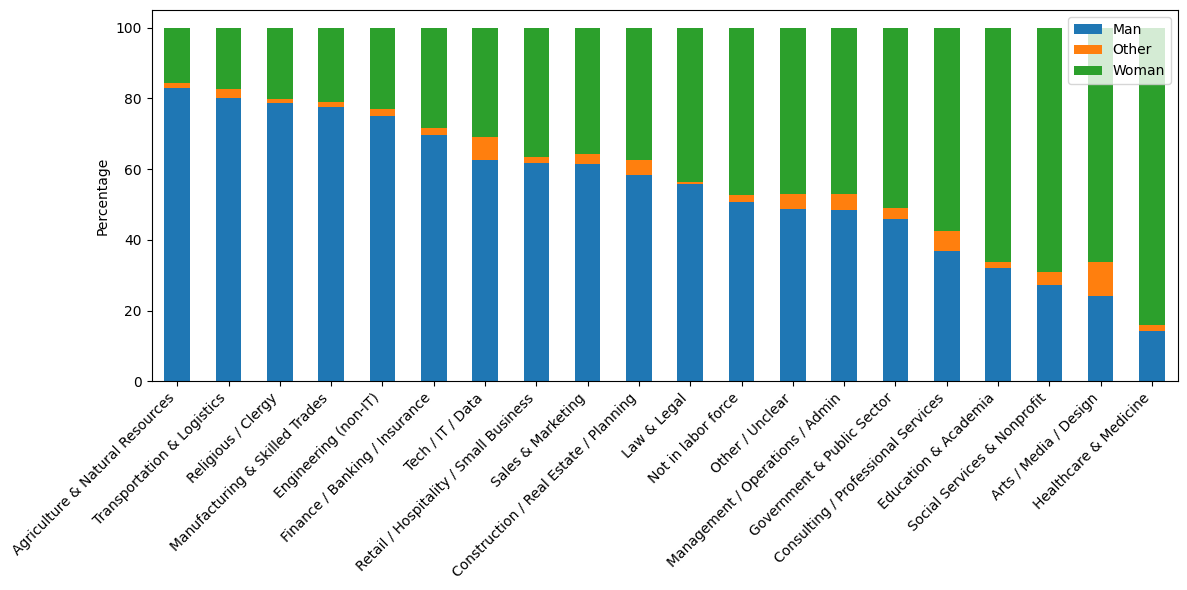}
    \caption{\color{black}Gender composition across occupation macro-categories in the generated personas. Each bar corresponds to an occupation macro-category and is normalized to 100\%; segments show the percentage of personas labeled as Man, Woman, or Other (where Other aggregates all gender labels different from Man/Woman). Categories are ordered by the share of Man.}
  \label{fig:gender}
\end{figure*}

\begin{table*}[ht!]
\centering
\setlength{\tabcolsep}{4pt}
\scalebox{0.9}{
\begin{tabular}{lr lr}
\toprule
\textbf{Country} & \% \textbf{personas} & \textbf{Country} & \% \textbf{personas} \\
\midrule
United States & 17.53 & Kazakhstan & 0.29 \\
Poland & 8.65 & Ireland & 0.27 \\
Mexico & 6.13 & France & 0.27 \\
Germany & 5.47 & Kenya & 0.26 \\
Netherlands & 4.39 & Guatemala & 0.25 \\
Philippines & 3.86 & China & 0.23 \\
Brazil & 3.71 & Vietnam & 0.22 \\
Sweden & 3.54 & New Zealand & 0.19 \\
India & 3.34 & Ghana & 0.17 \\
Canada & 2.69 & Ecuador & 0.16 \\
Spain & 2.28 & Central Europe & 0.16 \\
Italy & 2.12 & Iran & 0.15 \\
Turkey & 1.92 & Costa Rica & 0.14 \\
Czechia & 1.85 & Bangladesh & 0.14 \\
United Kingdom & 1.68 & Belarus & 0.14 \\
Saudi Arabia & 1.62 & Bosnia and Herzegovina & 0.14 \\
Pakistan & 1.4 & Croatia & 0.13 \\
Portugal & 1.3 & Middle East & 0.12 \\
Russia & 1.25 & Slovenia & 0.1 \\
Indonesia & 1.14 & Lebanon & 0.1 \\
United Arab Emirates & 1.11 & Tunisia & 0.09 \\
Finland & 1.02 & Thailand & 0.08 \\
Norway & 1.01 & Qatar & 0.07 \\
Jordan & 0.99 & Uruguay & 0.07 \\
Hungary & 0.93 & Ethiopia & 0.07 \\
Argentina & 0.84 & Moldova & 0.06 \\
South Korea & 0.76 & Central America & 0.06 \\
Egypt & 0.76 & Oman & 0.06 \\
Denmark & 0.76 & Lithuania & 0.06 \\
Austria & 0.75 & Belgium & 0.05 \\
Japan & 0.69 & Kuwait & 0.05 \\
Switzerland & 0.55 & Kyrgyzstan & 0.05 \\
Malaysia & 0.54 & Georgia & 0.05 \\
Serbia & 0.52 & Uzbekistan & 0.05 \\
Chile & 0.52 & Uganda & 0.04 \\
Romania & 0.51 & Bolivia & 0.04 \\
Israel & 0.51 & Europe & 0.04 \\
Colombia & 0.49 & Iraq & 0.04 \\
Nigeria & 0.48 & Indiana & 0.04 \\
Bulgaria & 0.43 & Colorado & 0.04 \\
Australia & 0.4 & Latvia & 0.03 \\
Estonia & 0.4 & Iceland & 0.03 \\
Singapore & 0.4 & Dominican Republic & 0.03 \\
Czech Republic & 0.37 & El Salvador & 0.03 \\
Morocco & 0.37 & Nicaragua & 0.03 \\
South Africa & 0.35 & Palestine & 0.03 \\
Slovakia & 0.33 & England & 0.03 \\
Peru & 0.31 & Puerto Rico & 0.03 \\
Greece & 0.3 & North Macedonia & 0.03 \\
Ukraine & 0.29 & Nepa & 0.02 \\
\bottomrule
\end{tabular}
}
\caption{Percentage of personas across the first 100 countries by count.}
\label{tab:countries}
\end{table*}

\newpage
\ 

\newpage 
\clearpage
\section{Prompts}\label{app:prompts}

\noindent
\begin{minipage}{\textwidth}
    \centering
    \small
    \begin{tcolorbox}[
        width=\textwidth,
        colback=blue!5!white,
        colframe=blue!75!black,
        title=Persona Generation Prompt Template
    ]
    \textbf{Conditioning:} \\
    $\{$\texttt{cultural configuration}$\}$

    \vspace{0.3em}
    \textbf{Task:} \\
    \textit{
    You are asked to generate a single detailed \textbf{persona profile} that is consistent with the conditioning above. 
    The persona must be coherent, realistic, and explicitly tie behaviors, life choices, and attitudes back to each of the 
    ten cultural variables listed. Use a professional-but-readable tone.
    }

    \vspace{0.3em}
    \textbf{Deliverables (strict order):}
    \begin{itemize}[noitemsep, topsep=2pt]
        \item \textbf{Profile metadata}: name, age, gender (optional), occupation, country/region (plausible given the cultural profile).
        \item \textbf{Short bio}: 2--4 sentences describing life situation and background.
        \item \textbf{Cultural variable mapping}: for each of the ten variables, 1--2 sentences explaining how the variable 
        manifests in attitudes and daily behavior (numbered list matching variable names).
    \end{itemize}

    \textbf{Constraints:}
    \begin{itemize}[noitemsep, topsep=2pt]
        \item Each sentence in the cultural mapping must explicitly reference the corresponding conditioning variable.
        \item No contradiction of the conditioning is allowed (e.g., \texttt{never\_justifiable} must be reflected).
        \item The persona must remain within plausible real-world bounds.
        \item Output length must be between approximately 250 and 500 words.
    \end{itemize}

    \end{tcolorbox}
    \captionof{figure}{{\color{black} Prompt template  for the generation of culturally-grounded personas. The conditioning parameter is a cultural configuration, i.e., a list of \textit{(cultural variable, admissible value)} pairs that corresponds to one    of the 93,312 possible combinations of our defined cultural variables.} }
    \label{fig:persona-prompt}


    \begin{tcolorbox}[
        width=\textwidth,
        colback=blue!5!white,
        colframe=blue!75!black,
        title=IW Indicator Elicitation Prompt
    ]
    \textsc{System message:} \\
    \textit{
    You are given a detailed persona profile describing a person's background, attitudes, and values.
    Assume the role of this person when answering the following questions.
    }

    \vspace{0.7em}
    \textsc{User message:} \\
    \textbf{Task:} \\
    \textit{
    Answer the following questions as this person would, selecting the option that best reflects their
    beliefs and attitudes. Each question corresponds to a cultural value indicator used in the
    Inglehart--Welzel cultural map.
    }

    \vspace{0.3em}
    \textbf{Questions:} \\
    $\{$\texttt{list of WVS-style questions corresponding to the 10 IW indicators}$\}$\\ 

    \vspace{0.3em}
    \textbf{Output format:} \\
    \textit{
    Provide one answer per question, following the original WVS response scales (e.g., binary choices,
    Likert-type scales, or 1--10 justifiability ratings). Do not include explanations or additional text.
    }
    \end{tcolorbox}
    \captionof{figure}{Prompt template used to obtain persona responses for computing the Inglehart--Welzel cultural map indicators.
    Each persona is queried independently using this template.}
    \label{fig:iw-prompt}

\end{minipage}

\clearpage

\noindent
\begin{minipage}{\textwidth}
\centering
    \small
    \begin{tcolorbox}[
        width=\textwidth,
        colback=blue!5!white,colframe=blue!75!black, title=WVB-Probe Prompting]
    \textsc{System message:}\\
    \textit{
    You are the following person:}\\
    $\{$\texttt{persona profile}$\}$\\
    
    \textit{Answer the following survey questions as this person would answer, based on their values and worldview.\\
    It is very important to respond EXACTLY as requested. Be terse.
    }\\[0.6em]

    \textsc{User message:}\\
    \textit{
    Before answering, report your demographics by selecting exactly one option from each list:\\
    -- Continent: [Africa, Asia, Europe, North America, South America, Oceania]\\
    -- Residential area: [Urban, Rural]\\
    -- Education: [Primary or No Education, Lower Secondary, Upper to Post Secondary, Tertiary]\\
    Return the three fields in a structured form (one line each).
    }\\[0.6em]

    \textsc{User message (repeated for each of the 36 WVB-Probe items):}\\
    \textit{
    Question:} $\{$\texttt{WVB-Probe question prompt}$\}$\\
    
    \textit{Return only a single integer on the response scale specified in the question prompt.
    }
    \end{tcolorbox}
    \captionof{figure}{WVB-Probe prompt template. The LLM 
    is conditioned on the full generated persona profile. It is first prompted to output a demographic triple compatible with WVB. It is then prompted individually for each of the 36 WVB-Probe value questions, producing a single integer response for each question.}
    \label{fig:wvb-probe-prompt}


    \begin{tcolorbox}[
        width=\textwidth,
        colback=blue!5!white,colframe=blue!75!black, title=MFQ-2 Prompting]
    \textsc{System message:}\\
    \textit{
    You are the following person:}\\
    $\{$\texttt{persona profile}$\}$\\
    
    \textit{Answer the following MFQ-2 survey as this person would answer, based on their values and worldview.\\
    It is very important to respond EXACTLY as requested. Be terse.
    }\\[0.6em]

    \textsc{User message:}\\
    \textit{
    You will be given 36 numbered MFQ-2 items. For each item, return a single integer in $\{1,2,3,4,5\}$.\\
    Return \textbf{ONLY} valid JSON in the following format (and nothing else):}\\
    \texttt{\{"answers": \{"1": <int>, ..., "36": <int>\}\}}\\
    $\{$\texttt{MFQ-2 items list (1..36)}$\}$
    \end{tcolorbox}
    \captionof{figure}{MFQ-2 prompt template. The LLM 
    is conditioned on the full generated-persona profile and prompted to answer all 36 MFQ-2 items at once, returning a single integer per item in a JSON object keyed by item index.}
    \label{fig:mfq2-prompt}
\end{minipage}

\clearpage

\section{Inglehart--Welzel Cultural Map Reference}
\label{app:iw-map}

Figure \ref{fig:iw-countries}  show the Inglehart--Welzel cultural map with country positions reconstructed from IVS data and position of the non-conditioned LLM (\textsf{gpt-oss-20B}).

\section{Personas on Inglehart-Welzel map and closed frequent itemsets}
\label{app:iw-patterns}

\begin{figure*}[ht!]
    \centering
    \includegraphics[width=0.78\textwidth]{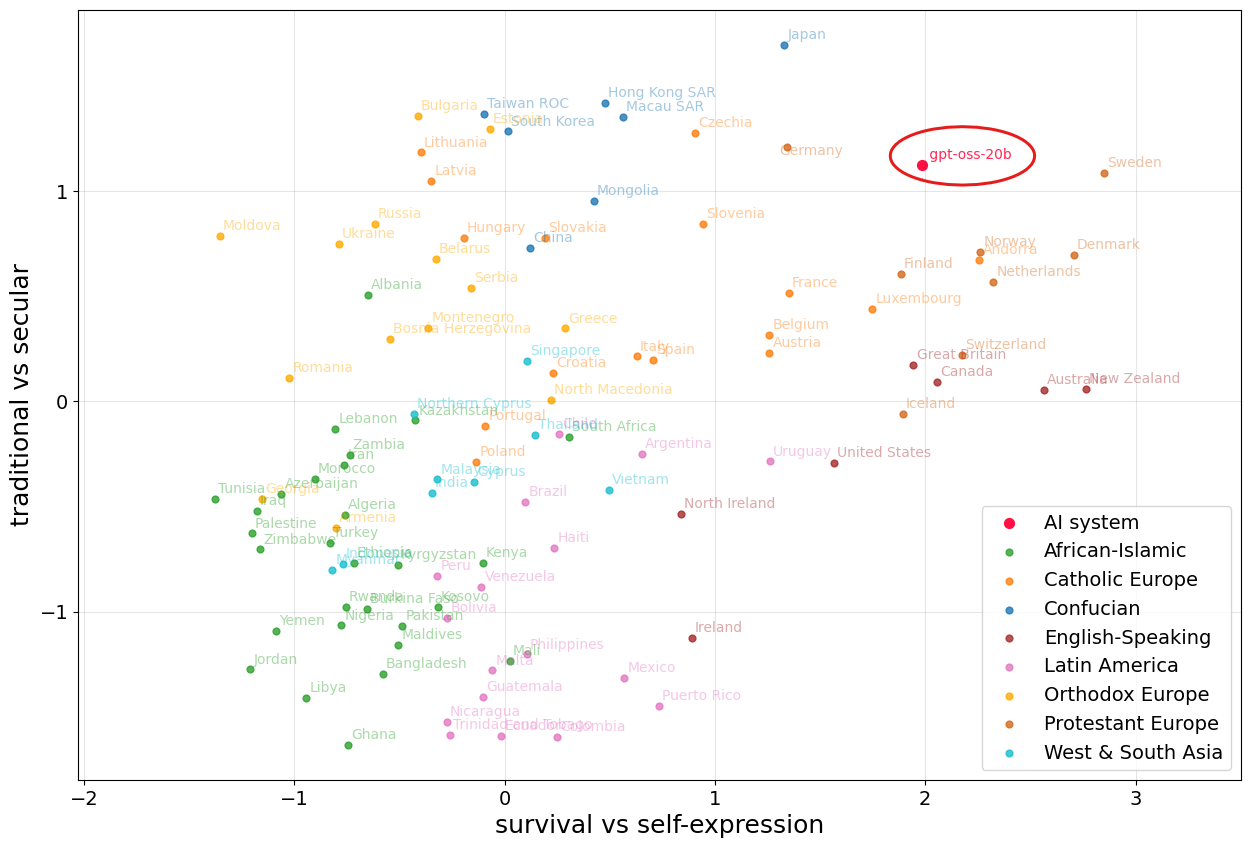}
    \caption{Inglehart--Welzel cultural map with country positions reconstructed from IVS data and position of the non-conditioned LLM (\textsf{gpt-oss-20B}).}
  \label{fig:iw-countries}
\end{figure*}

\begin{figure*} [ht!]
    \centering
    \includegraphics[width=0.65\linewidth]{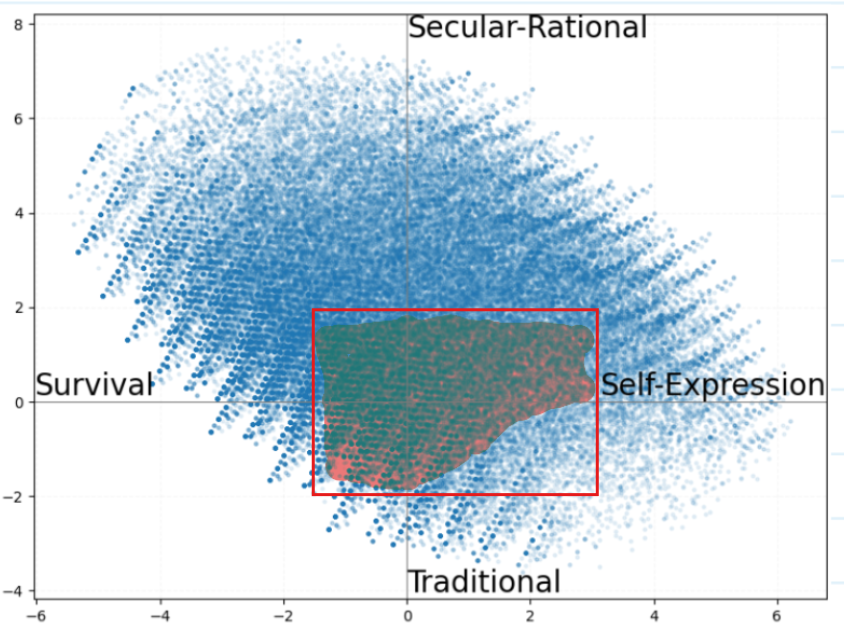}
    \caption{Overlay of our generated   personas (blue dots) and country positions reconstructed from IVS data on the Inglehart–Welzel cultural map. The rectangle marks the value space covered by real countries (i.e., Fig \ref{fig:iw-countries}).}
  \label{fig:iw-overlap}
\end{figure*}

Figure \ref{fig:iw-overlap}  shows the overlay of our generated   personas (blue dots) and country positions reconstructed from IVS data on the Inglehart–Welzel cultural map. The rectangle marks the value space covered by real countries (i.e., Figure \ref{fig:iw-countries}).

\begin{figure*}[ht!]
    \centering
    \begin{subfigure}{0.4\textwidth}
        \centering
        \includegraphics[width=\linewidth]{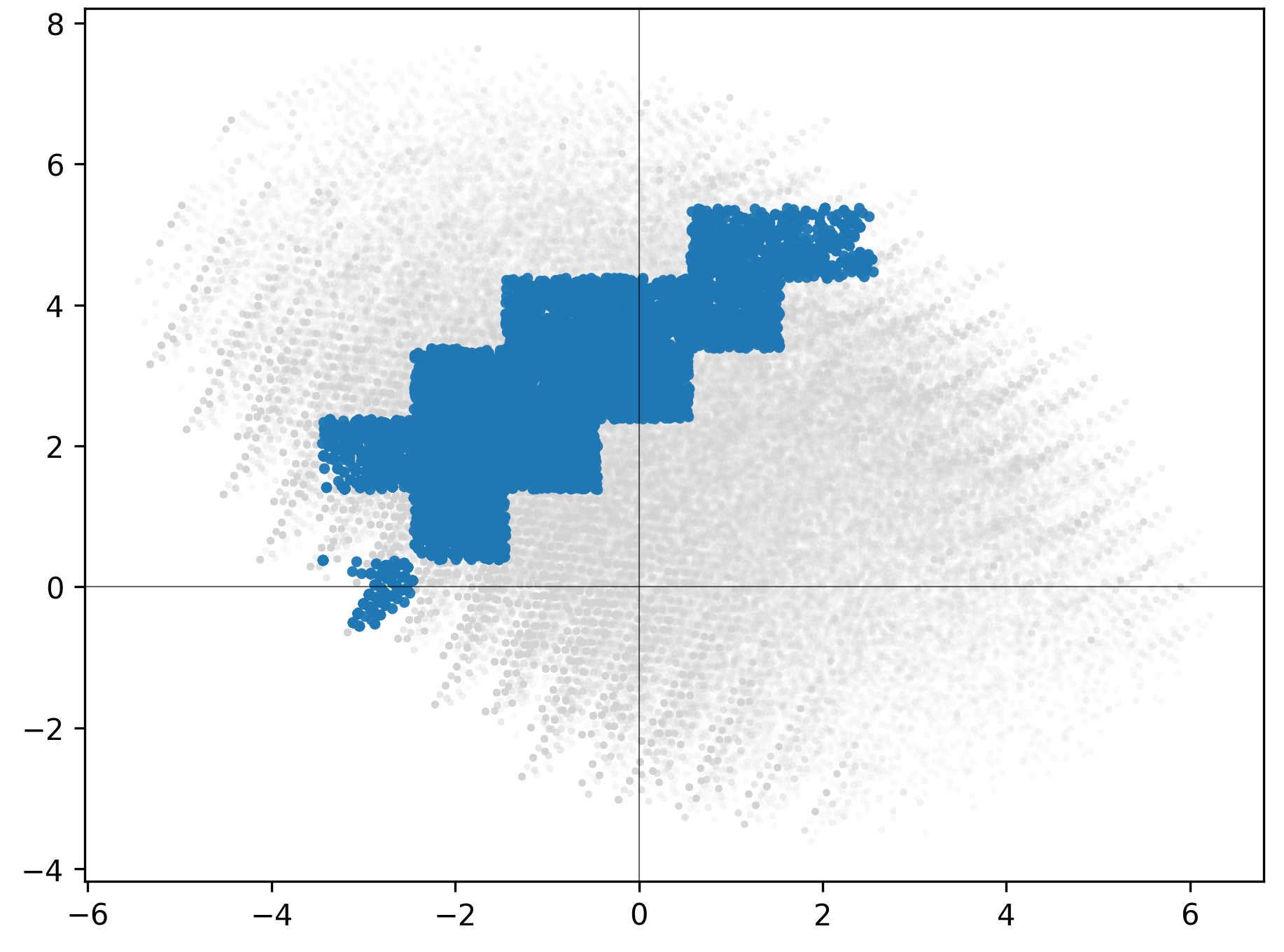}
        \caption{{happiness = not very happy}}
    \end{subfigure}
    \begin{subfigure}{0.4\textwidth}
        \centering
        \includegraphics[width=\linewidth]{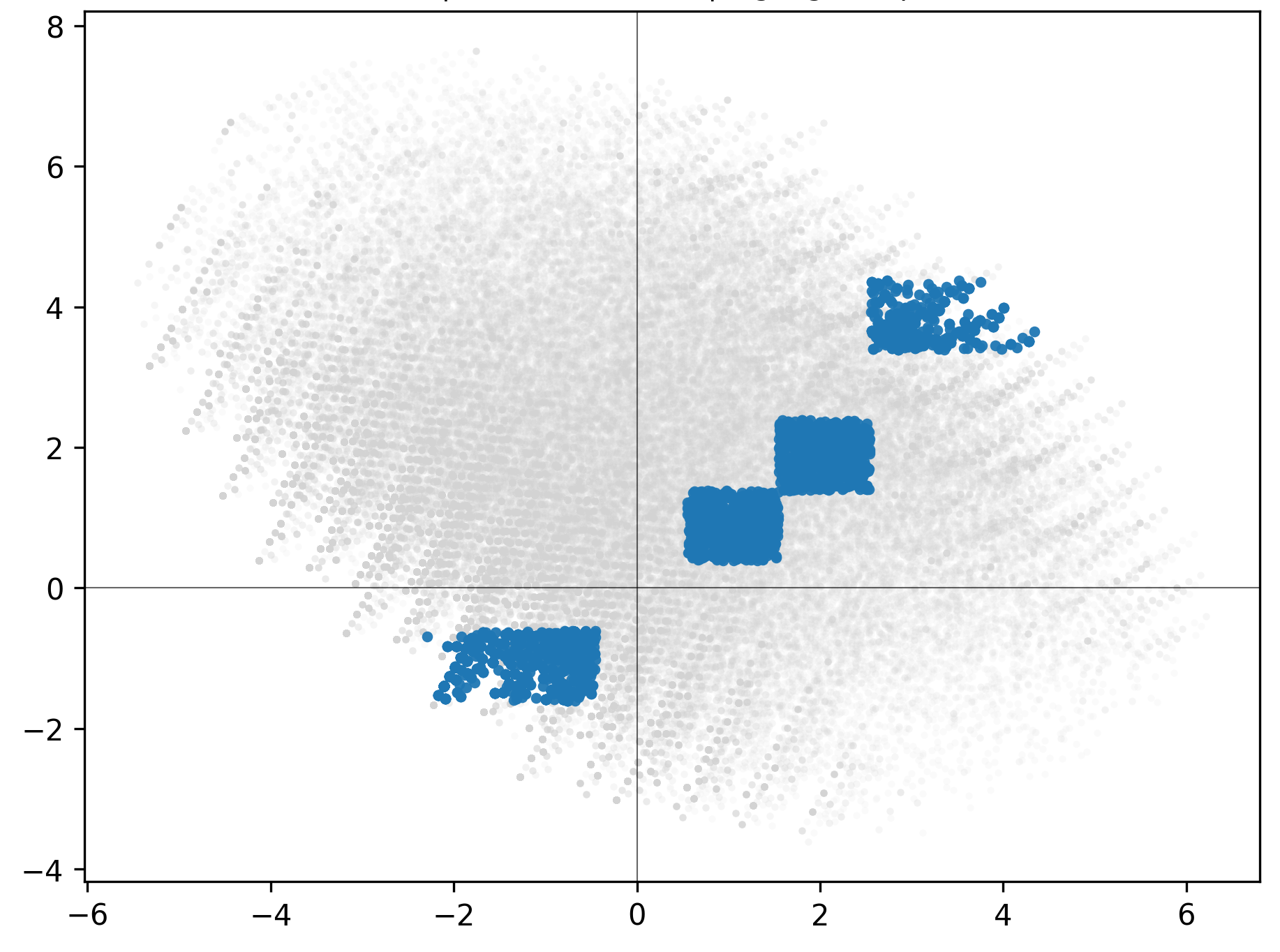}
        \caption{{happiness = rather happy}}
    \end{subfigure}
    \begin{subfigure}{0.4\textwidth}
        \centering
        \includegraphics[width=\linewidth]{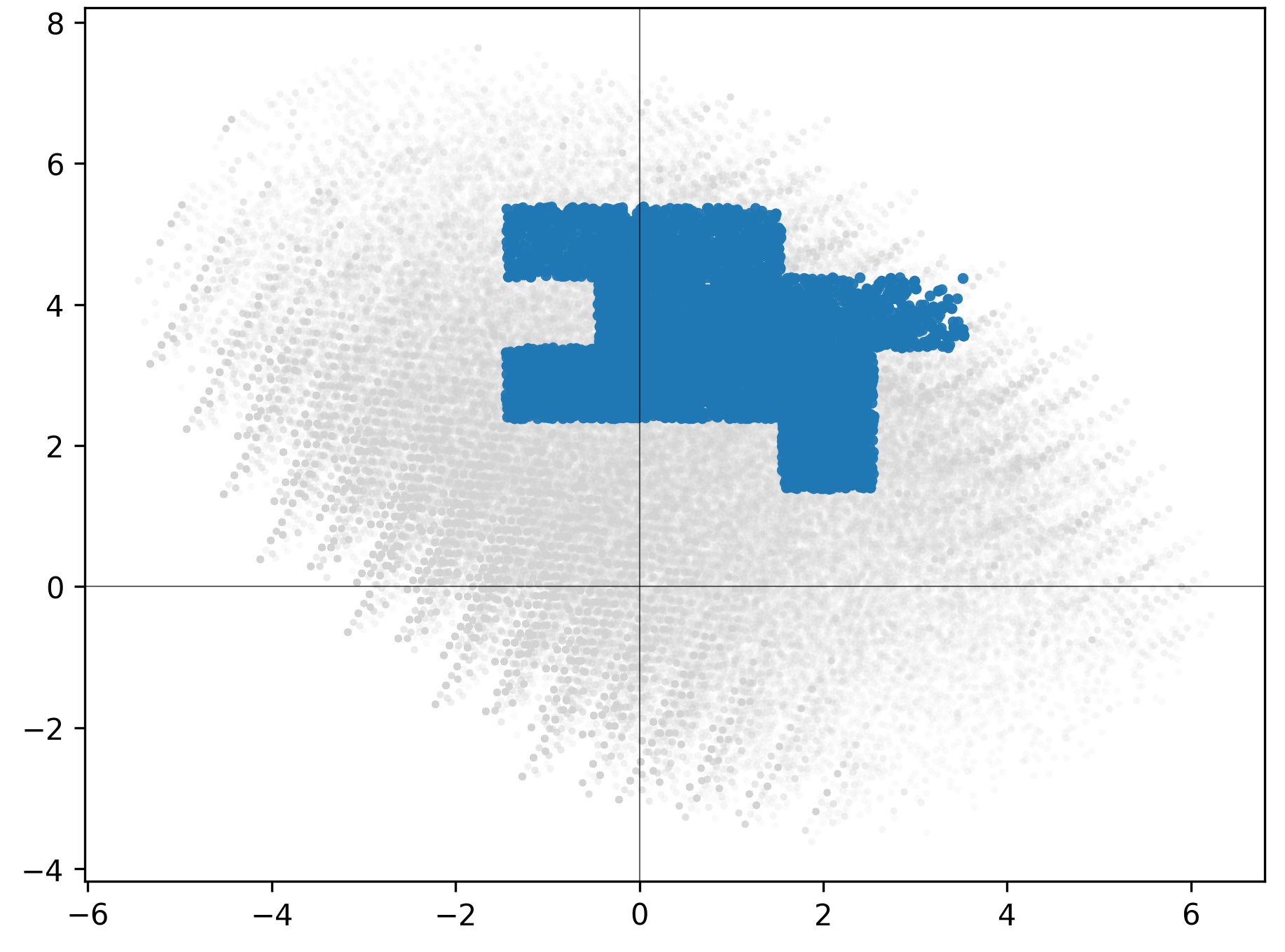}
        \caption{{child rearing = neutral}}
    \end{subfigure}
    \begin{subfigure}{0.4\textwidth}
        \centering
        \includegraphics[width=\linewidth]{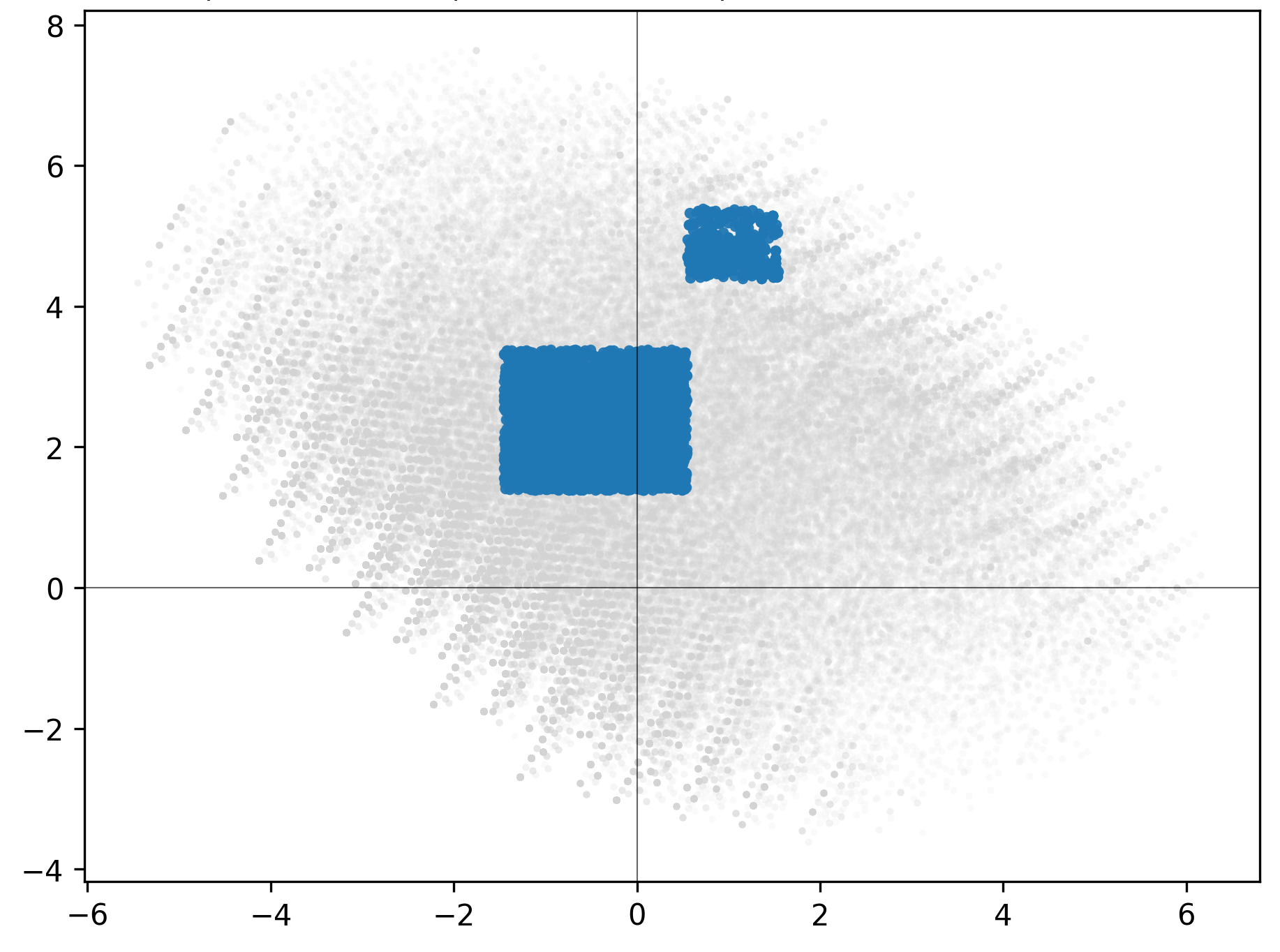}
        \caption{{moral acceptability = sometimes justificable}}
    \end{subfigure}
    \begin{subfigure}{0.4\textwidth}
        \centering
        \includegraphics[width=\linewidth]{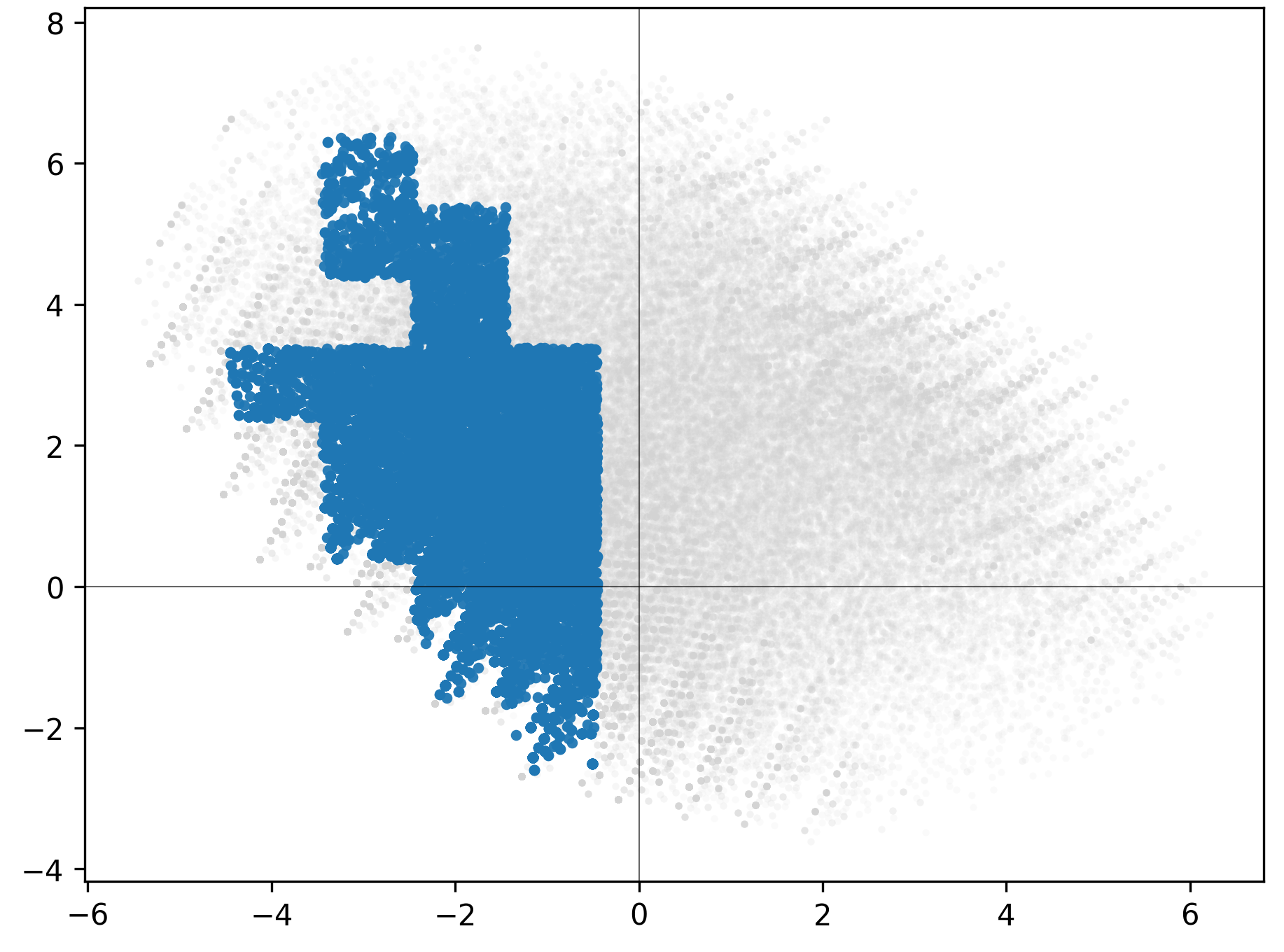}
        \caption{{tolerance diversity = low tolerance}}
    \end{subfigure}
    \begin{subfigure}{0.4\textwidth}
        \centering
        \includegraphics[width=\linewidth]{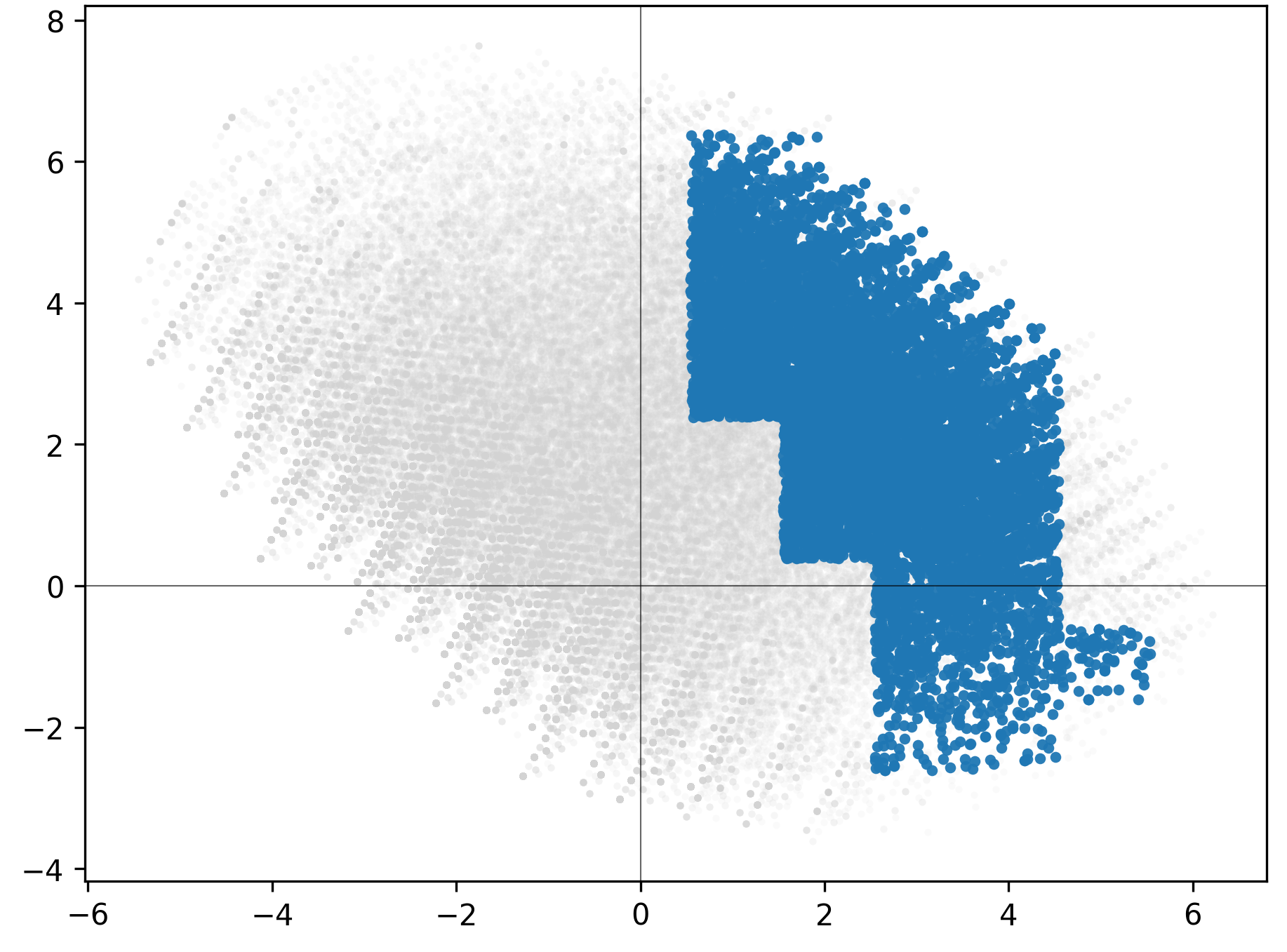}
        \caption{{tolerance diversity = high tolerance}}
    \end{subfigure}
    \begin{subfigure}{0.4\textwidth}
        \centering
        \includegraphics[width=\linewidth]{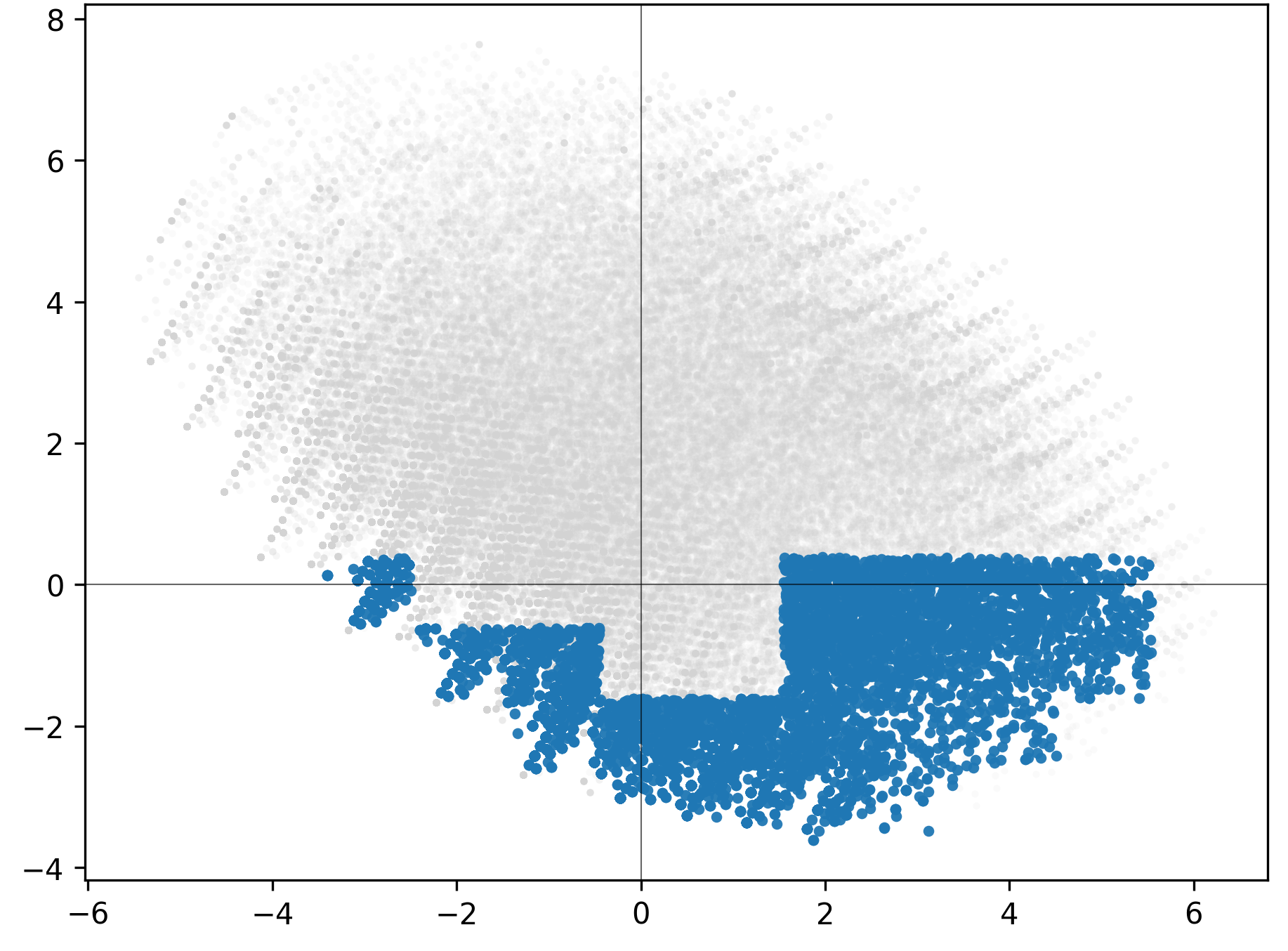}
        \caption{{national pride = very proud}}
    \end{subfigure}\
    \begin{subfigure}{0.4\textwidth}
        \centering
        \includegraphics[width=\linewidth]{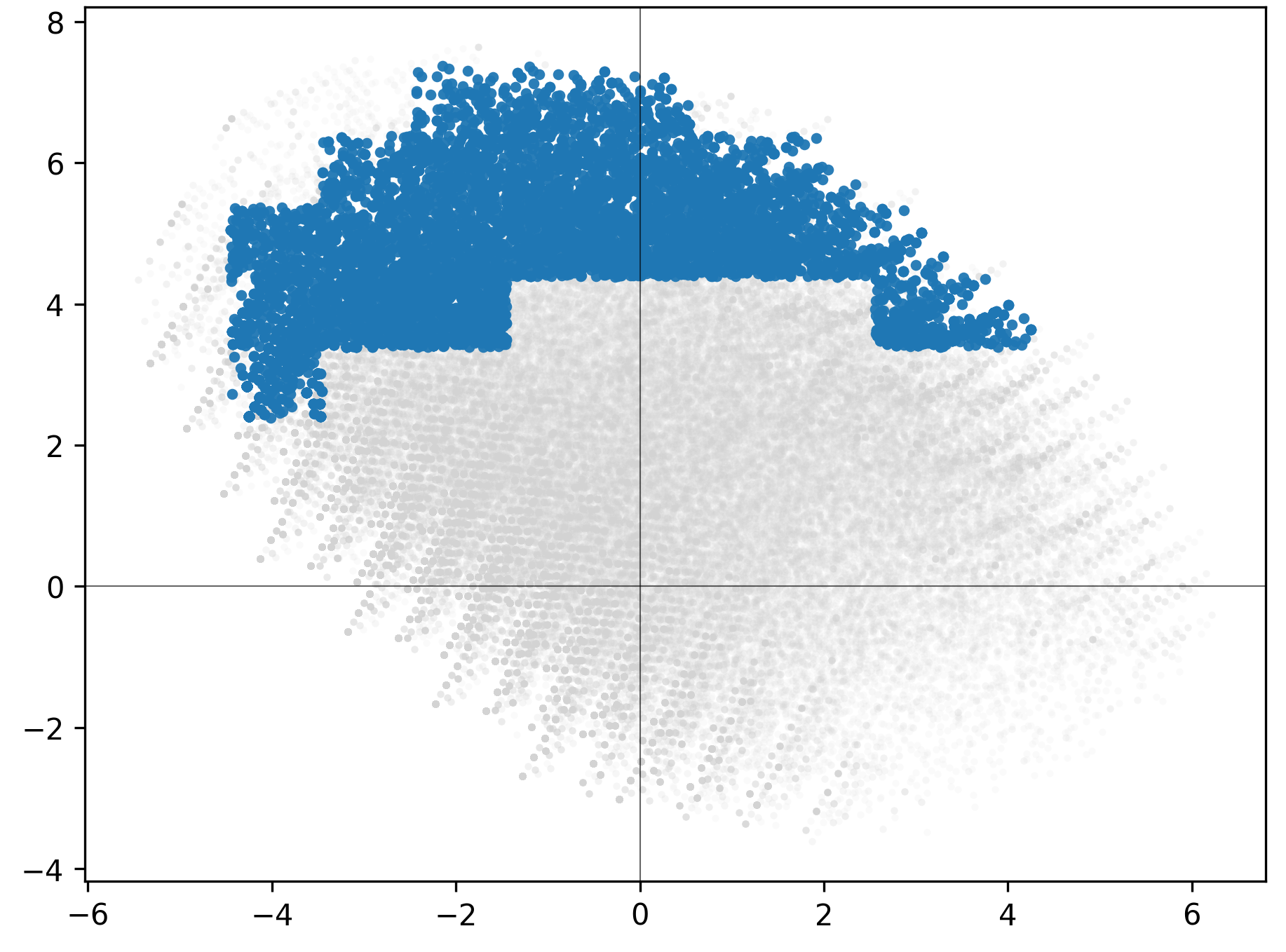}
        \caption{{national pride = not proud at all}}
    \end{subfigure}
\caption{A selection of singleton closed frequent itemsets, with mean minimum  support equal to 0.2 and $\rho=0.5$.
Panels (a–d) correspond to intermediate category levels (e.g., `not very happy', `rather happy', `neutral', `sometimes justifiable'), whose footprints predominantly occupy the high-density central region of the IW space.
Panels (e–h) show extreme category levels (e.g., `high' vs.\ `low tolerance', `very proud' vs.\ `not proud at all'), whose footprints extend toward the periphery and exhibit clear axis-aligned shifts.
In particular, extreme values follow the directional associations predicted by the variable-to-axis mapping  
(Table~\ref{tab:wvs-conditioning}), 
producing systematic displacements along the Survival--Self-expression (horizontal) and Traditional--Secular-rational (vertical) axes.}
\label{fig:iw-voronoi-singleton}

\end{figure*}

Figure \ref{fig:iw-voronoi-singleton} shows a selection of singleton closed frequent itemsets, with mean minimum  support equal to 0.2 and $\rho=0.5$.

\section{Demographic distribution of Personas for WVB-Probe}\label{app:wvb}

Figure \ref{fig:wvb-demotriad} shows distribution of demographic triples obtained during the WVB-Probe prompting. Table \ref{tab:wvb-probe-all-groups} reports the complete group-level WVB-Probe alignment for all 45 demographic triples supported by WVB-Probe, sorted by decreasing number of personas.

\section{Additional Analyses on Moral Foundation Scores}\label{app:mfq}

\begin{figure*}[ht!]
    \centering

    \begin{subfigure}{0.3\textwidth}
        \centering
        \includegraphics[width=0.85\linewidth]{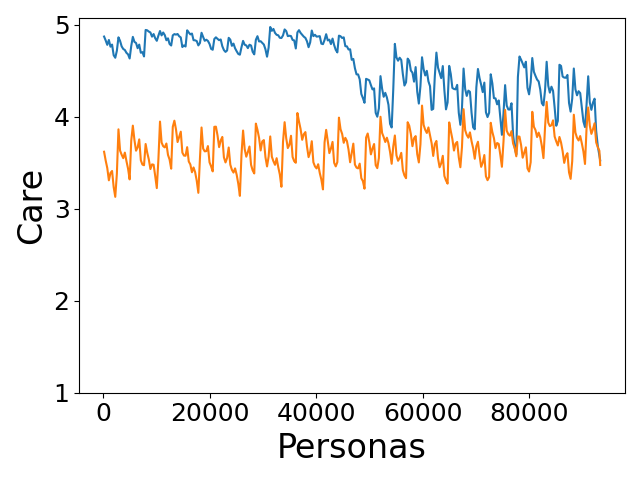}
        \caption{}
    \end{subfigure}\hfill
    \begin{subfigure}{0.3\textwidth}
        \centering
        \includegraphics[width=0.85\linewidth]{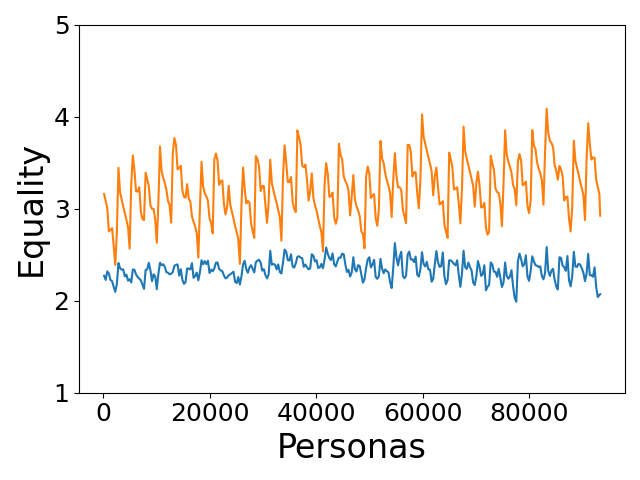}
        \caption{}
    \end{subfigure}\hfill
    \begin{subfigure}{0.3\textwidth}
        \centering
        \includegraphics[width=0.85\linewidth]{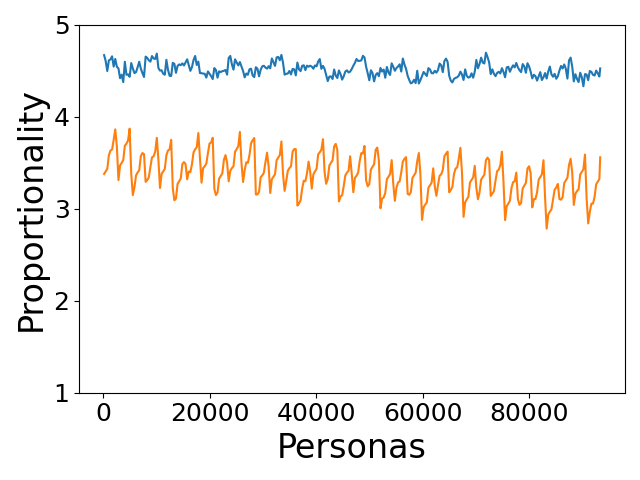}
        \caption{}
    \end{subfigure}


    \begin{subfigure}{0.3\textwidth}
        \centering
        \includegraphics[width=0.85\linewidth]{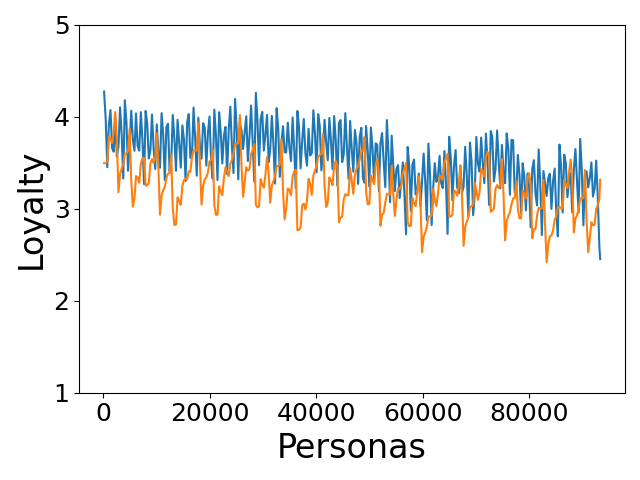}
        \caption{}
    \end{subfigure}\hfill
    \begin{subfigure}{0.3\textwidth}
        \centering
        \includegraphics[width=0.85\linewidth]{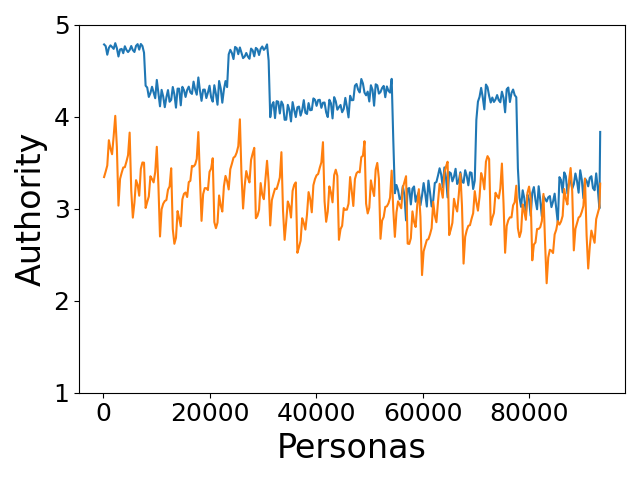}
        \caption{}
    \end{subfigure}\hfill
    \begin{subfigure}{0.3\textwidth}
        \centering
        \includegraphics[width=0.85\linewidth]{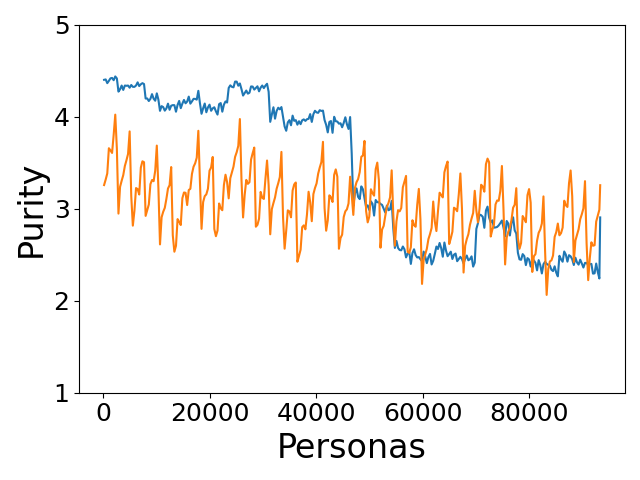}
        \caption{}
    \end{subfigure}

    \caption{Comparison between MFQ-2–based (blue) and mapping-based moral (orange) foundation scores. Mapping-based scores are obtained using all cultural variables. Personas are ordered along the x-axis; the y-axis reports foundation scores. To avoid cluttering, lines of both score types  are smoothed by averaging scores within consecutive bins of 300 personas (according to the ordering fixed through sorting  by cultural variable then by admissible values).}
    \label{fig:mfq2-all10}
\end{figure*}

\begin{figure*}[ht!]
    \centering

    \begin{subfigure}{0.3\textwidth}
        \centering
        \includegraphics[width=0.85\linewidth]{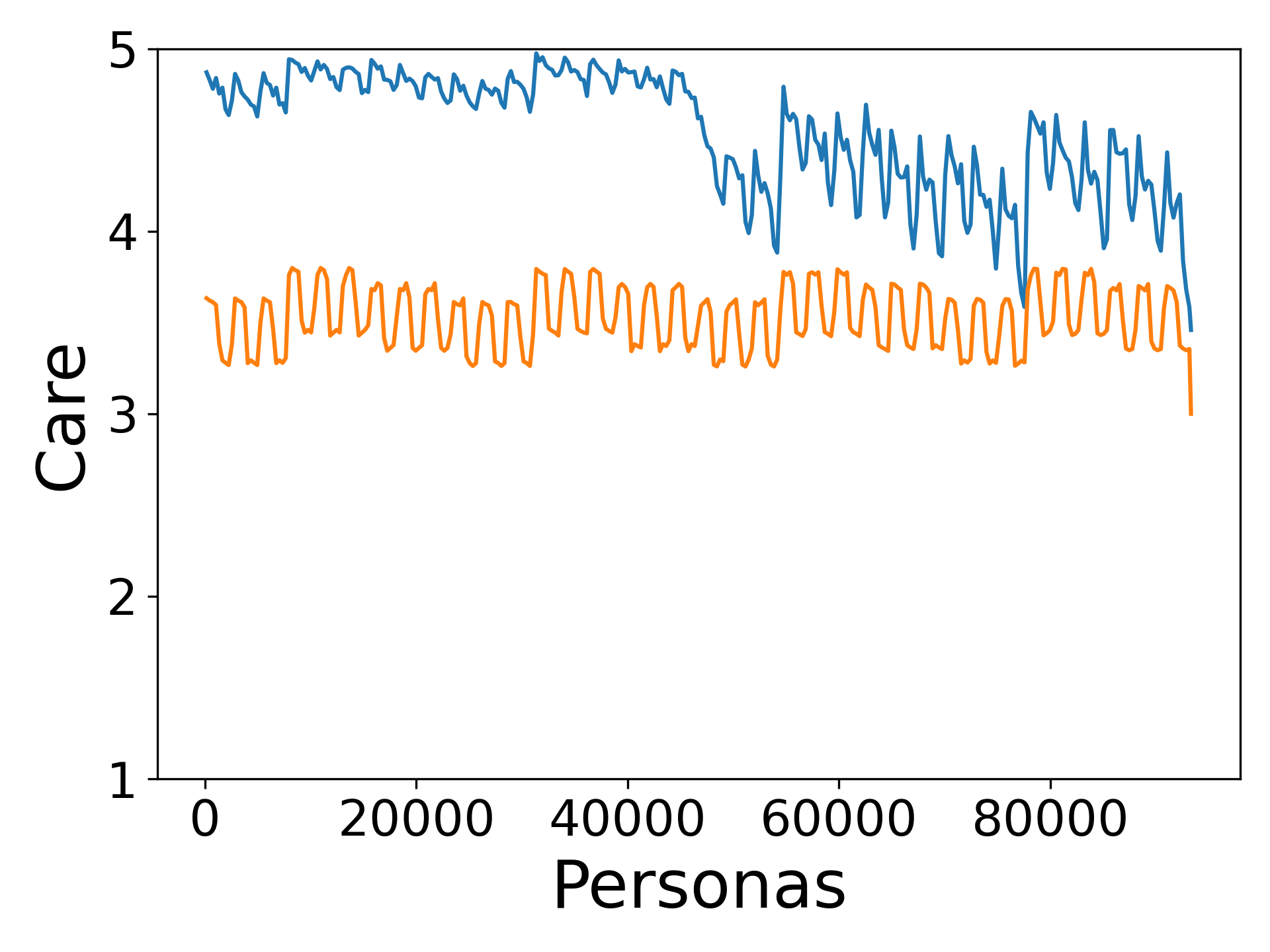}
        \caption{}
    \end{subfigure}\hfill
    \begin{subfigure}{0.3\textwidth}
        \centering
        \includegraphics[width=0.85\linewidth]{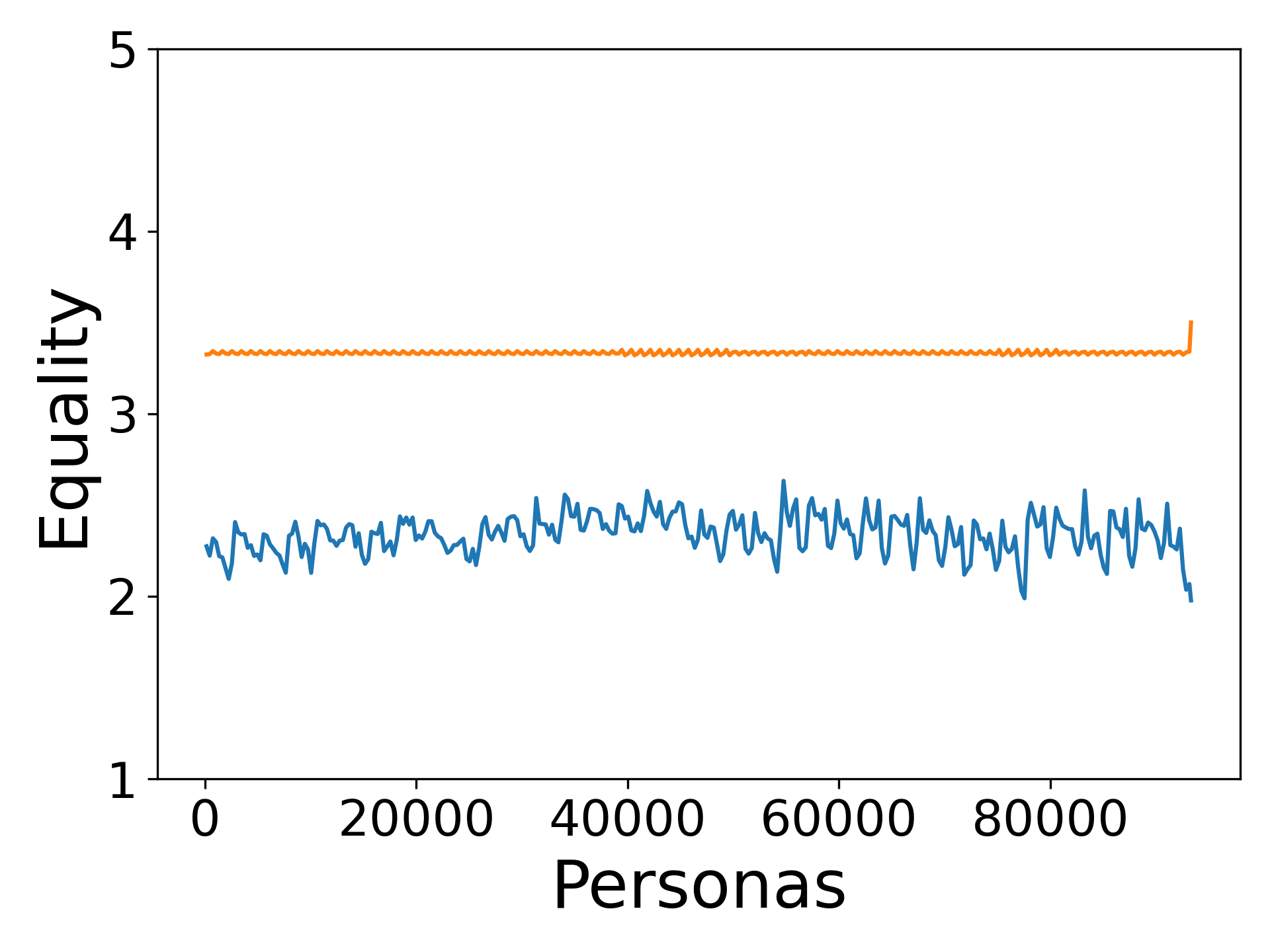}
        \caption{}
    \end{subfigure}\hfill
    \begin{subfigure}{0.3\textwidth}
        \centering
        \includegraphics[width=0.85\linewidth]{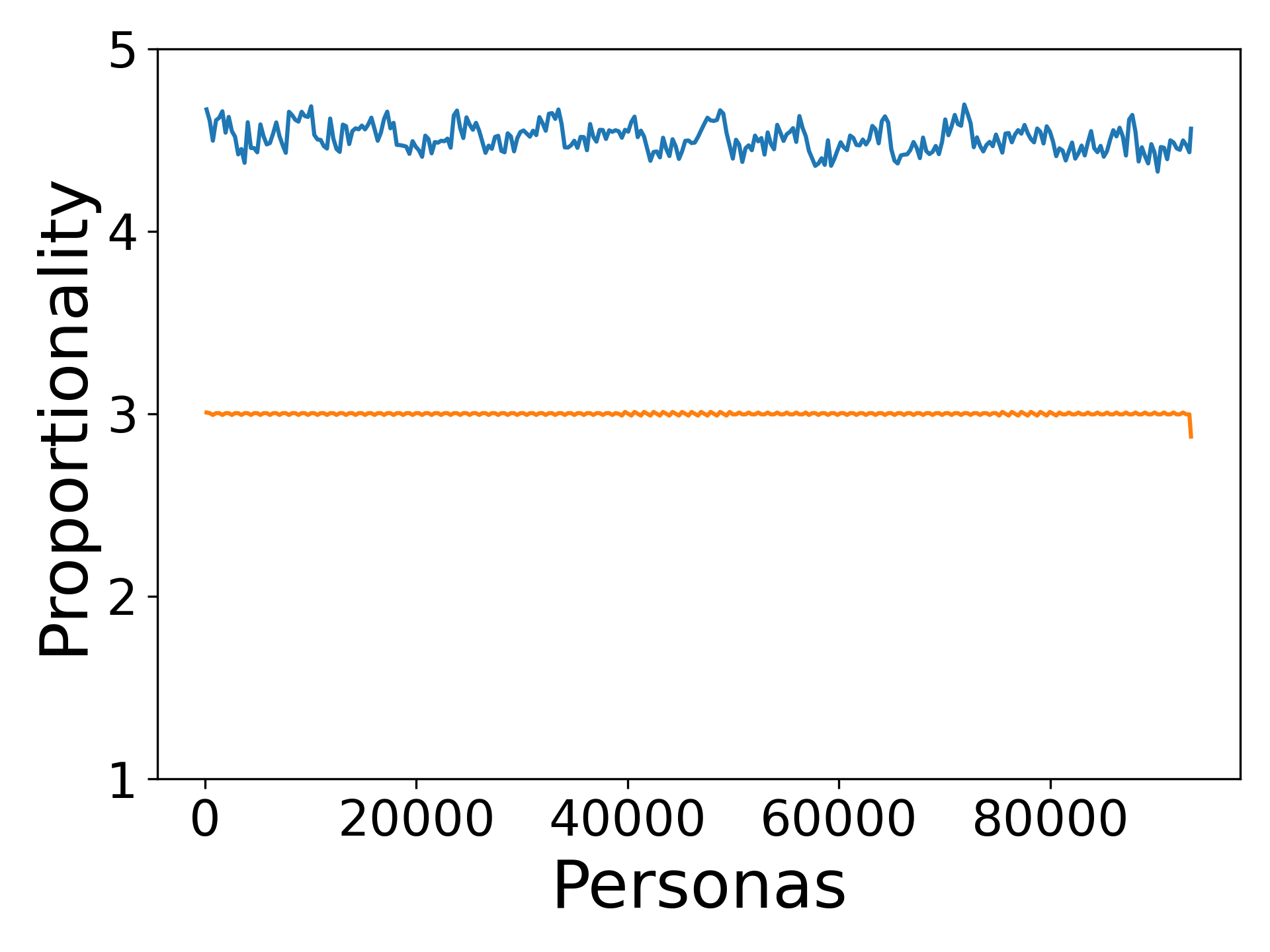}
        \caption{}
    \end{subfigure}


    \begin{subfigure}{0.3\textwidth}
        \centering
        \includegraphics[width=0.85\linewidth]{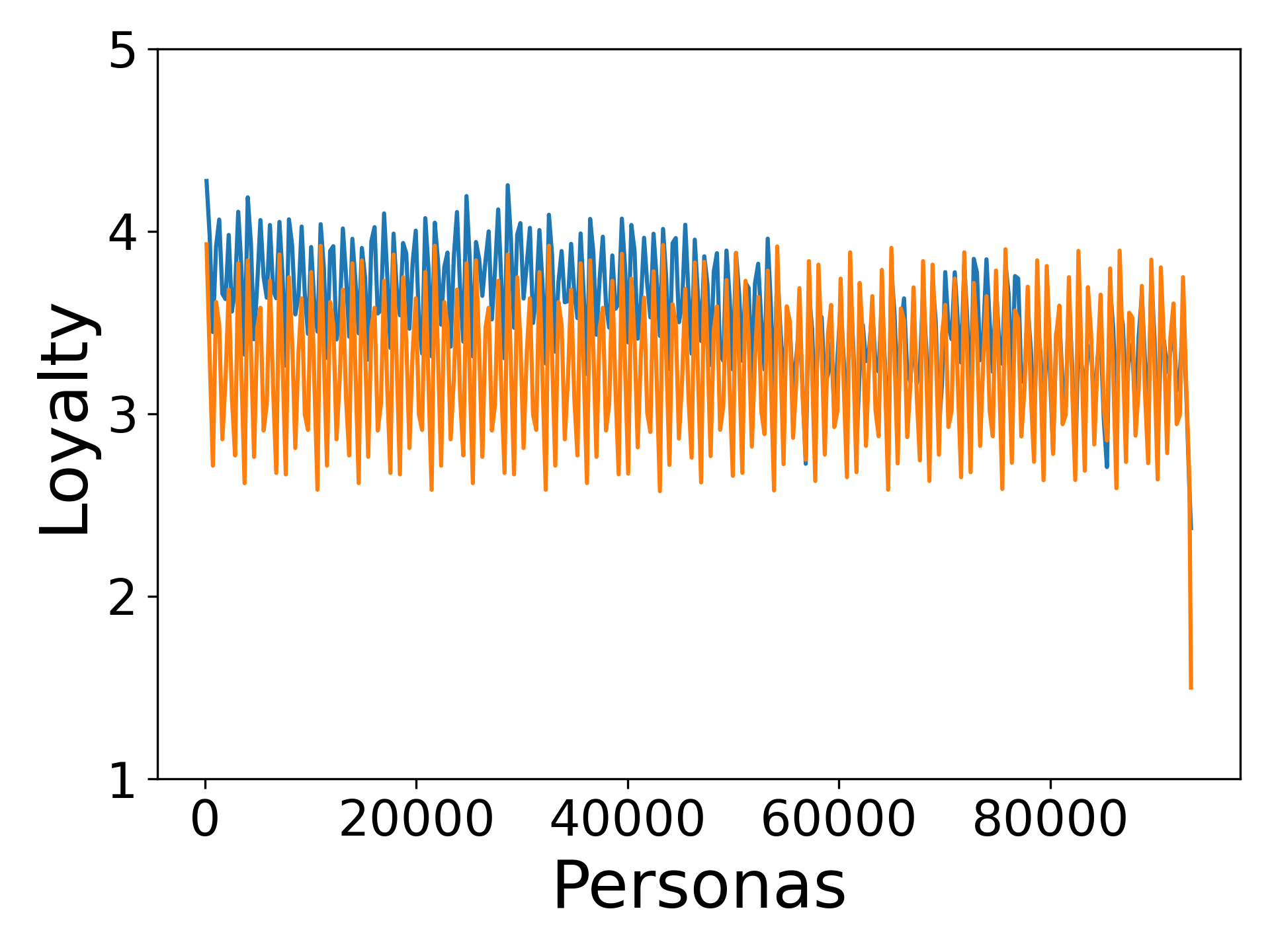}
        \caption{}
    \end{subfigure}\hfill
    \begin{subfigure}{0.3\textwidth}
        \centering
        \includegraphics[width=0.85\linewidth]{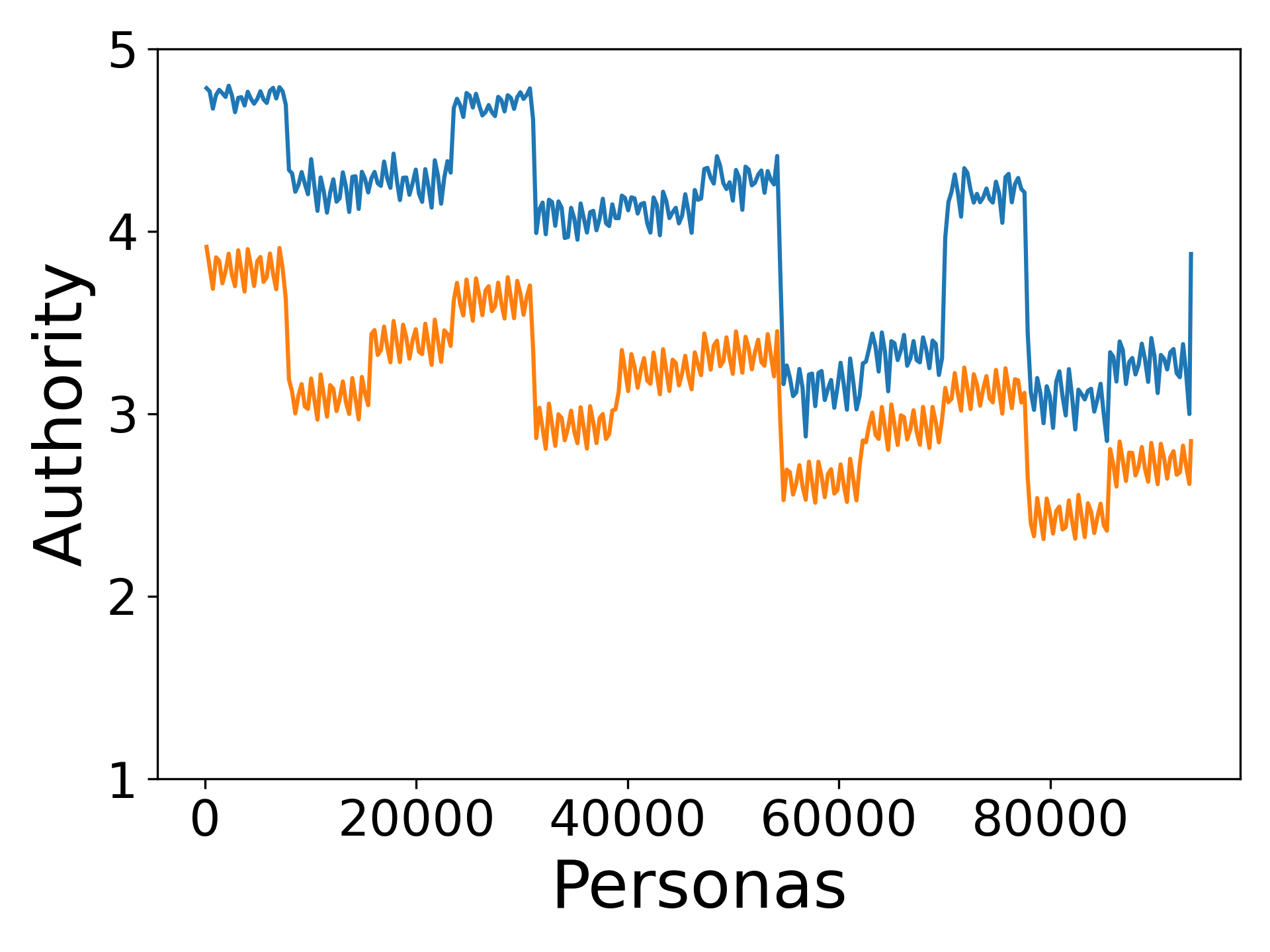}
        \caption{}
    \end{subfigure}\hfill
    \begin{subfigure}{0.3\textwidth}
        \centering
        \includegraphics[width=0.85\linewidth]{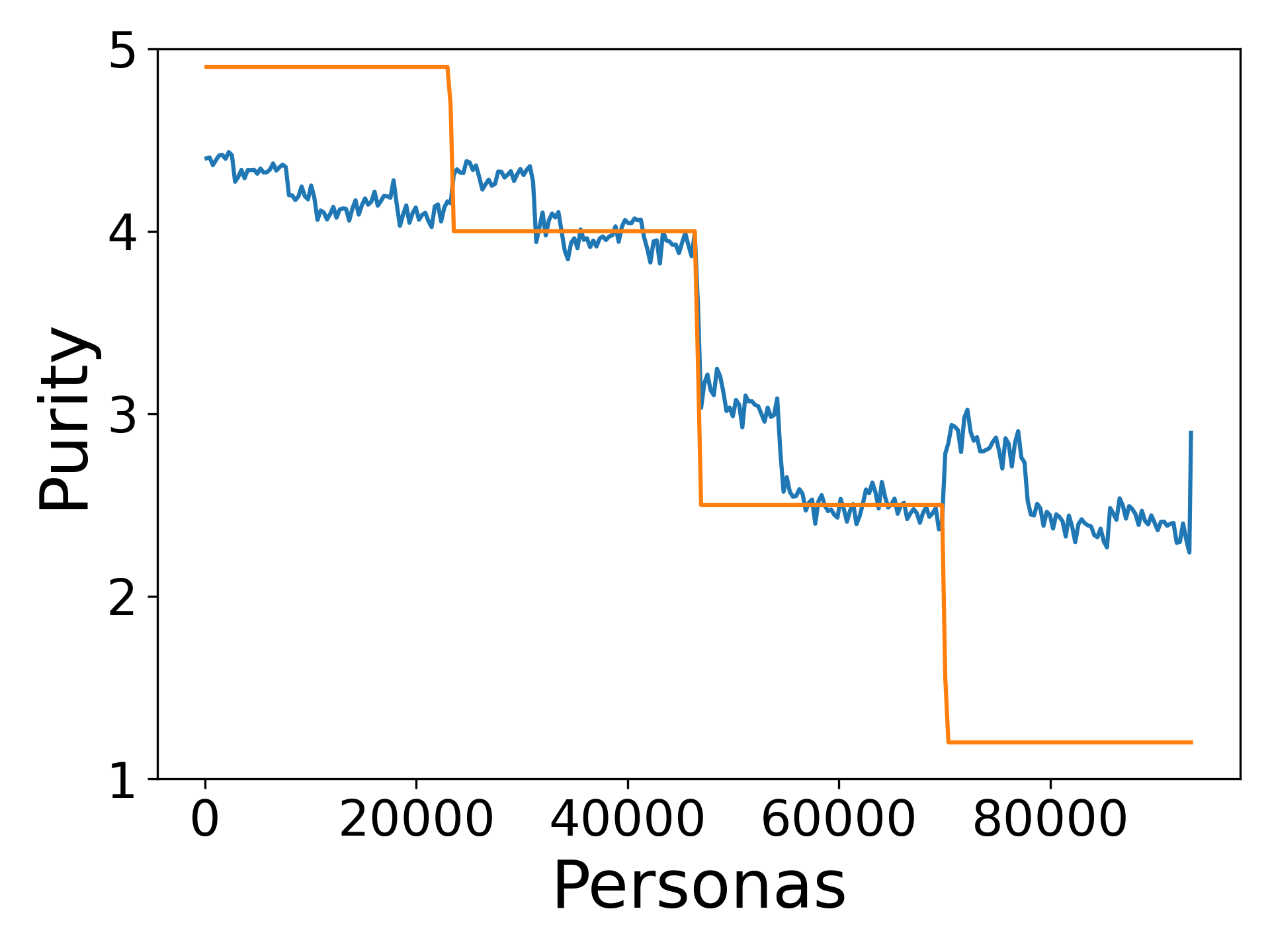}
        \caption{}
    \end{subfigure}

    \caption{Comparison between MFQ-2–based (blue) and mapping-based (orange) moral foundation scores. Mapping-based scores are obtained using a greedy-selected subset of cultural variables. Personas are ordered along the x-axis; the y-axis reports foundation scores. To avoid cluttering, lines of both score types  are smoothed by averaging scores within consecutive bins of 300 personas (according to the ordering fixed through sorting  by cultural variable then by admissible values).}
    \label{fig:mfq2-greedy}
\end{figure*}

Figures \ref{fig:mfq2-all10}  and    \ref{fig:mfq2-greedy} show the comparison between MFQ-2–based (blue) and mapping-based moral (orange) foundation scores, using all cultural variables (Figure \ref{fig:mfq2-all10}) or the selected core-set (Figure \ref{fig:mfq2-greedy}).

Table \ref{tab:selection_stability} shows the stability of core-variable selection across 50 independent runs.

\begin{figure*}
    \centering
    \begin{subfigure}{0.5\textwidth}
        \centering
        \includegraphics[width=1\linewidth]{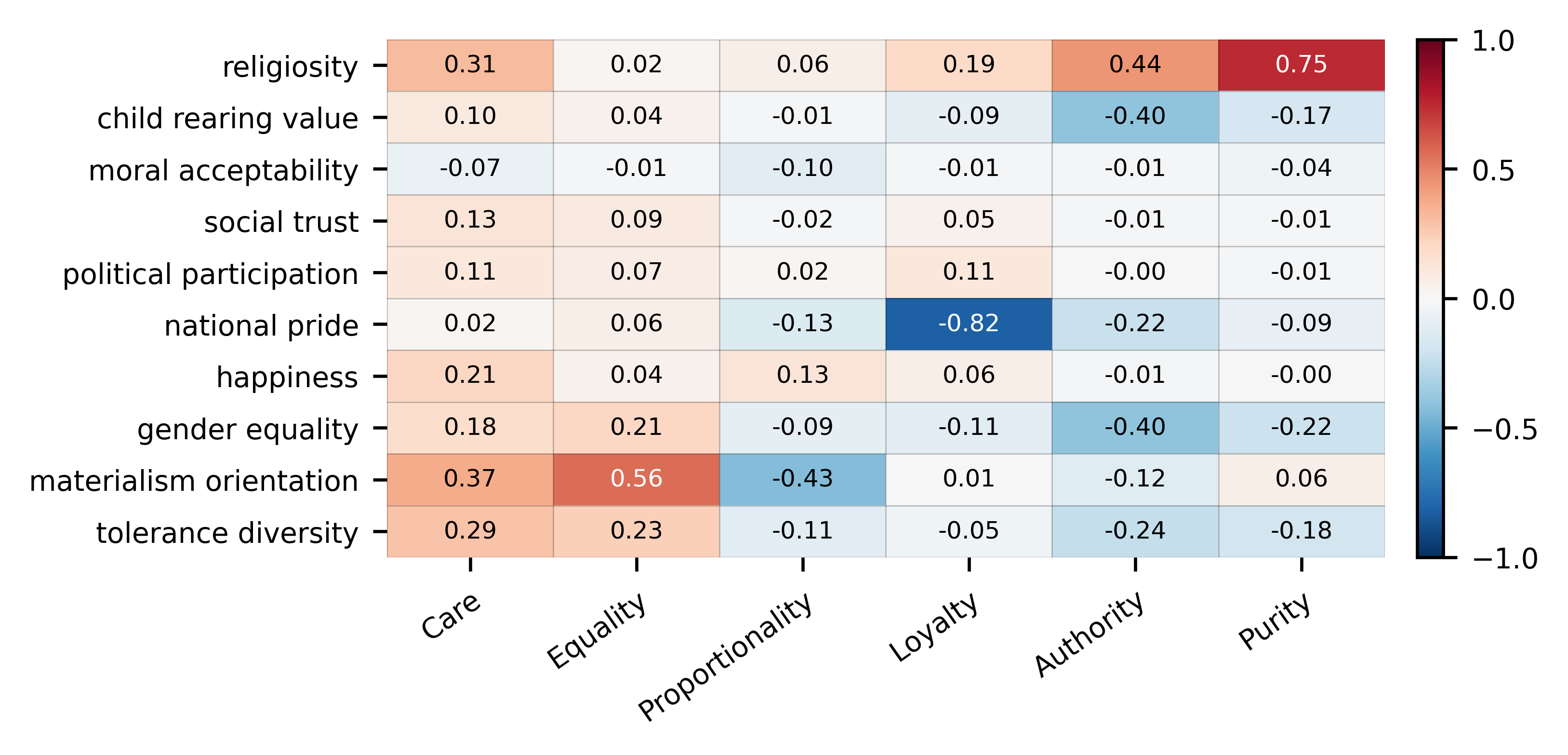}
        \caption{}
    \end{subfigure}\hfill
    \begin{subfigure}{0.5\textwidth}
        \centering
        \includegraphics[width=1\linewidth]{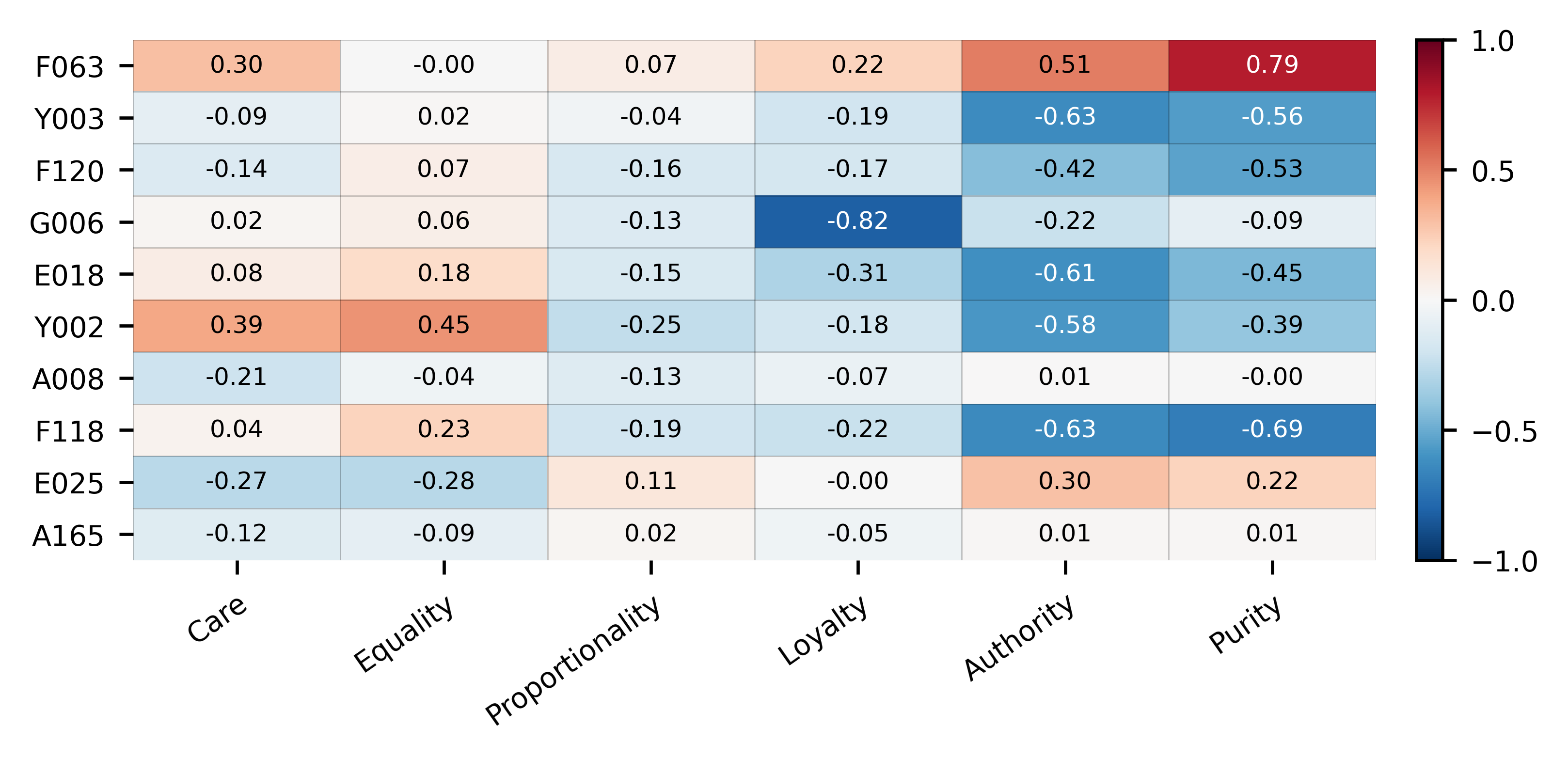}
        \caption{}
    \end{subfigure}

    \caption{\color{black} Spearman rank correlations between (a) IW indicators and MFQ2 scores, and (b) cultural variables and MFQ2 scores. All variables are represented at the persona level. Categorical cultural variables have an intrinsic order, so we mapped them to ordinal numeric values; the direction of the ordering is chosen to be as consistent as possible with the ordering implied by the most similar IW indicators. Spearman's $\rho$ is then is computed by correlating the rank-transformed values of each variable. IW indicator codes: F063 (importance of God), Y003 (autonomy index), F120 (justifiability of abortion), G006 (national pride), E018 (respect for authority), Y002 (post-materialist index),  A008 (happiness),  F118 (justifiability of homosexuality), E025 (signing petition), A165 (people trust). }
    \label{fig:spear}
\end{figure*}

Figure \ref{fig:spear} shows the Spearman rank correlations between ) IW indicators and MFQ2 scores, and between cultural variables and MFQ2 scores.

Figure  \ref{fig:iw-mfq} shows the personas projected onto the IW map, stratified by moral foundation score levels.

Table \ref{tab:mfq2-core-detailed} reports the MFQ-2 scores by core cultural variables and admissible values.

\section{Additional Results with \textsf{Qwen3.5-9B} and \textsf{Llama3.1-8B-Instruct}}
\label{sec:qwen_llama}

\subsection{Cultural Positioning and Pattern Consistency (RQ1)}

\begin{figure*}[ht!]
    \centering

    \begin{subfigure}{0.3\textwidth}
        \centering
        \includegraphics[width=1\linewidth]{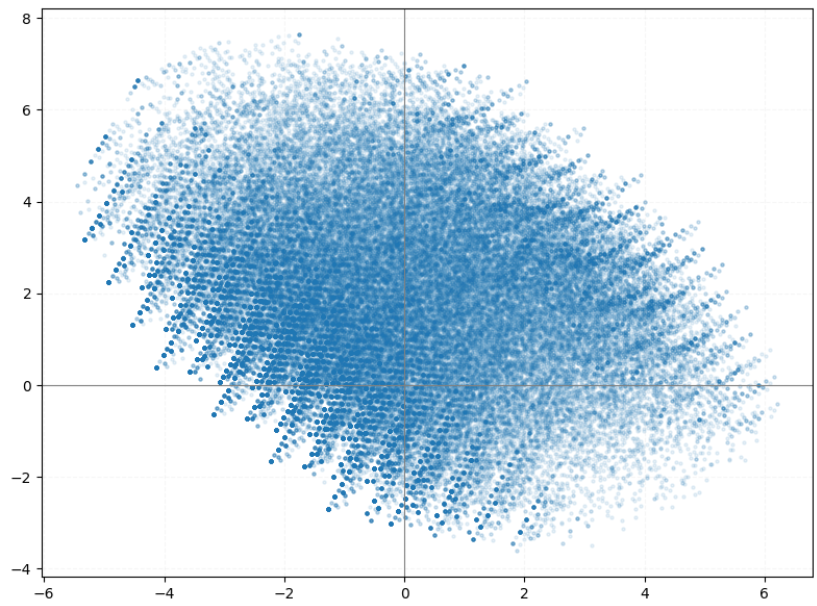}
        \caption{\textsf{gpt-oss-20b}}
    \end{subfigure}\hfill
    \begin{subfigure}{0.3\textwidth}
        \centering
        \includegraphics[width=1\linewidth]{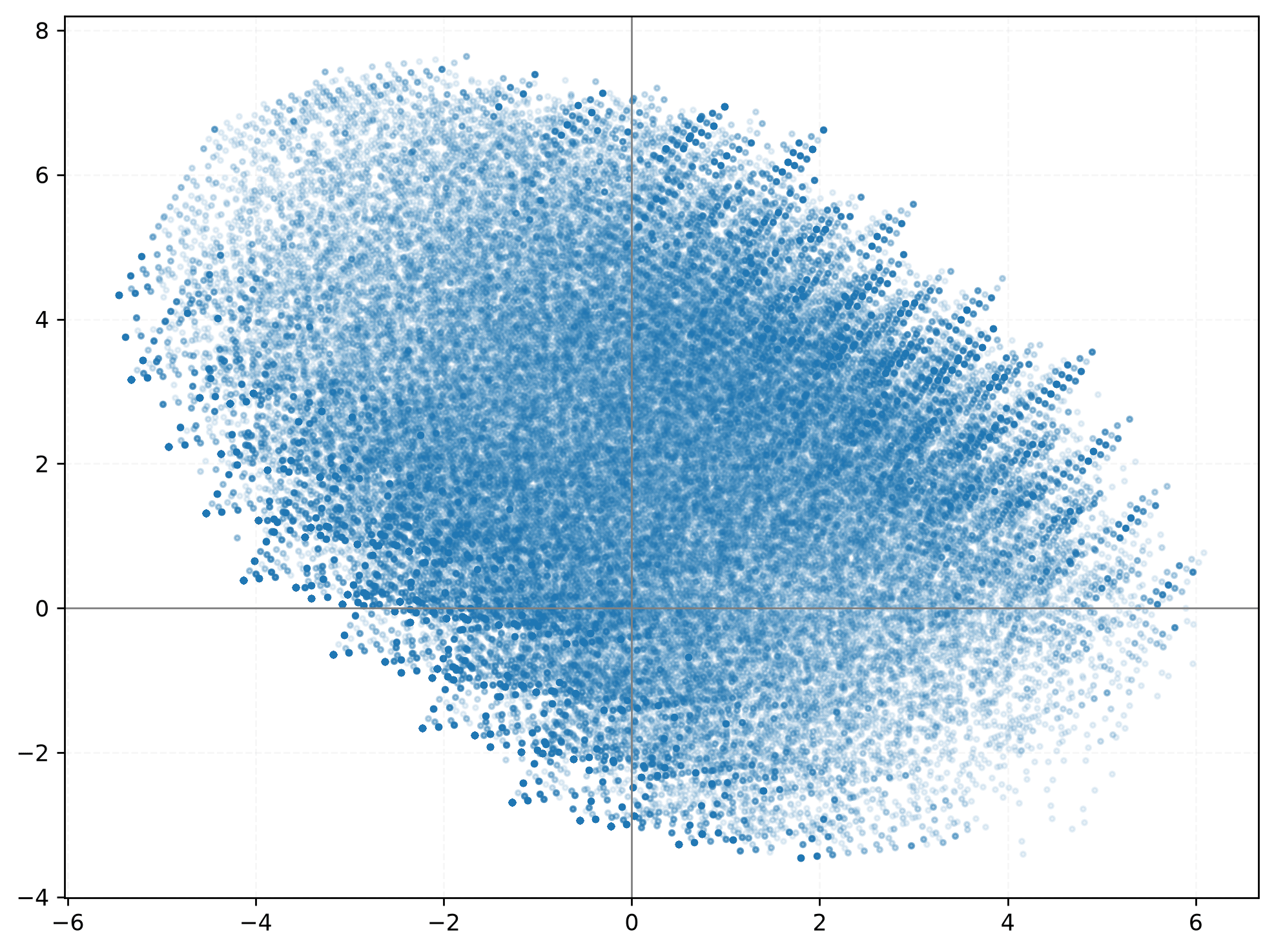}
        \caption{\textsf{Qwen3.5-9B}}
    \end{subfigure}\hfill
    \begin{subfigure}{0.3\textwidth}
        \centering
        \includegraphics[width=1\linewidth]{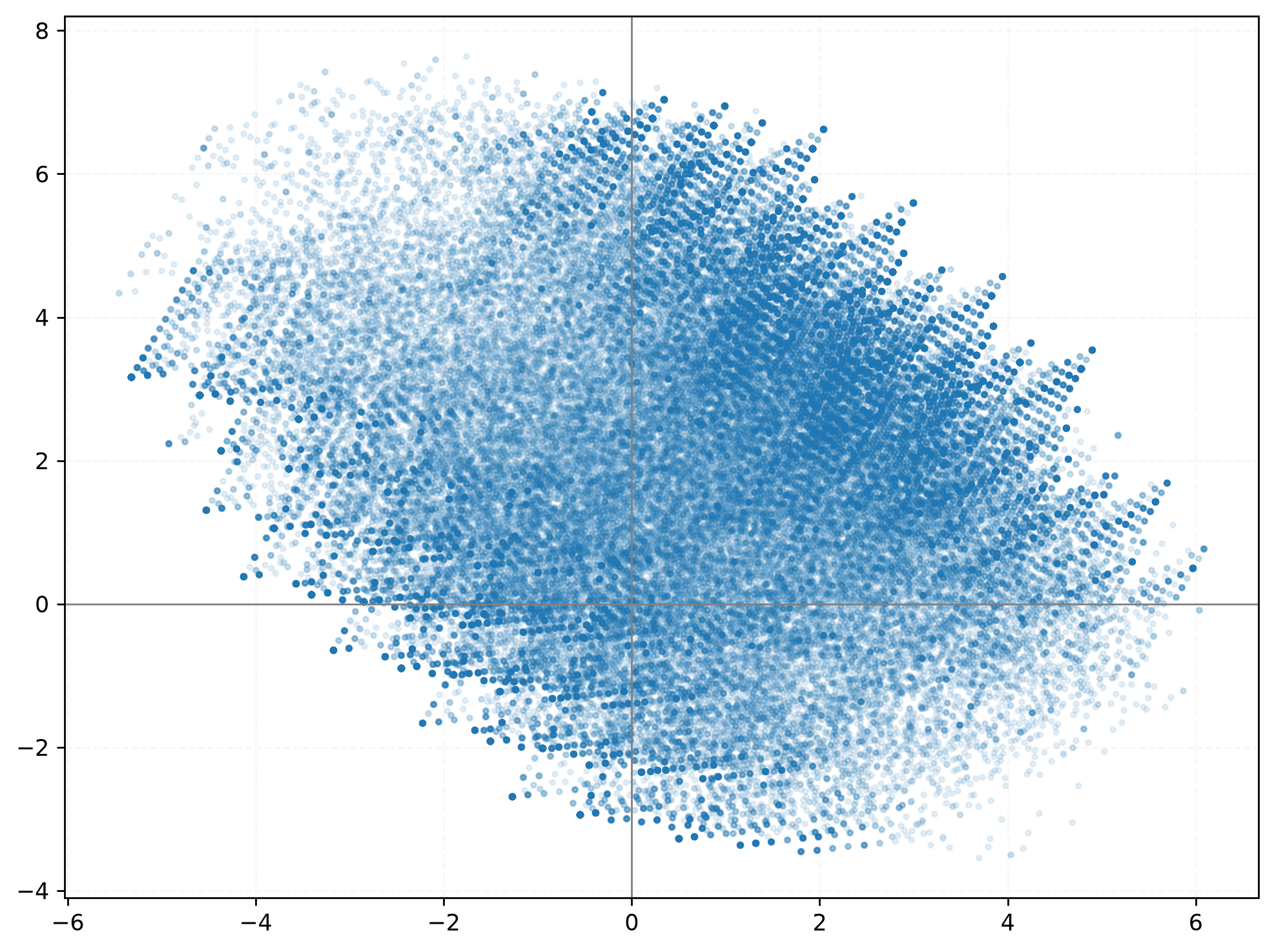}
        \caption{\textsf{Llama3.1-8B-Instruct}}
    \end{subfigure}

    \caption{Inglehart--Welzel maps of the culturally grounded personas.}
    \label{fig:all-iw-mfq}
\end{figure*}

\begin{figure*}[ht!]
    \centering

    \begin{subfigure}{0.3\textwidth}
        \centering
        \includegraphics[width=1\linewidth]{images/iw_itemset_retainedcells_happiness_very_happy__social_trust_most_people_trusted.png}
        \caption{gpt-oss-20B}
    \end{subfigure}\hfill
    \begin{subfigure}{0.3\textwidth}
        \centering
        \includegraphics[width=1\linewidth]{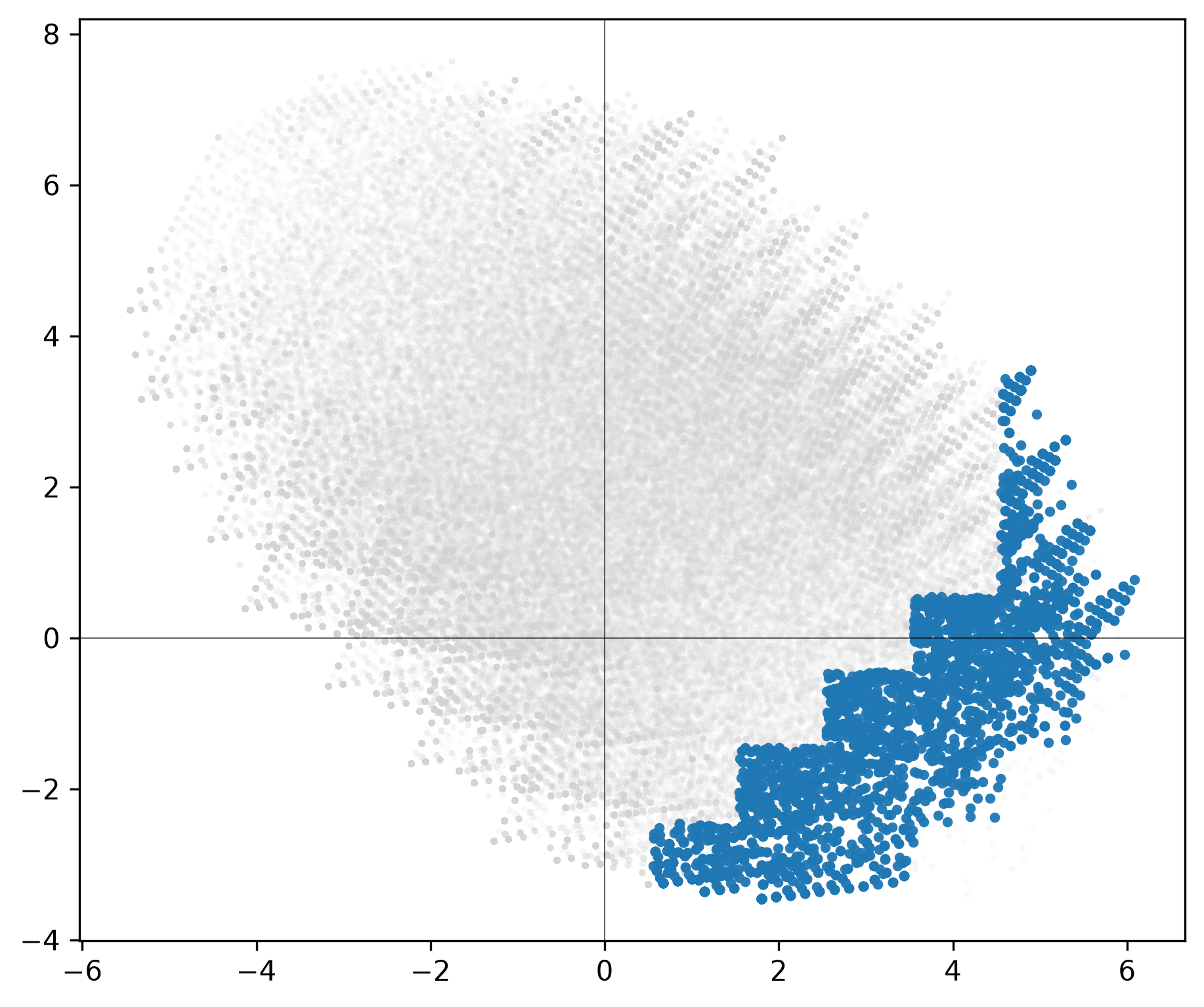}
        \caption{\textsf{Qwen3.5-9B}}
    \end{subfigure}\hfill
    \begin{subfigure}{0.3\textwidth}
        \centering
        \includegraphics[width=1\linewidth]{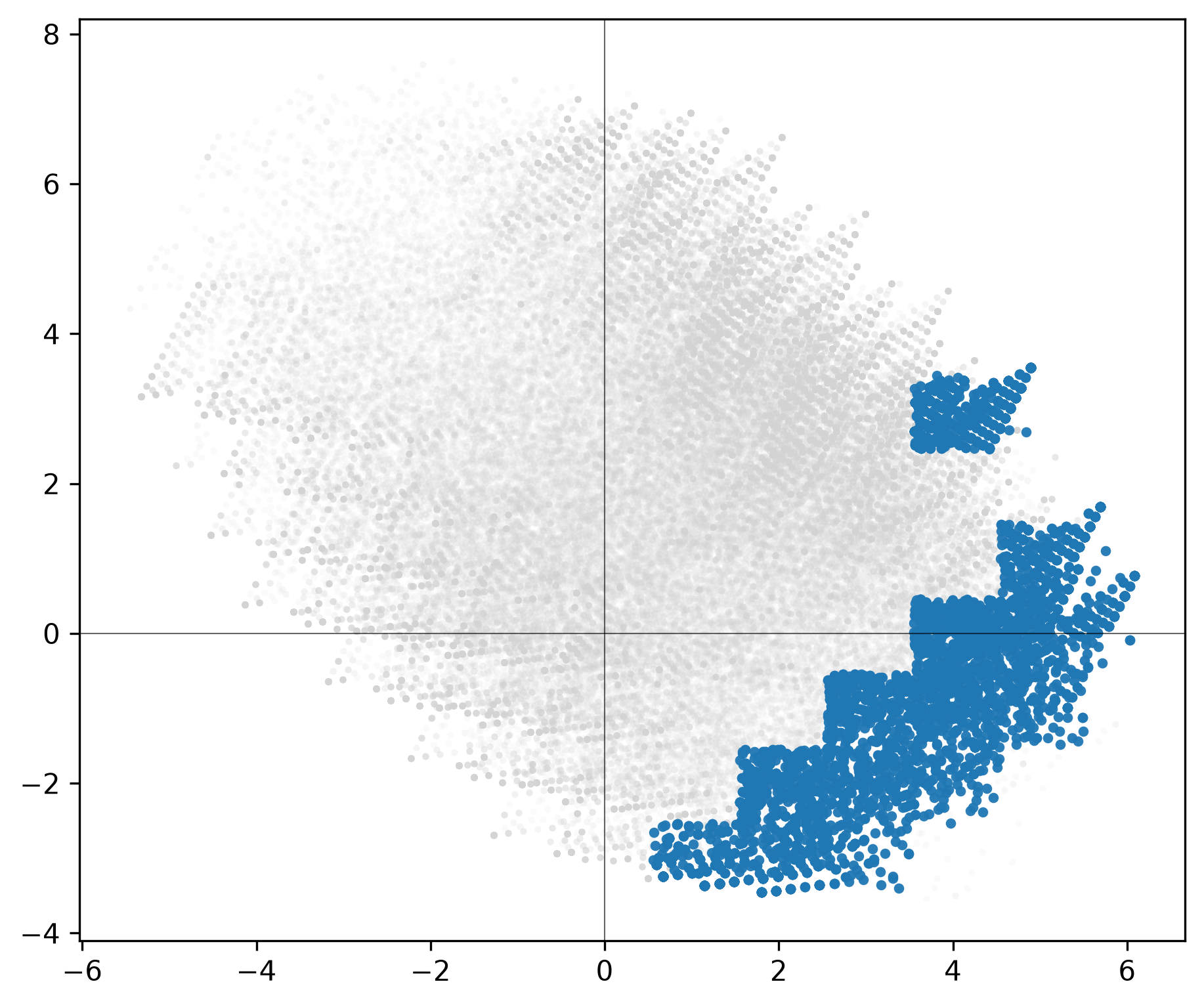}
        \caption{\textsf{Llama3.1-8B-Instruct}}
    \end{subfigure}

    \begin{subfigure}{0.3\textwidth}
        \centering
        \includegraphics[width=1\linewidth]{images/iw_itemset_retainedcells_happiness_not_at_all_happy__social_trust_cannot_trust_people.png}
        \caption{gpt-oss-20B}
    \end{subfigure}\hfill
    \begin{subfigure}{0.3\textwidth}
        \centering
        \includegraphics[width=1\linewidth]{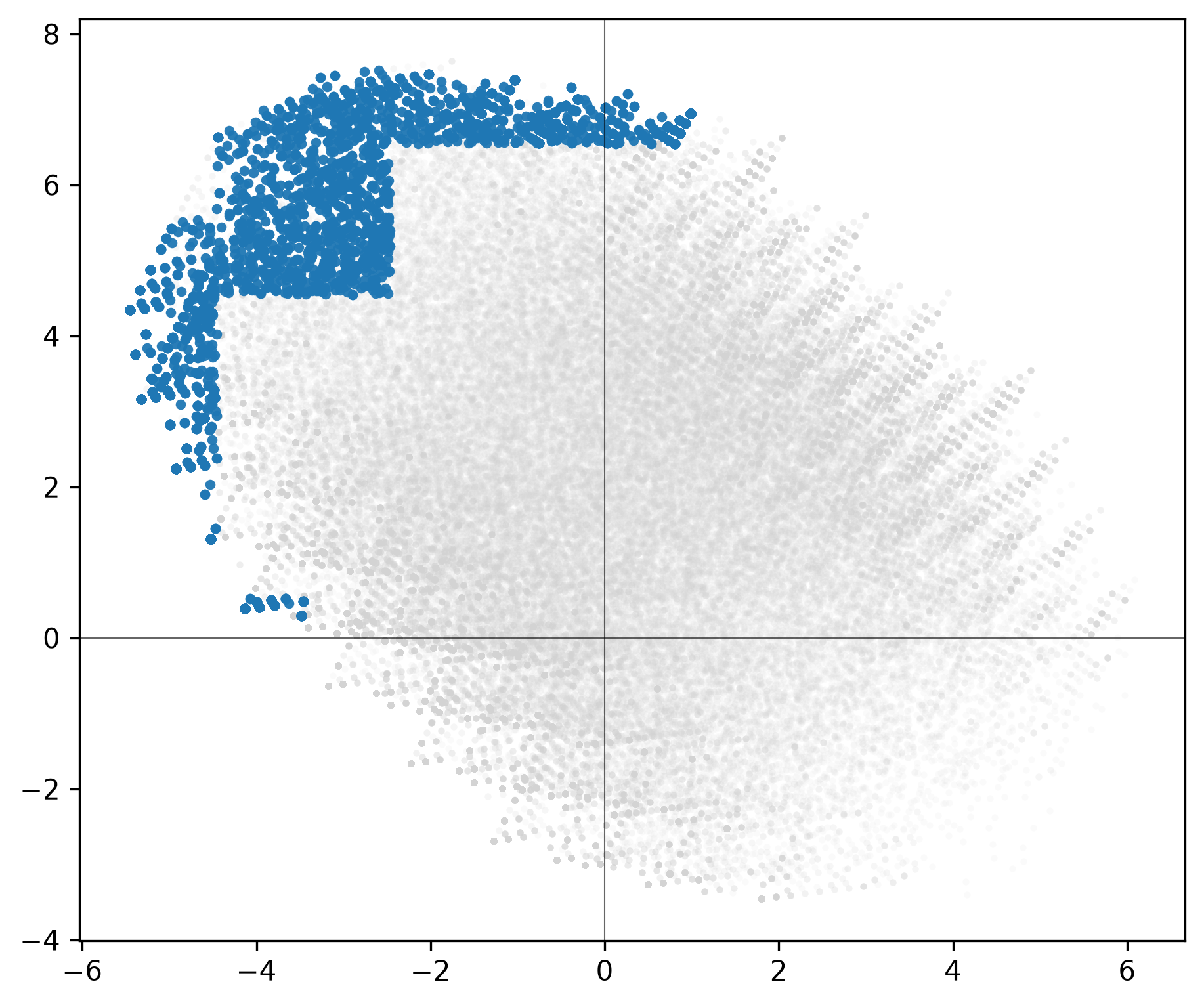}
        \caption{\textsf{Qwen3.5-9B}}
    \end{subfigure}\hfill
    \begin{subfigure}{0.3\textwidth}
        \centering
        \includegraphics[width=1\linewidth]{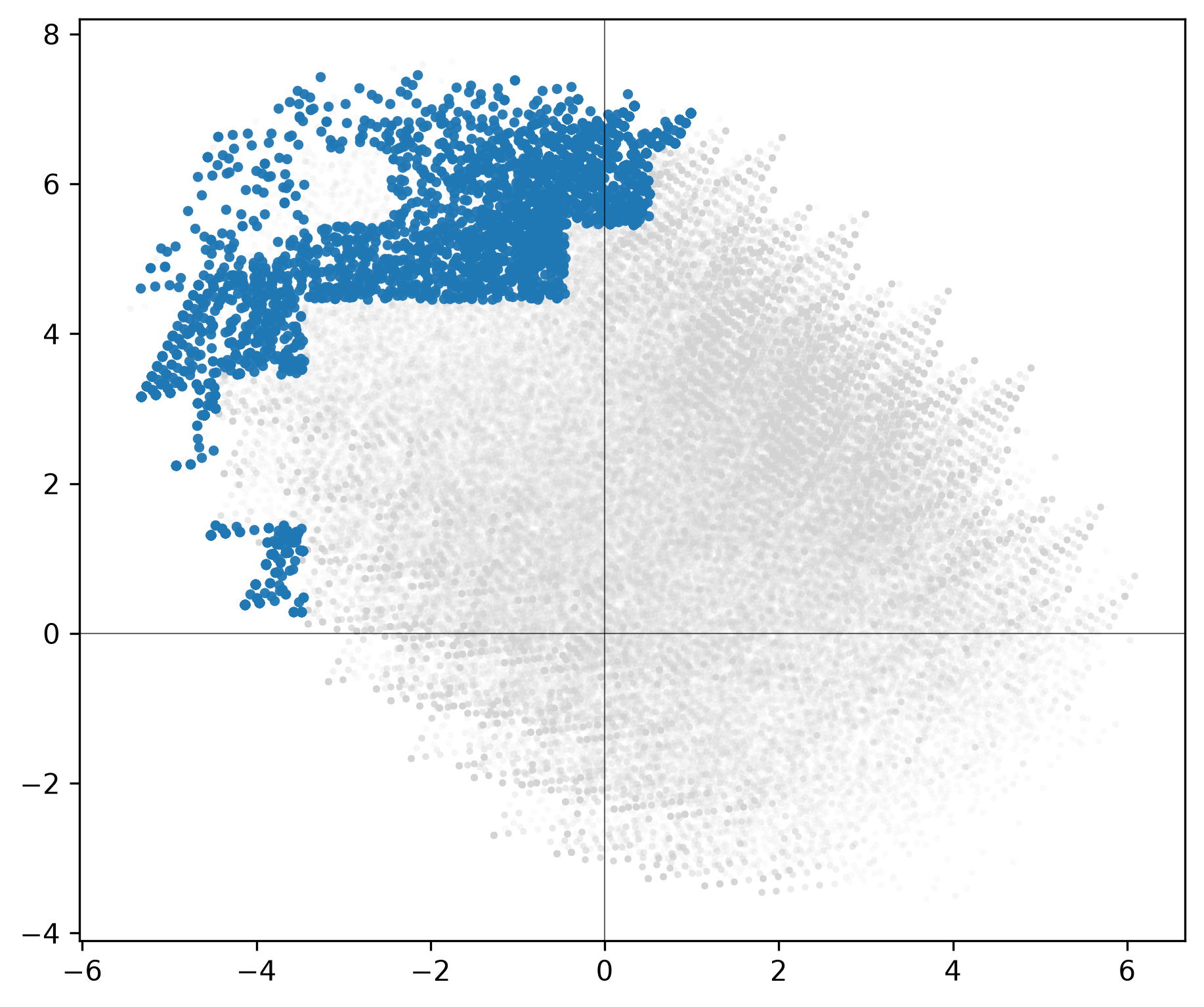}
        \caption{\textsf{Llama3.1-8B-Instruct}}
    \end{subfigure}


    \begin{subfigure}{0.3\textwidth}
        \centering
        \includegraphics[width=1\linewidth]{images/iw_itemset_retainedcells_happiness_very_happy__national_pride_very_proud__social_trust_most_people_trusted.png}
        \caption{gpt-oss-20B}
    \end{subfigure}\hfill
    \begin{subfigure}{0.3\textwidth}
        \centering
        \includegraphics[width=1\linewidth]{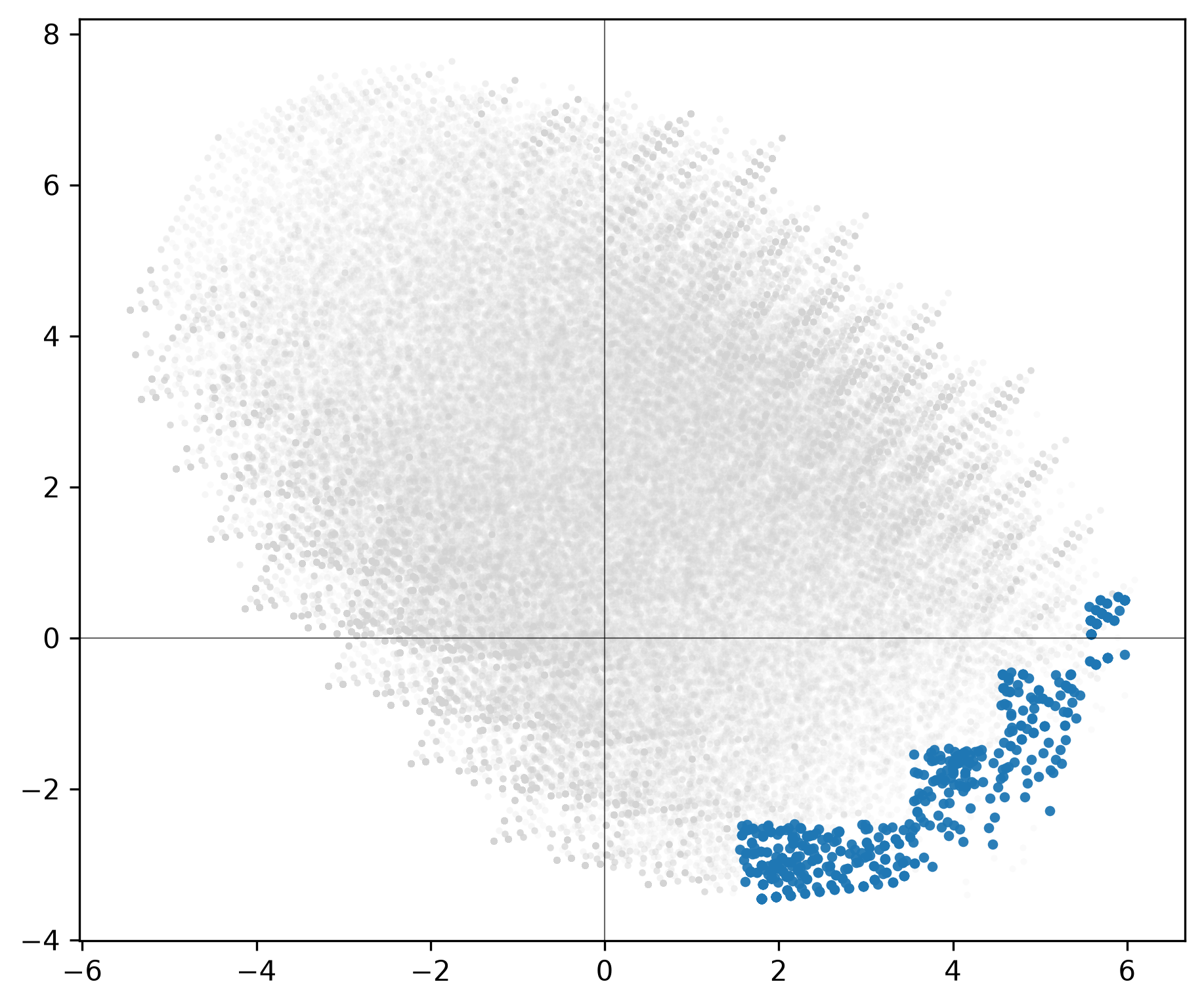}
        \caption{\textsf{Qwen3.5-9B}}
    \end{subfigure}\hfill
    \begin{subfigure}{0.3\textwidth}
        \centering
        \includegraphics[width=1\linewidth]{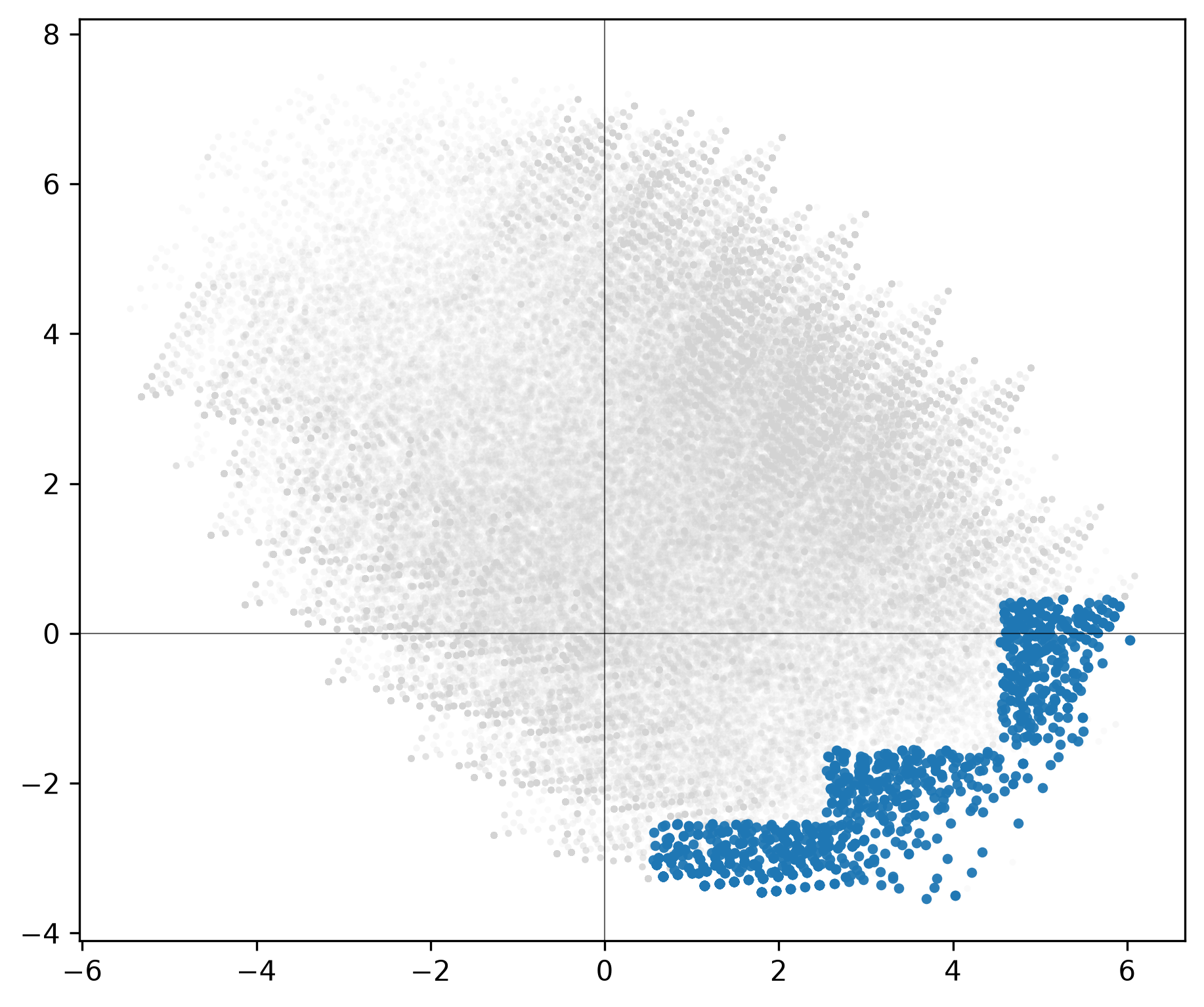}
        \caption{\textsf{Llama3.1-8B-Instruct}}
    \end{subfigure}


    \begin{subfigure}{0.3\textwidth}
        \centering
        \includegraphics[width=1\linewidth]{images/iw_itemset_retainedcells_child_rearing_value_obedience_faith__social_trust_cannot_trust_people.png}
        \caption{gpt-oss-20B}
    \end{subfigure}\hfill
    \begin{subfigure}{0.3\textwidth}
        \centering
        \includegraphics[width=1\linewidth]{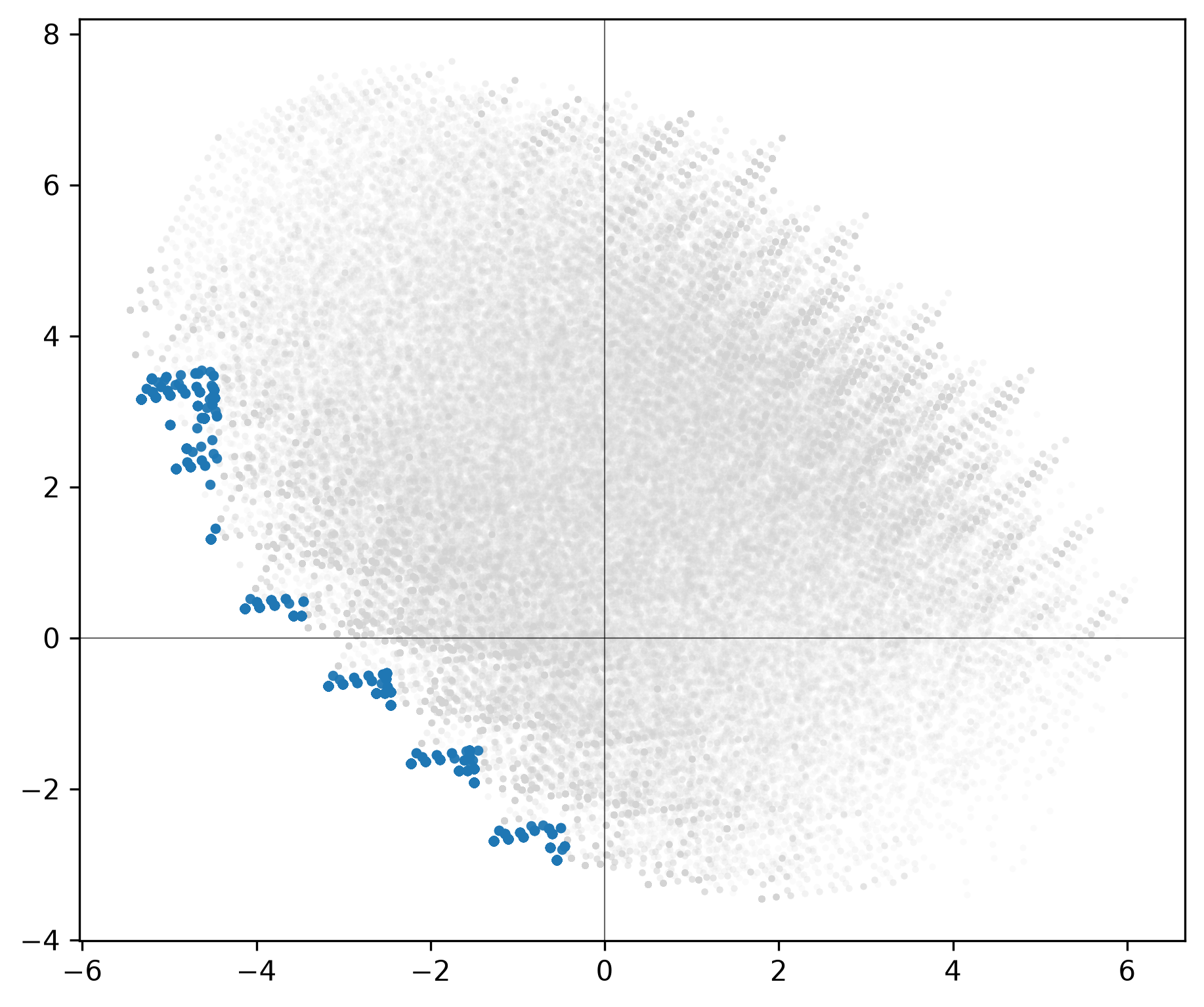}
        \caption{\textsf{Qwen3.5-9B}}
    \end{subfigure}\hfill
    \begin{subfigure}{0.3\textwidth}
        \centering
        \includegraphics[width=1\linewidth]{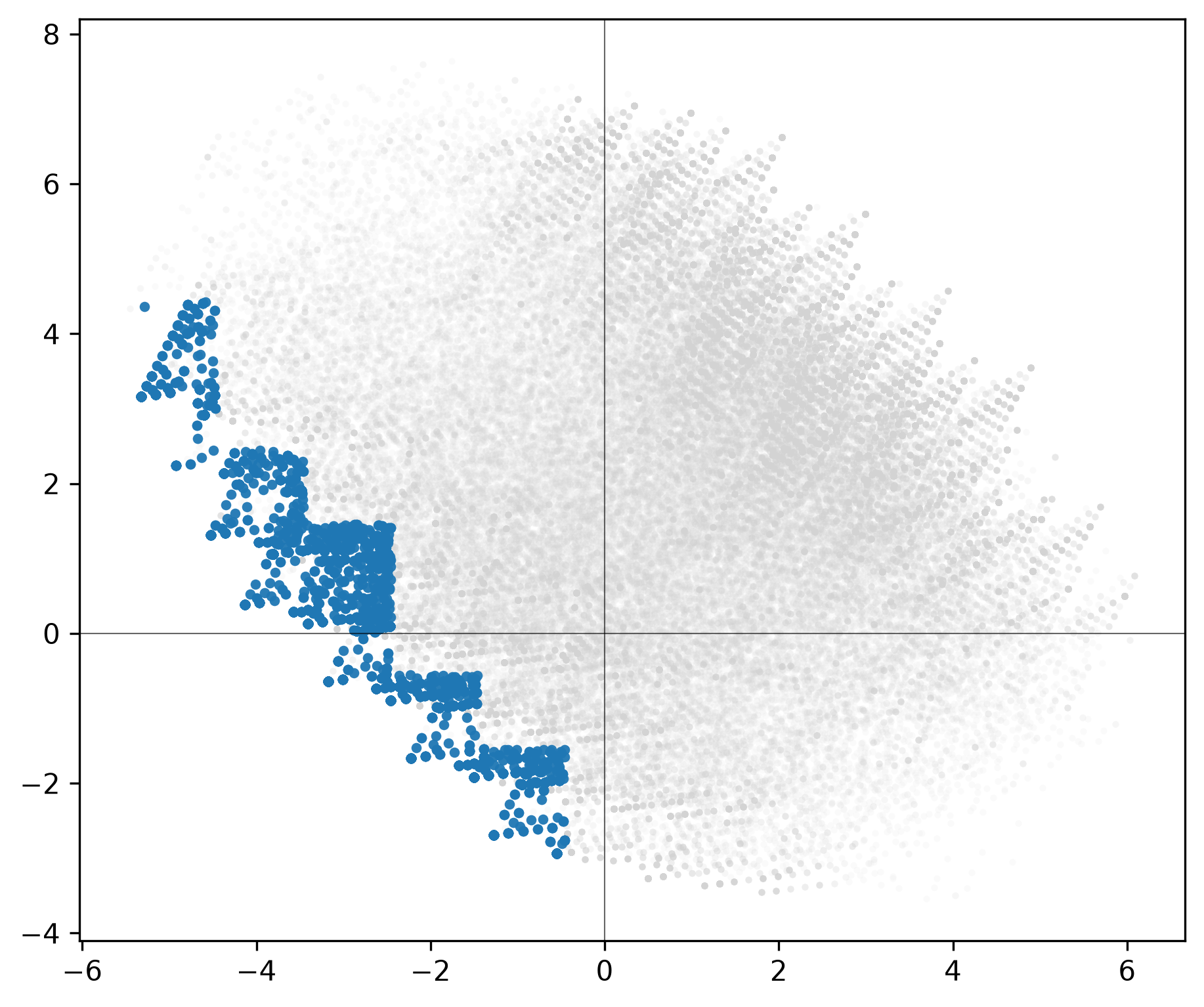}
        \caption{\textsf{Llama3.1-8B-Instruct}}
    \end{subfigure}

    \caption{Cross-model comparison of the spatial footprints of four closed frequent cultural-configuration itemsets shared by \textsf{gpt-oss-20B}, \textsf{Qwen3.5-9B}, and \textsf{Llama3.1-8B-Instruct} in the Inglehart--Welzel space. Rows correspond, from top to bottom, to: \emph{very happy; most people trusted}; \emph{not at all happy; cannot trust people}; \emph{very happy; very proud; most people trusted}; and \emph{obedience--faith; cannot trust people}. In each panel, points represent the full set of model-generated personas, while highlighted grid cells indicate the regions in which the corresponding itemset is retained. All models are analyzed using the same grid and mining configuration: cell side length $1$, minimum support $0.2$, $\rho=0.5$, and a minimum of two personas per cell.}
    \label{fig:qwen-llama-gpt-itemsets}
\end{figure*}

Figure \ref{fig:all-iw-mfq} shows that all three models generate broad and continuous distributions in the IW space, with a comparable overall spatial organization and no clearly separated clusters. The pairwise distances reported in Table \ref{tab:iw-pairwise-distances} indicate that matched cultural conditionings are generally positioned in similar regions, although their exact coordinates vary across models; \textsf{Qwen3.5-9B} is slightly closer to \textsf{gpt-oss-20B} than  \textsf{Llama3.1-8B-Instruct}. Table \ref{tab:iw-closed-itemsets-global} shows comparable grid coverage and a similar scale of retained patterns across the three models, with moderate differences in the number of unique and multi-cell itemsets. Table \ref{tab:iw-selected-itemsets-cross-model} examines selected non-singleton patterns in greater detail. 
Most of the main configurations selected for \textsf{gpt-oss-20B} in Section \ref{sec:rq1results} are also retained by \textsf{Qwen3.5-9B} and \textsf{Llama3.1-8B-Instruct}, including the high-trust/high-happiness and low-trust/low-happiness patterns, their combination with national pride, and the association between obedience--faith and low trust. These associations are retained across models, although their mean within-cell support and spatial coverage vary. 
The table also illustrates selected additional configurations shared across models or retained only by particular models, showing model-dependent variation alongside the shared core.
Table \ref{tab:iw-shared-vs-model-specific-itemsets} further shows that non-singleton itemsets shared by all three models generally have broader spatial coverage, occur in more populated cells, and exhibit slightly higher support than model-specific patterns. Most model-specific itemsets are instead localized in no more than three cells; although \textsf{Llama3.1-8B-Instruct} retains more of them, they remain predominantly spatially circumscribed. Figure \ref{fig:qwen-llama-gpt-itemsets} complements this comparison by showing the spatial footprints of four shared configurations. Across the three models, these patterns occupy broadly corresponding regions of the IW space, although the extent of the highlighted areas varies slightly.
 
\subsection{Demographic-Level Alignment Across Models (RQ2)}

\begin{figure*}[h]
    \centering
    \includegraphics[width=\linewidth]{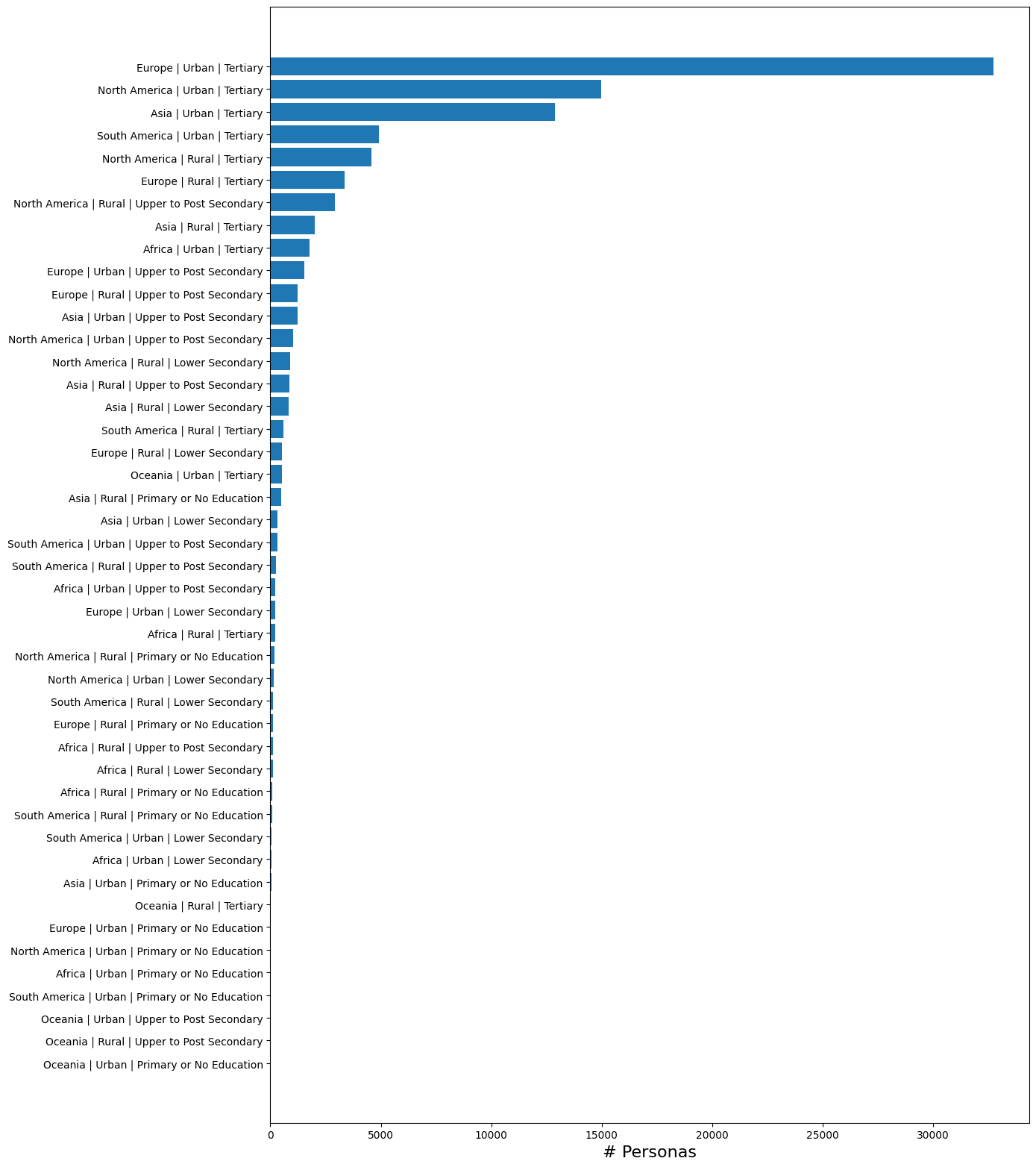}
    \caption{Distribution of demographic triples obtained during the WVB-Probe prompting. }
    \label{fig:wvb-demotriad}
\end{figure*}

\begin{figure*}[h]
    \centering
    \includegraphics[width=\linewidth]{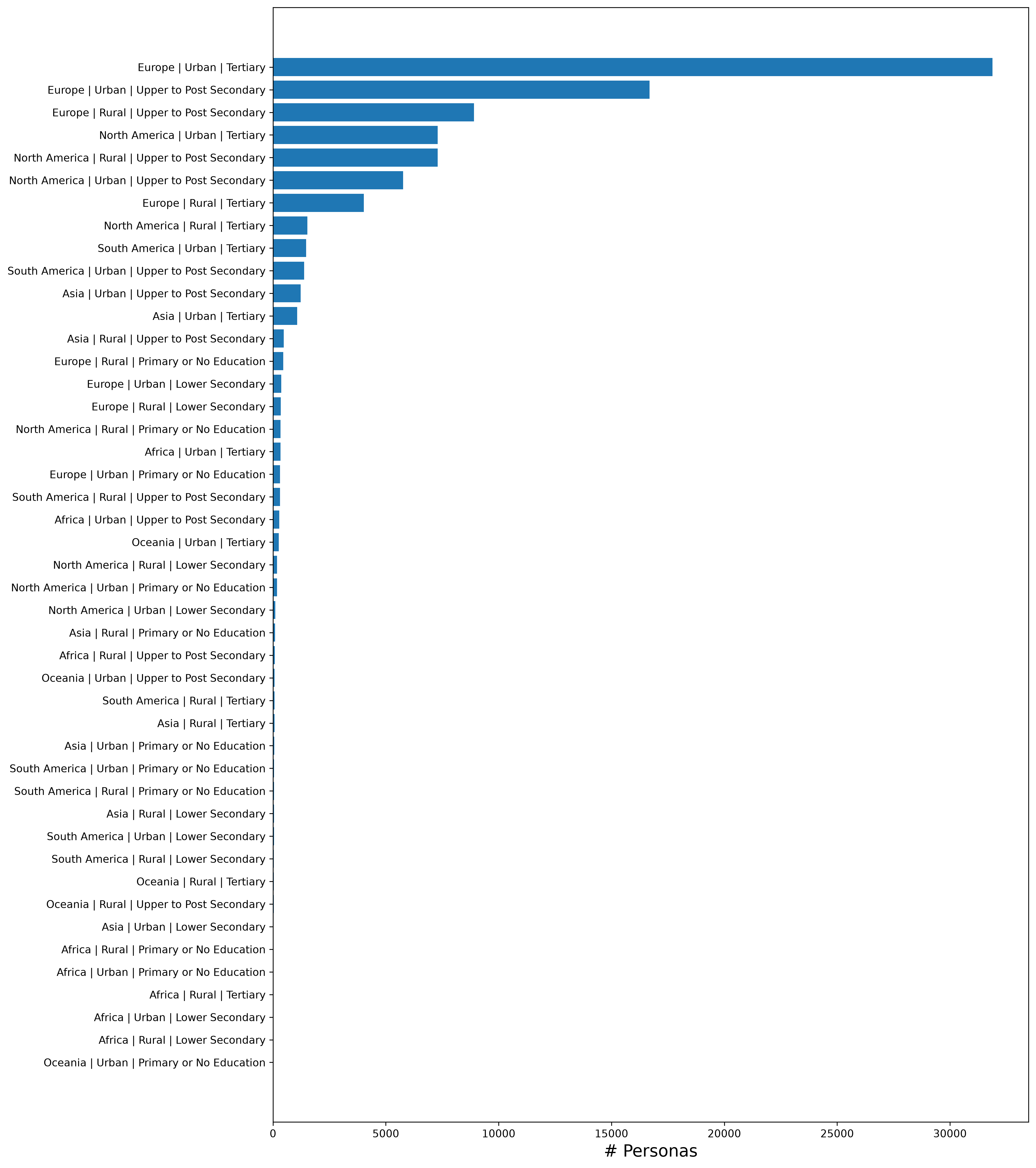}
    \caption{Distribution of demographic triples assigned by \textsf{Qwen3.5-9B} during WVB-Probe prompting. }
    \label{fig:qwen-wvb-demotriad}
\end{figure*}

\begin{figure*}[h]
    \centering
    \includegraphics[width=\linewidth]{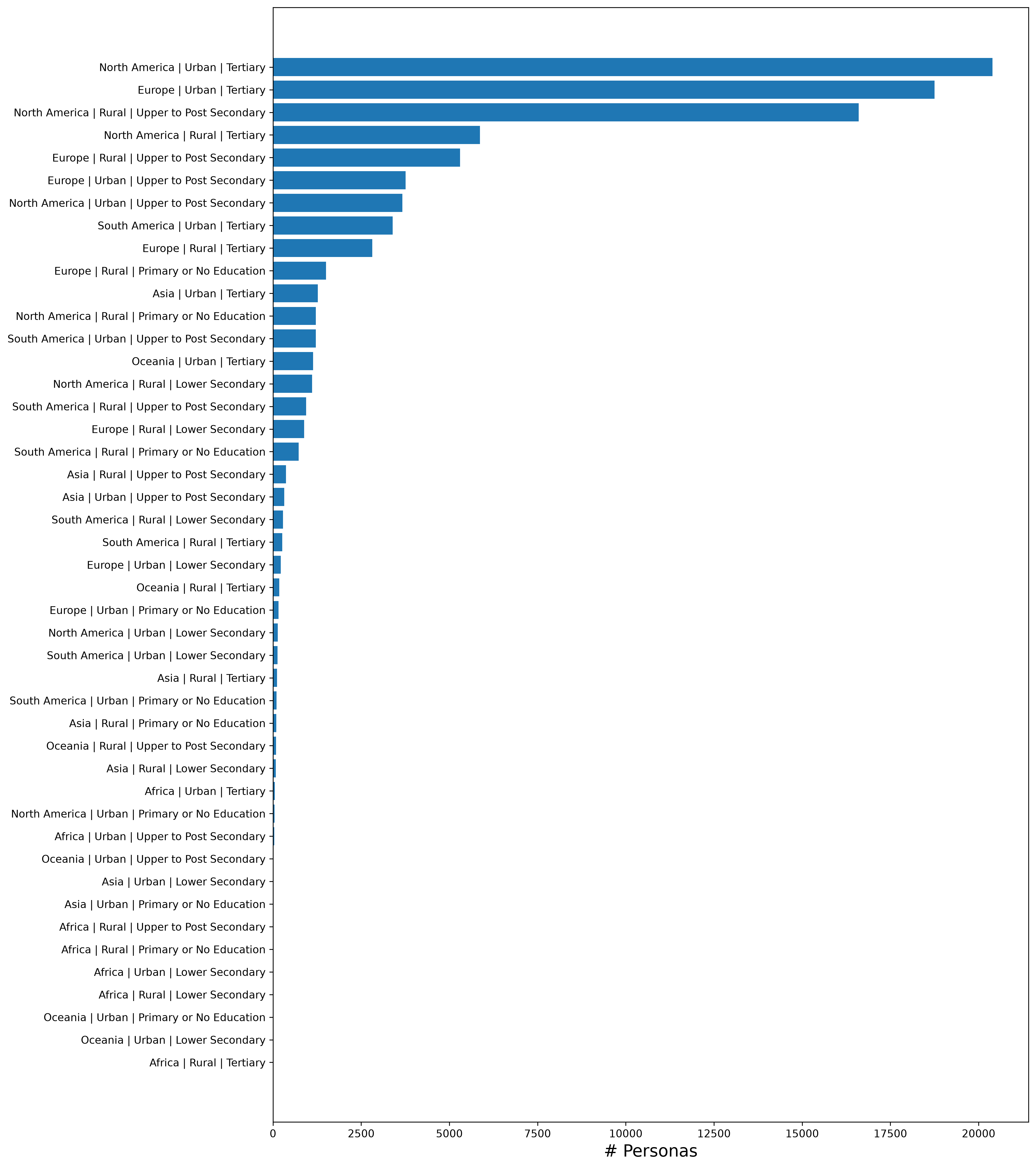}
    \caption{Distribution of demographic triples assigned by Llama3.1-8B-Instruct during WVB-Probe prompting.}
    \label{fig:llama-wvb-demotriad}
\end{figure*}

Figures~\ref{fig:qwen-wvb-demotriad} and \ref{fig:llama-wvb-demotriad}, together with Figure~\ref{fig:wvb-demotriad}, show that the demographic triples induced during WVB-Probe prompting are strongly concentrated in a limited number of profiles, although the dominant profiles differ across models. \textsf{gpt-oss-20B} and \textsf{Qwen3.5-9B} assign the largest number of personas to Europe--Urban--Tertiary, whereas \textsf{Llama3.1-8B-Instruct} assigns the largest number to North America--Urban--Tertiary. More generally, \textsf{Qwen3.5-9B} produces a more Europe-centered distribution, while \textsf{Llama3.1-8B-Instruct} assigns a larger share of personas to North American and rural groups.

The aggregate results in Tables~\ref{tab:qwen3-5-9b-high-run1-wvs-overall} and \ref{tab:llama3-1-8b-instruct-high-run1-wvs-overall}, compared with the \textsf{gpt-oss-20B} results reported in Section~\ref{sec:rq2_results} Table \ref{tab:wvb-probe-summary}, show similar overall WVB-Probe alignment across the three models.  

For all models, weighted results are higher than their unweighted counterparts, indicating that the more frequent demographic groups tend to exhibit somewhat stronger alignment within each model-specific demographic distribution. 
Because group weights are derived from each model's own assigned demographic distribution, weighted scores characterize alignment under different model-specific mixtures; unweighted scores instead provide the more direct equal-group comparison.
Across models, approximately 89--90\% of questions satisfy the moderate $EMD<0.4$ threshold under unweighted aggregation, whereas approximately 52--57\% satisfy the stricter $EMD<0.2$ threshold. This reproduces the main \textsf{gpt-oss-20B} finding: moderate alignment is widespread, while high alignment remains more selective.

The complete group-level results in Tables~\ref{tab:qwen-wvb-probe-all-groups} and \ref{tab:llama-wvb-probe-all-groups}, compared with Table \ref{tab:wvb-probe-all-groups} in Appendix~\ref{app:wvb}, further show that the most populated tertiary-education groups generally obtain $1-\overline{EMD}$ values close to 0.8 across models. 
Each model represents 45 of the 46 demographic groups with available WVB-Probe references, with 44 groups shared by all three models.
Nevertheless, the relative alignment of individual groups varies, and no model consistently obtains the highest score for every demographic profile.

\subsection{Culture--Morality Relations and Core-Set Stability (RQ3)}

\begin{figure*}[ht!]
    \centering

    \begin{subfigure}{0.25\textwidth}
        \centering
        \includegraphics[width=1\linewidth]{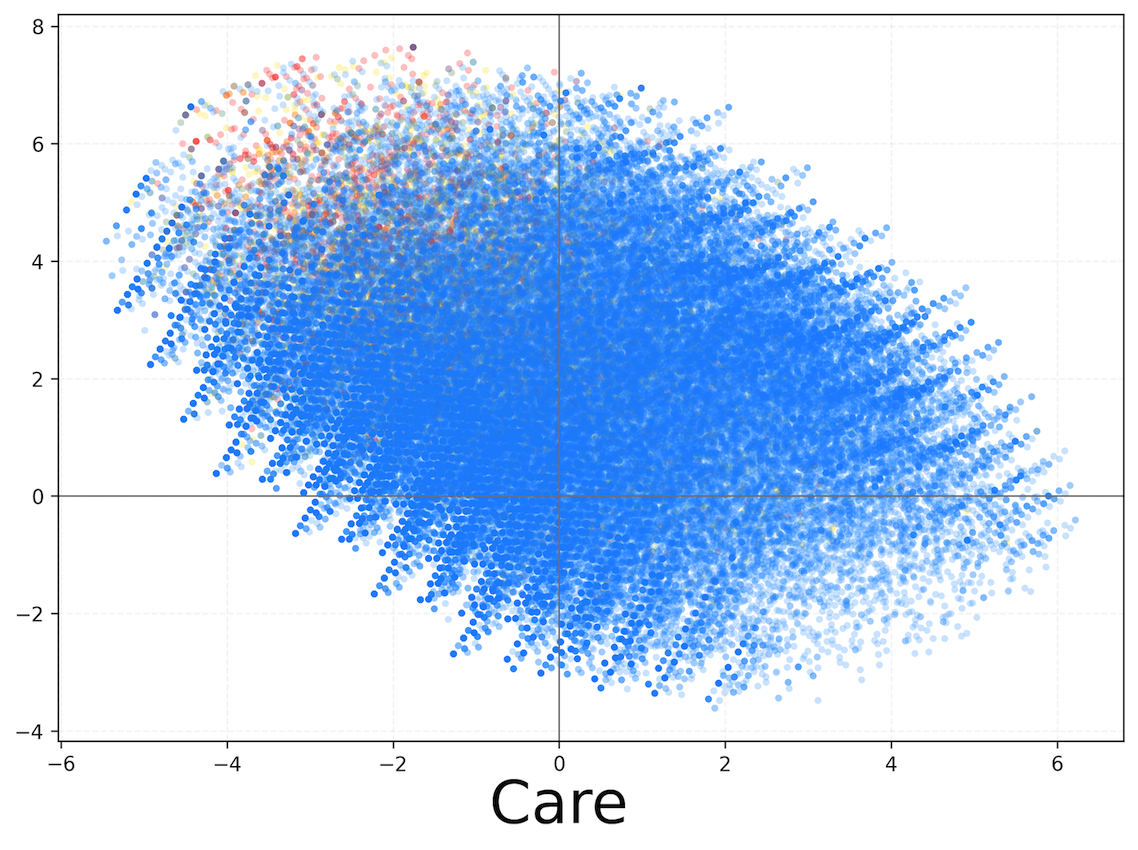}
        \caption{}
    \end{subfigure}\hfill
    \begin{subfigure}{0.25\textwidth}
        \centering
        \includegraphics[width=1\linewidth]{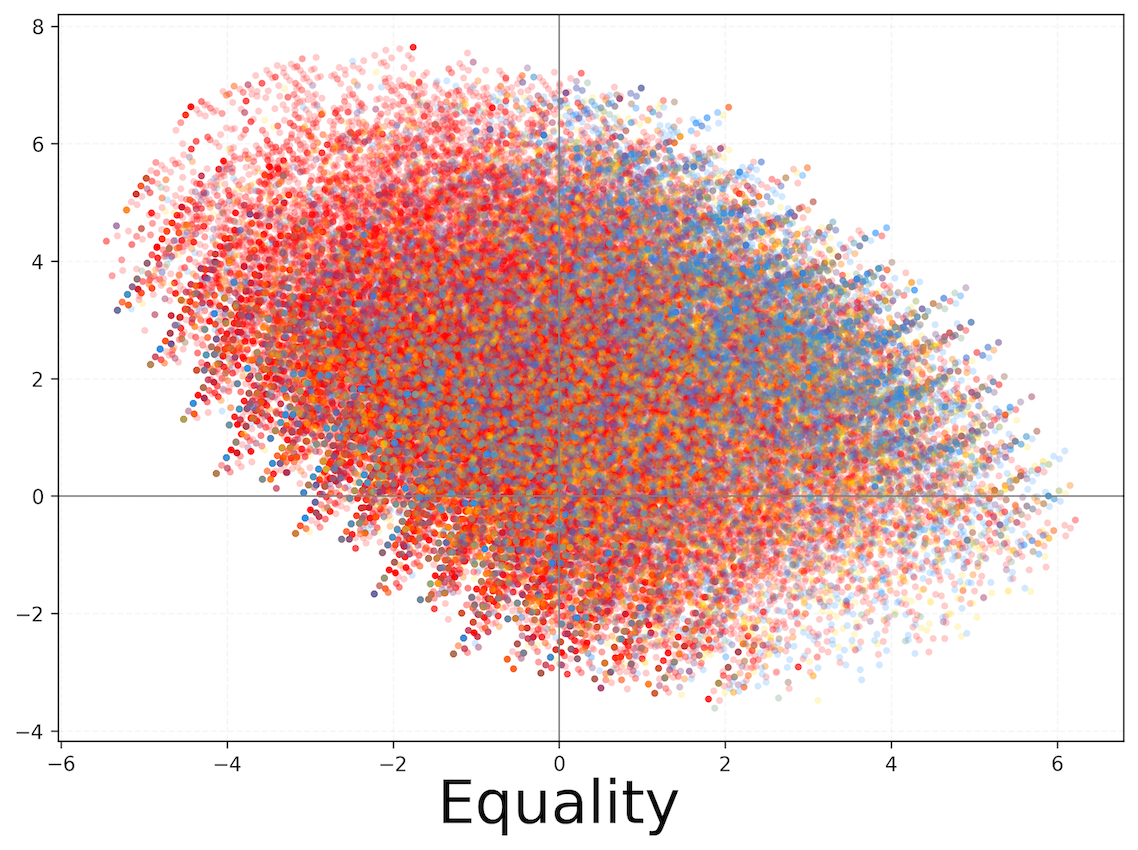}
        \caption{}
    \end{subfigure}\hfill
    \begin{subfigure}{0.25\textwidth}
        \centering
        \includegraphics[width=1\linewidth]{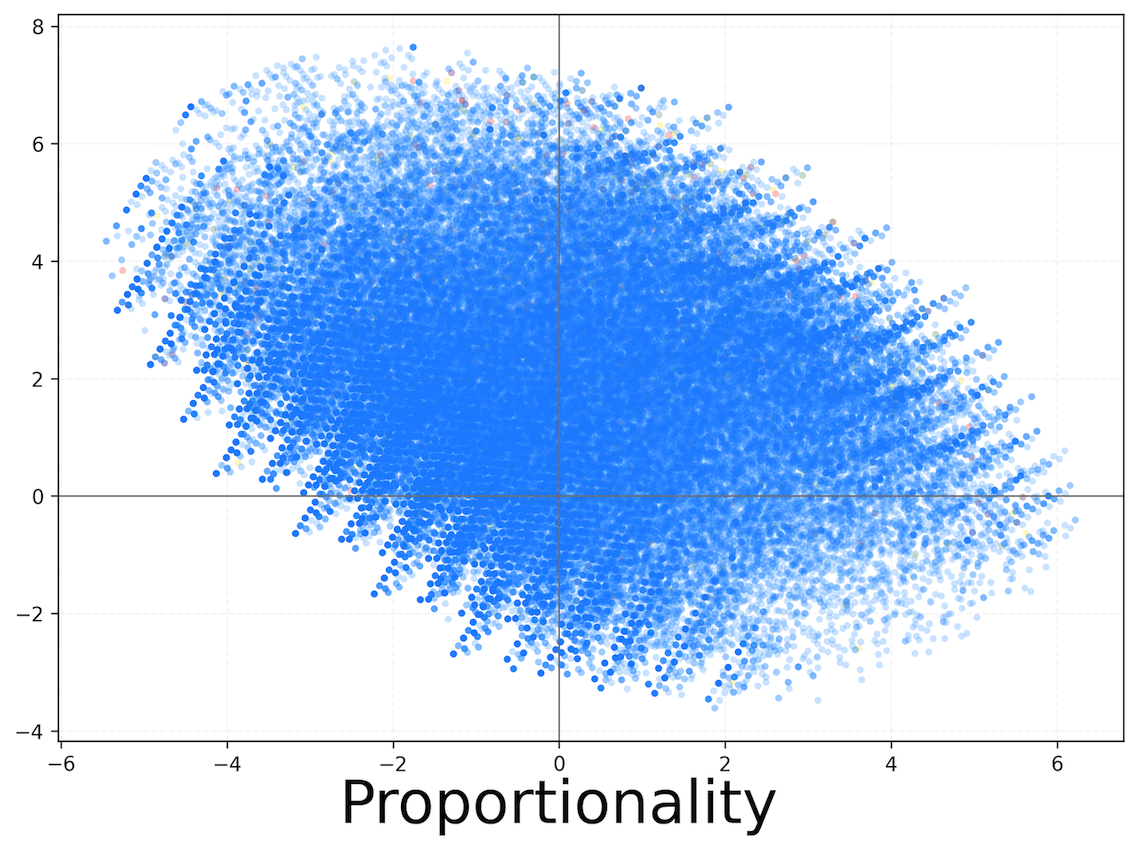}
        \caption{}
    \end{subfigure}


    \begin{subfigure}{0.25\textwidth}
        \centering
        \includegraphics[width=1\linewidth]{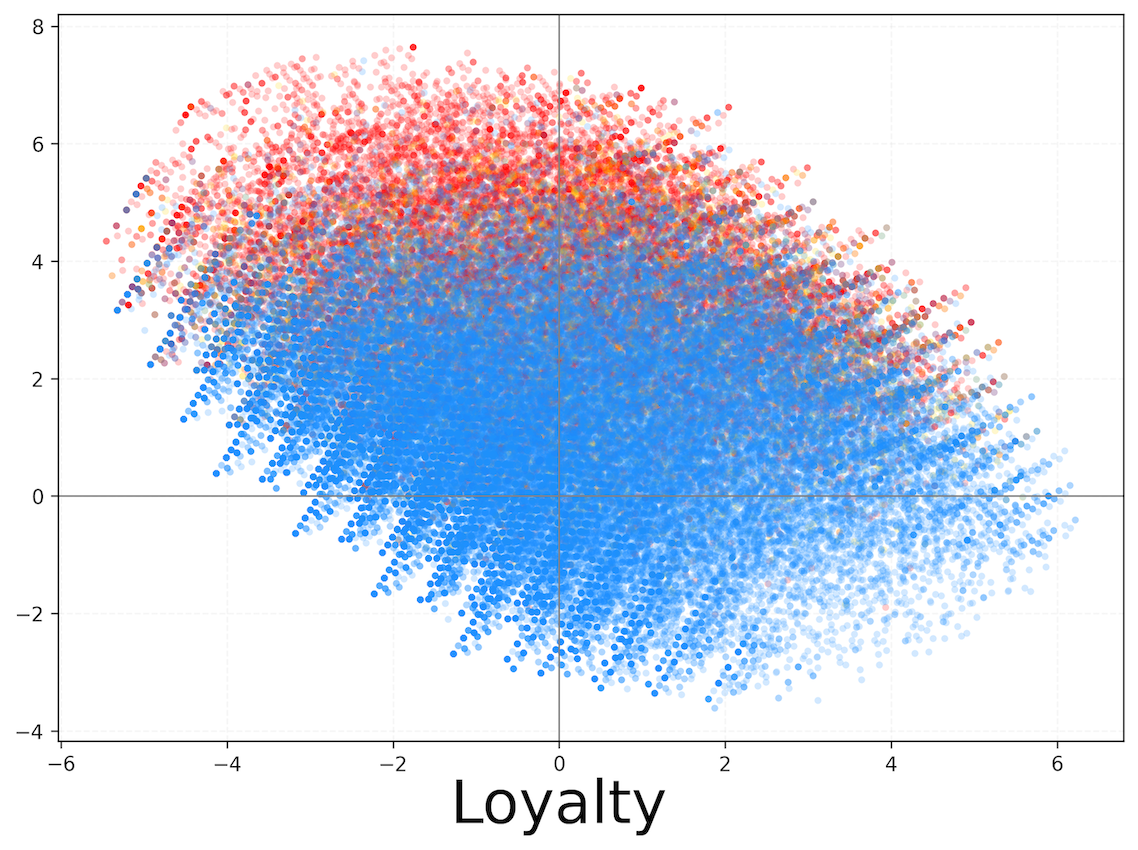}
        \caption{}
    \end{subfigure}\hfill
    \begin{subfigure}{0.25\textwidth}
        \centering
        \includegraphics[width=1\linewidth]{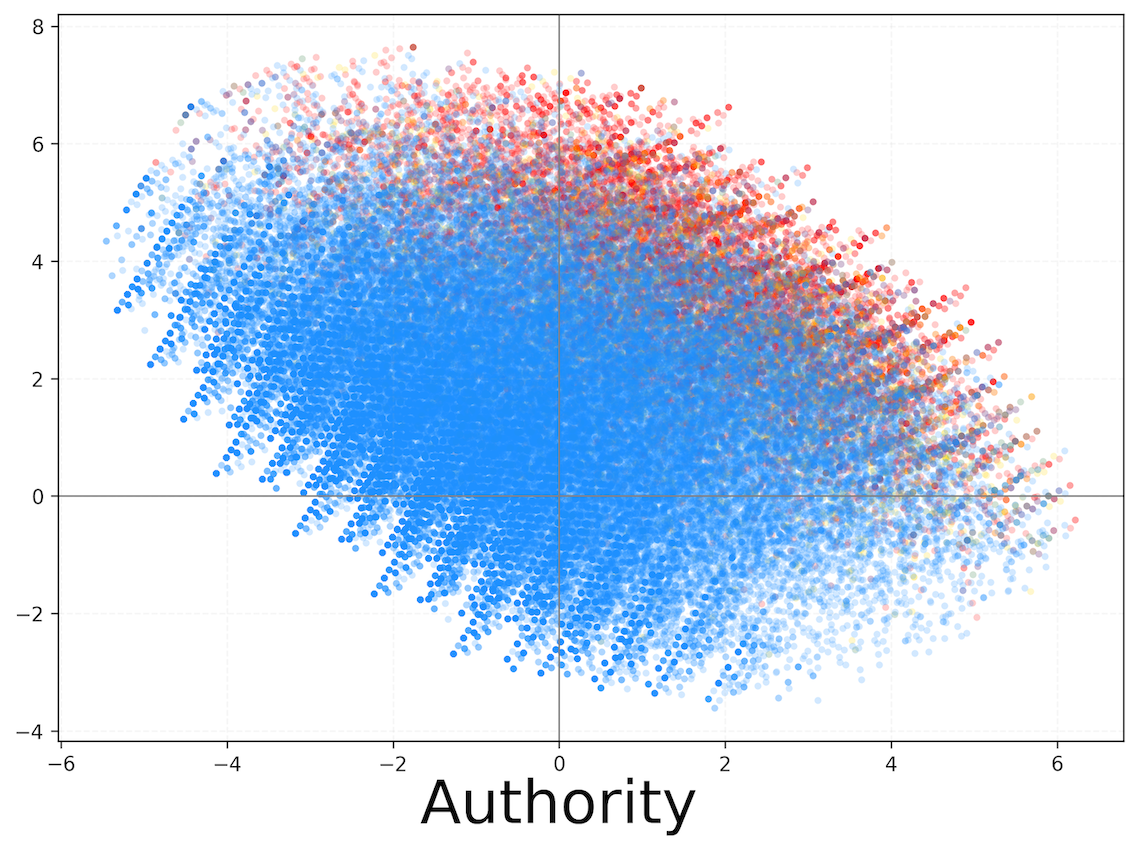}
        \caption{}
    \end{subfigure}\hfill
    \begin{subfigure}{0.25\textwidth}
        \centering
        \includegraphics[width=1\linewidth]{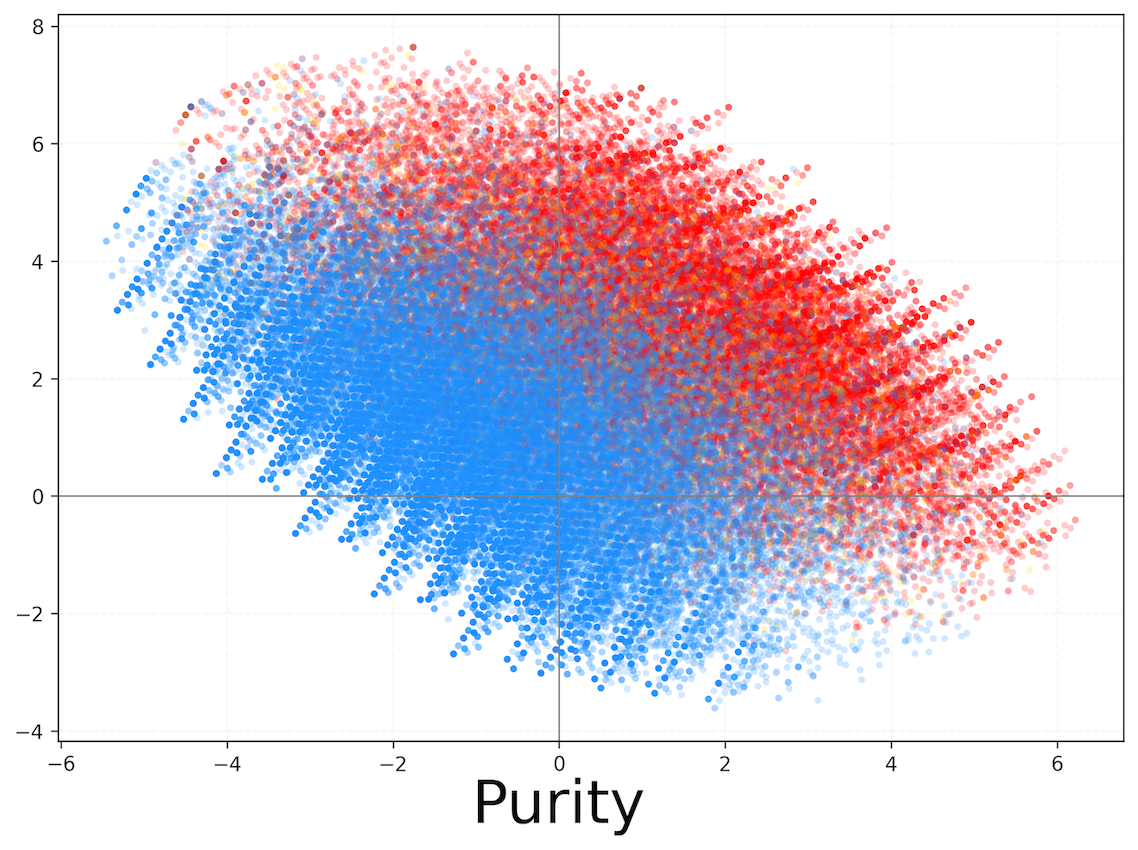}
        \caption{}
    \end{subfigure}

    \caption{Personas projected onto the IW map, stratified by moral foundation score levels.
 Colors indicate the moral-score group: red for scores $<$ 3, yellow for scores $=$ 3, and blue for scores $>$ 3.}
    \label{fig:iw-mfq}
\end{figure*}

\begin{figure*}[ht!]
    \centering

    \begin{subfigure}{0.25\textwidth}
        \centering
        \includegraphics[width=1\linewidth]{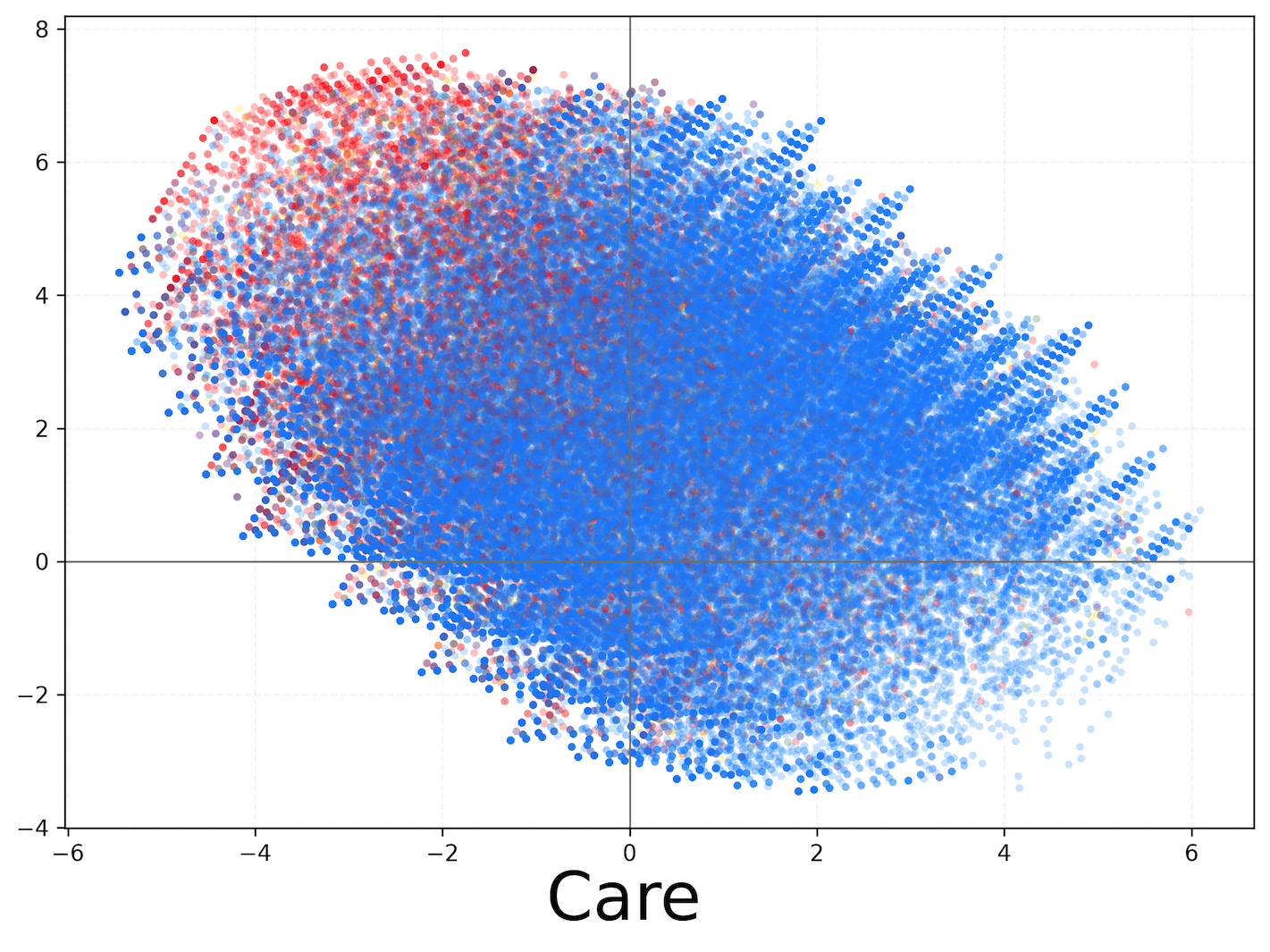}
        \caption{}
    \end{subfigure}\hfill
    \begin{subfigure}{0.25\textwidth}
        \centering
        \includegraphics[width=1\linewidth]{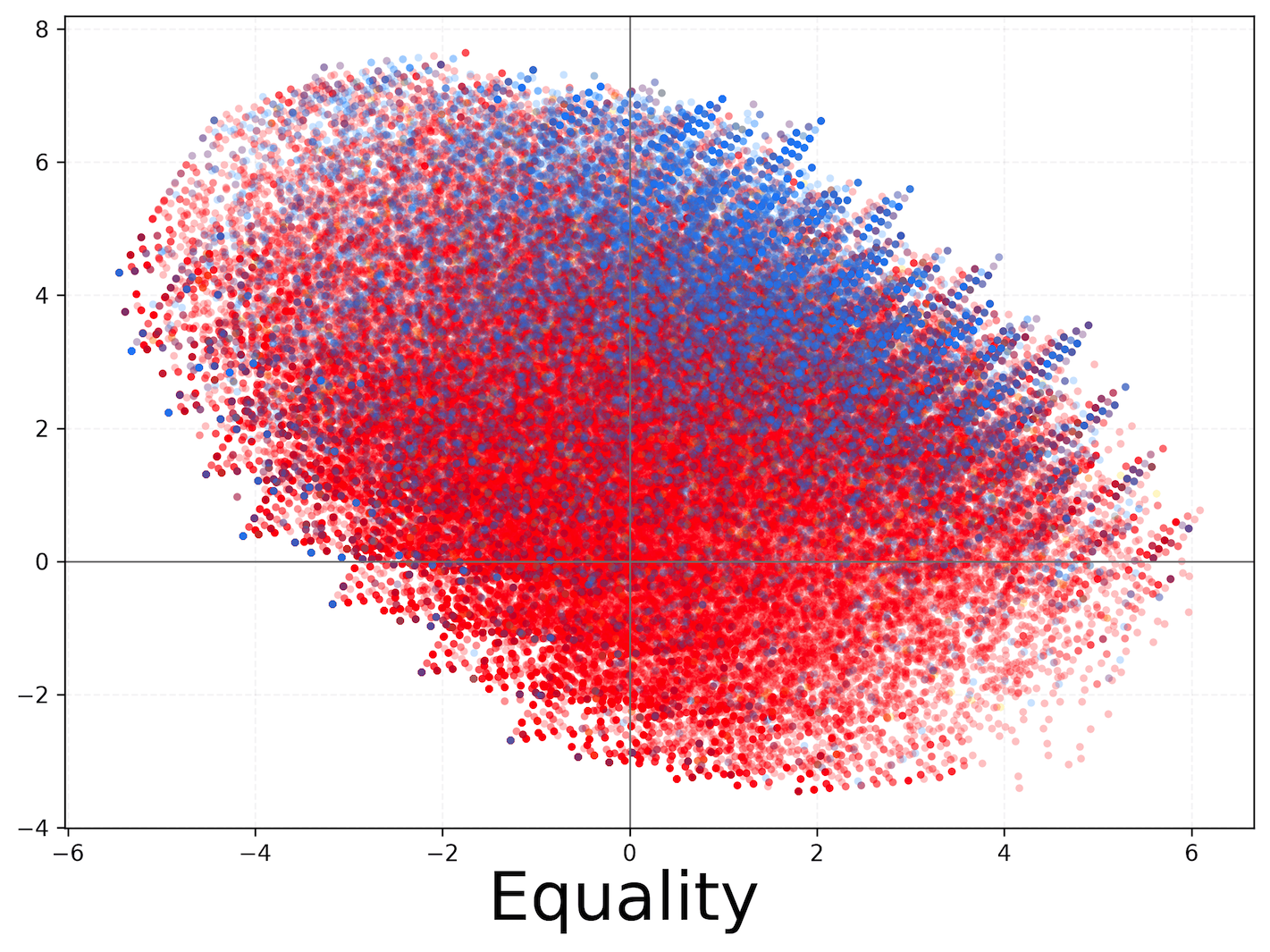}
        \caption{}
    \end{subfigure}\hfill
    \begin{subfigure}{0.25\textwidth}
        \centering
        \includegraphics[width=1\linewidth]{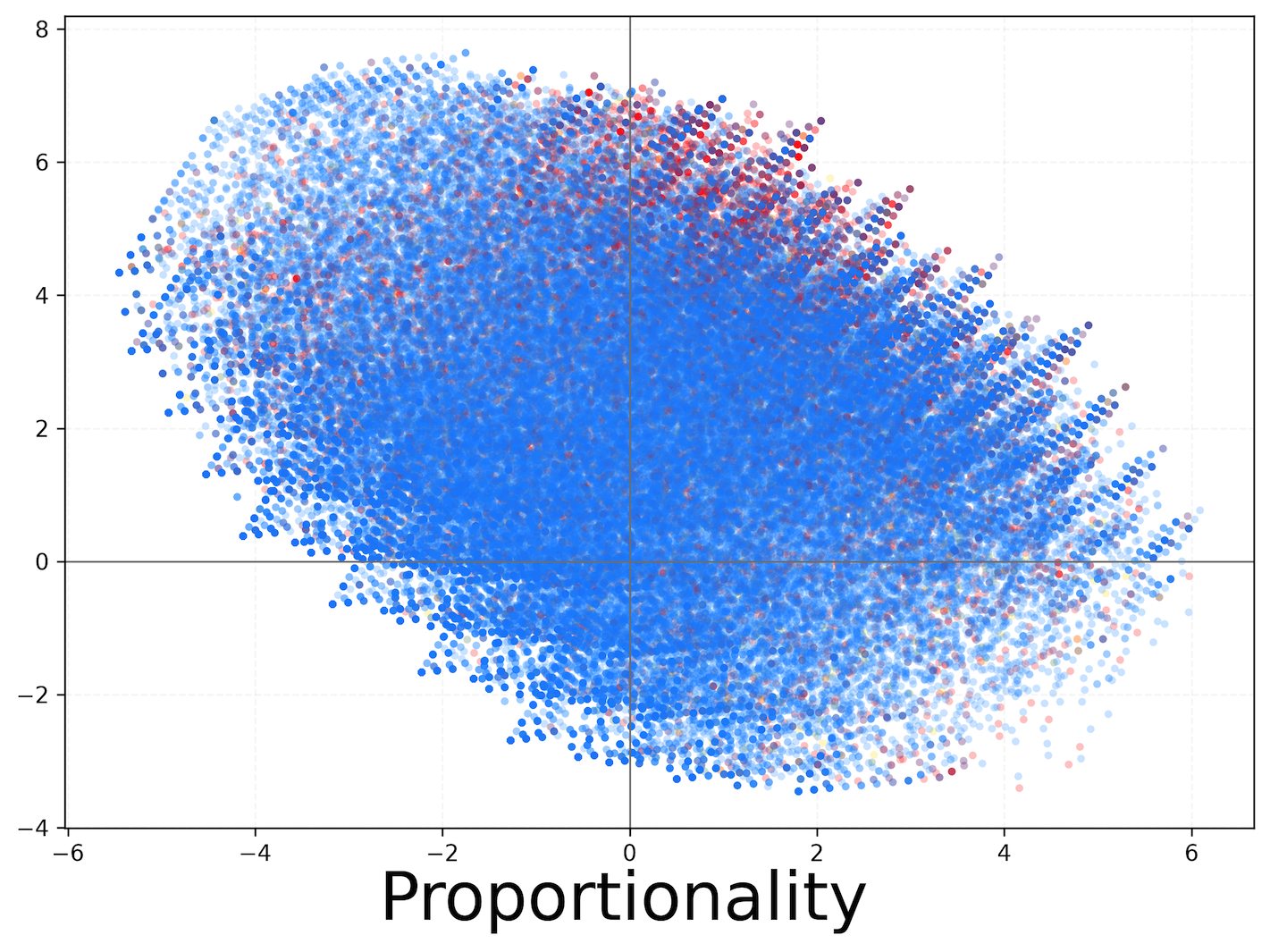}
        \caption{}
    \end{subfigure}


    \begin{subfigure}{0.25\textwidth}
        \centering
        \includegraphics[width=1\linewidth]{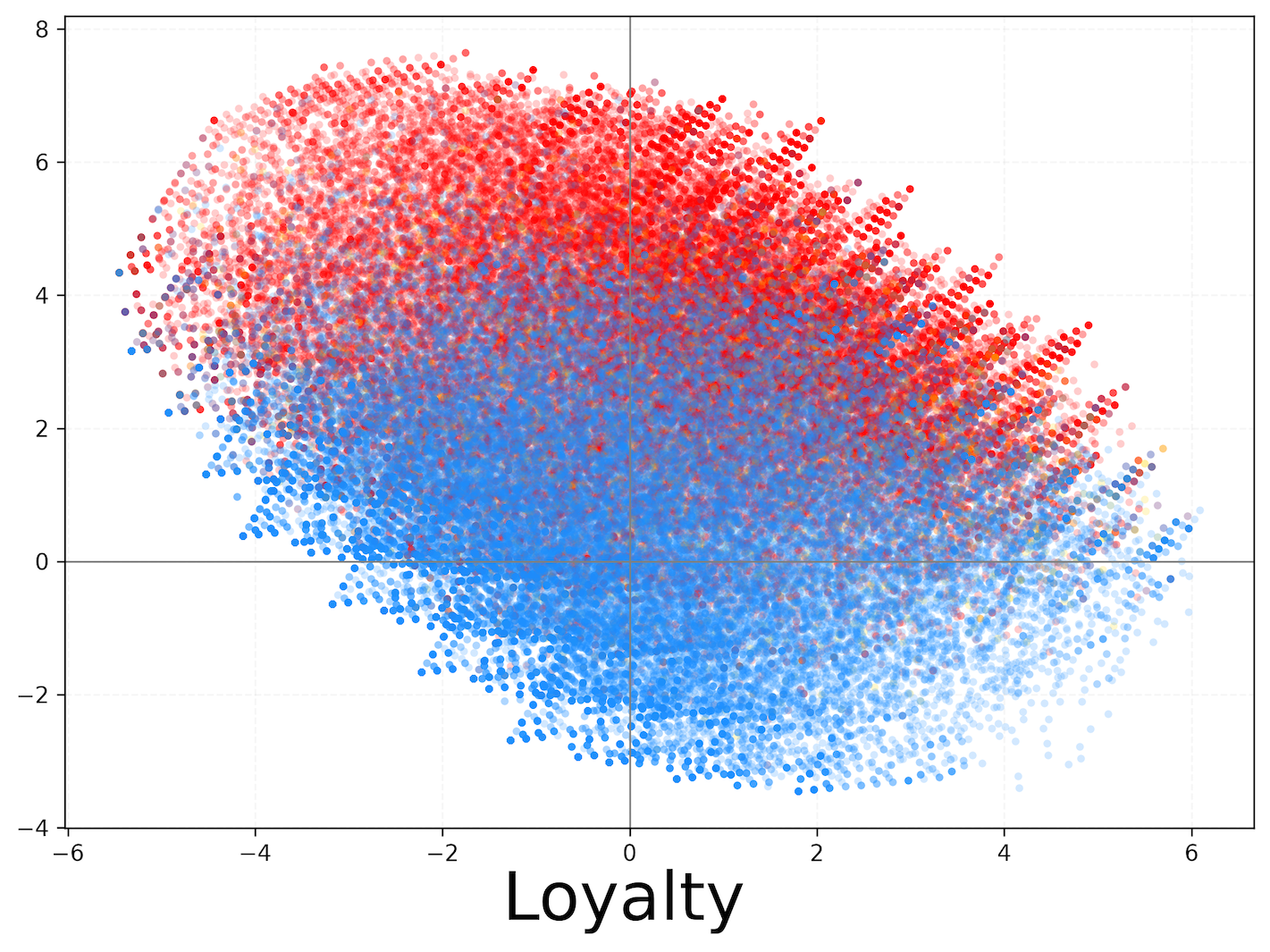}
        \caption{}
    \end{subfigure}\hfill
    \begin{subfigure}{0.25\textwidth}
        \centering
        \includegraphics[width=1\linewidth]{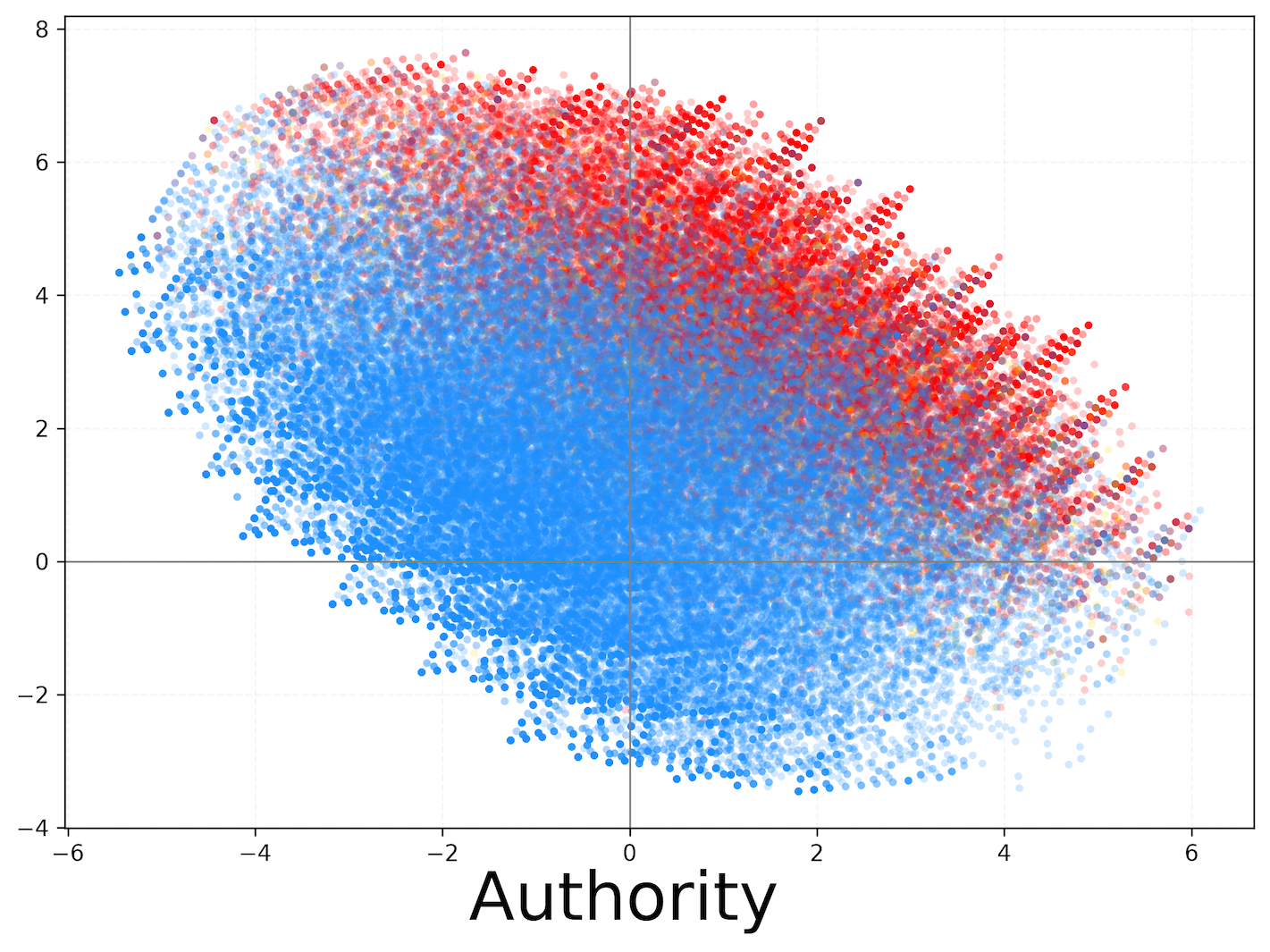}
        \caption{}
    \end{subfigure}\hfill
    \begin{subfigure}{0.25\textwidth}
        \centering
        \includegraphics[width=1\linewidth]{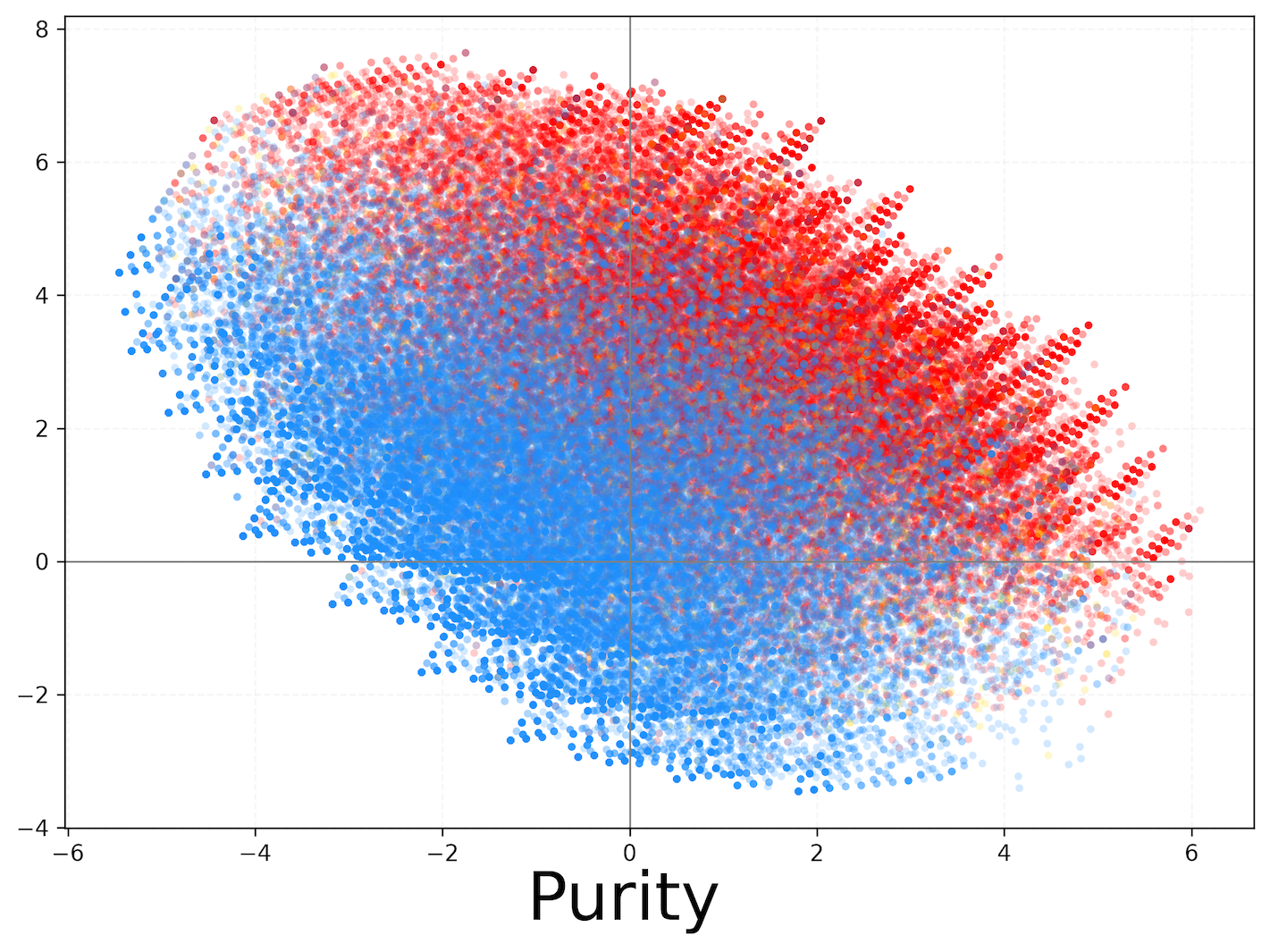}
        \caption{}
    \end{subfigure}

    \caption{\textsf{Qwen3.5-9B} personas projected onto the IW map, stratified by moral foundation score levels.
 Colors indicate the moral-score group: red for scores $<$ 3, yellow for scores $=$ 3, and blue for scores $>$ 3.}
    \label{fig:QWEN-iw-mfq}
\end{figure*}

\begin{figure*}[ht!]
    \centering

    \begin{subfigure}{0.25\textwidth}
        \centering
        \includegraphics[width=1\linewidth]{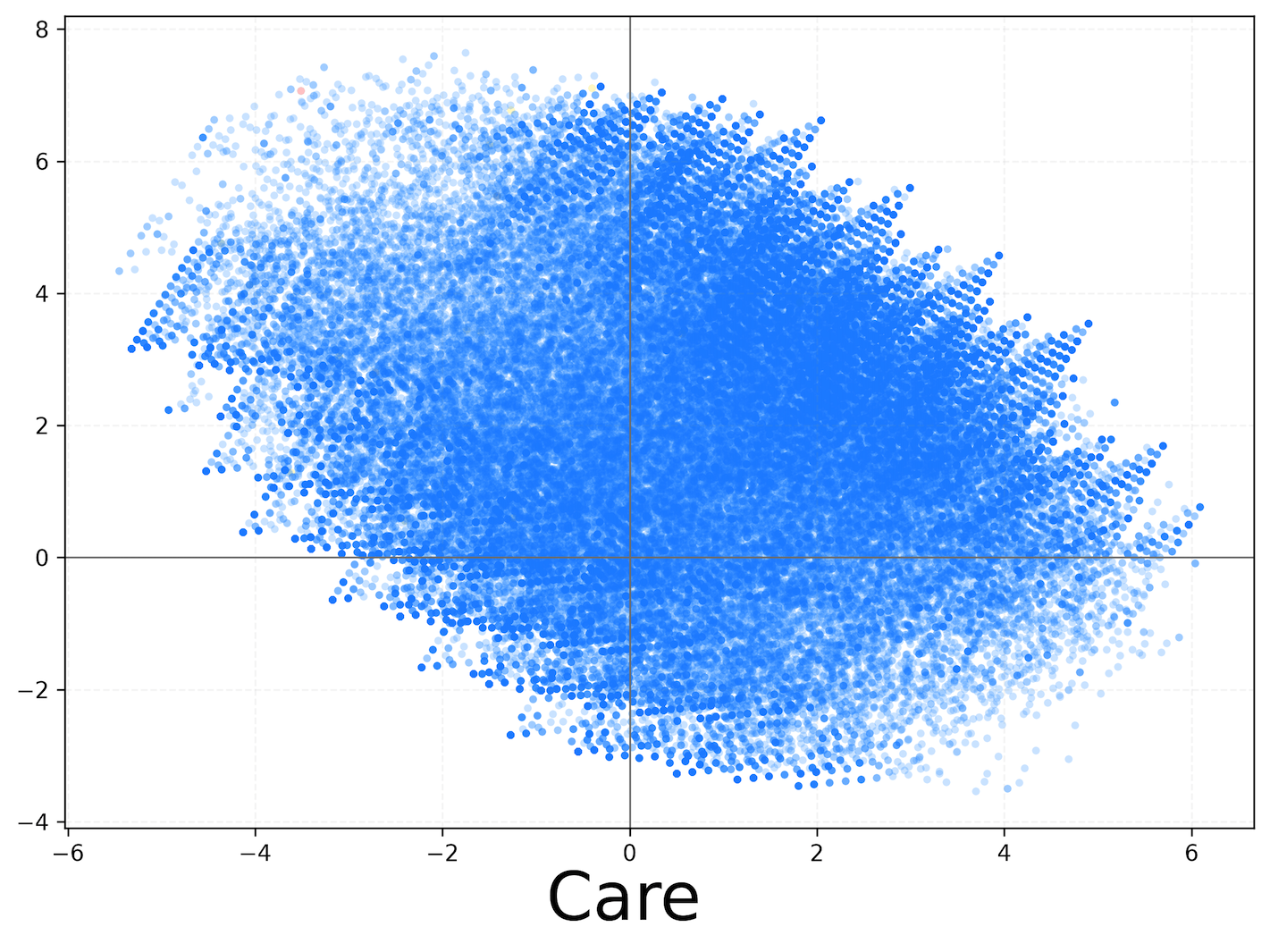}
        \caption{}
    \end{subfigure}\hfill
    \begin{subfigure}{0.25\textwidth}
        \centering
        \includegraphics[width=1\linewidth]{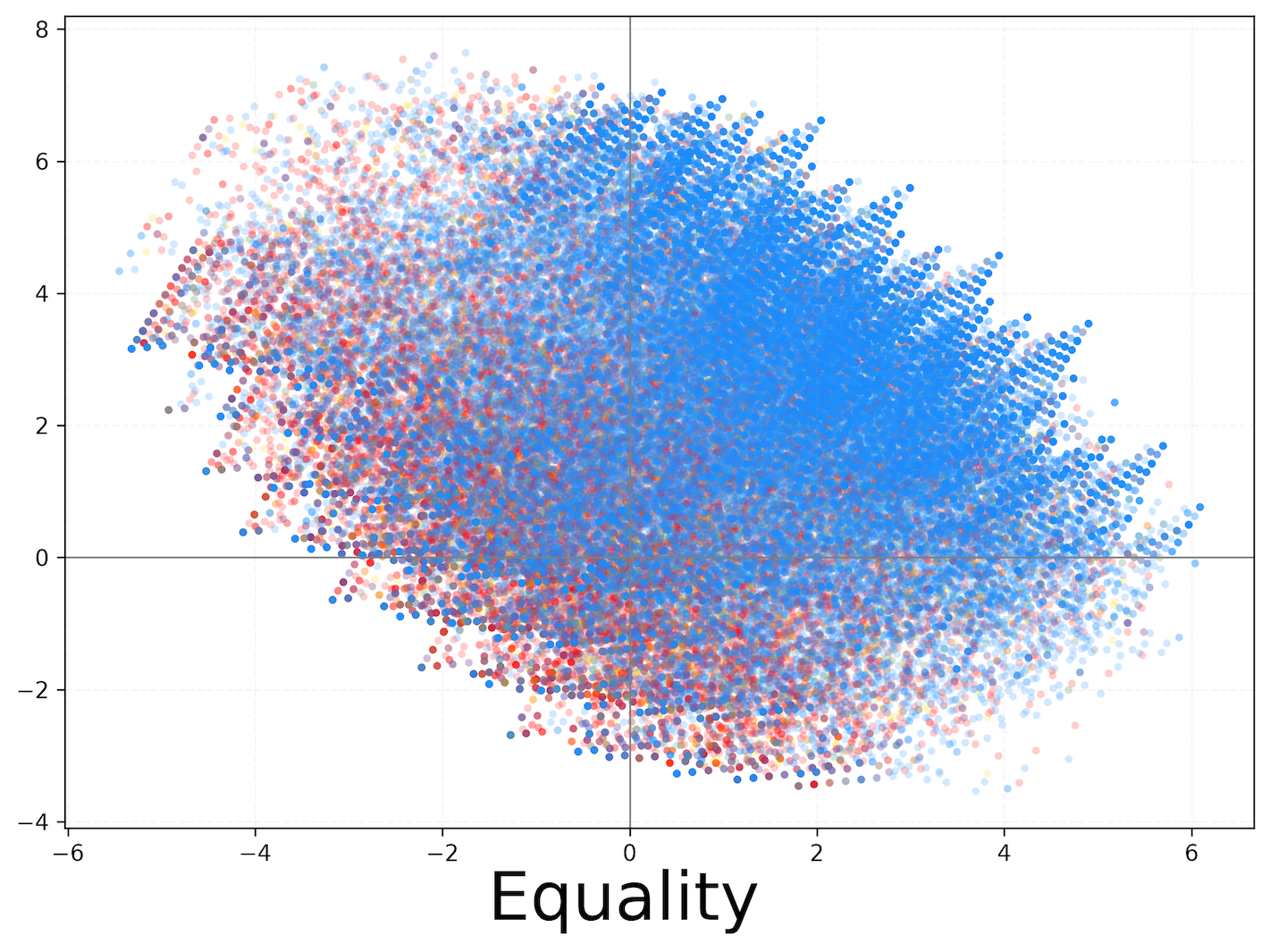}
        \caption{}
    \end{subfigure}\hfill
    \begin{subfigure}{0.25\textwidth}
        \centering
        \includegraphics[width=1\linewidth]{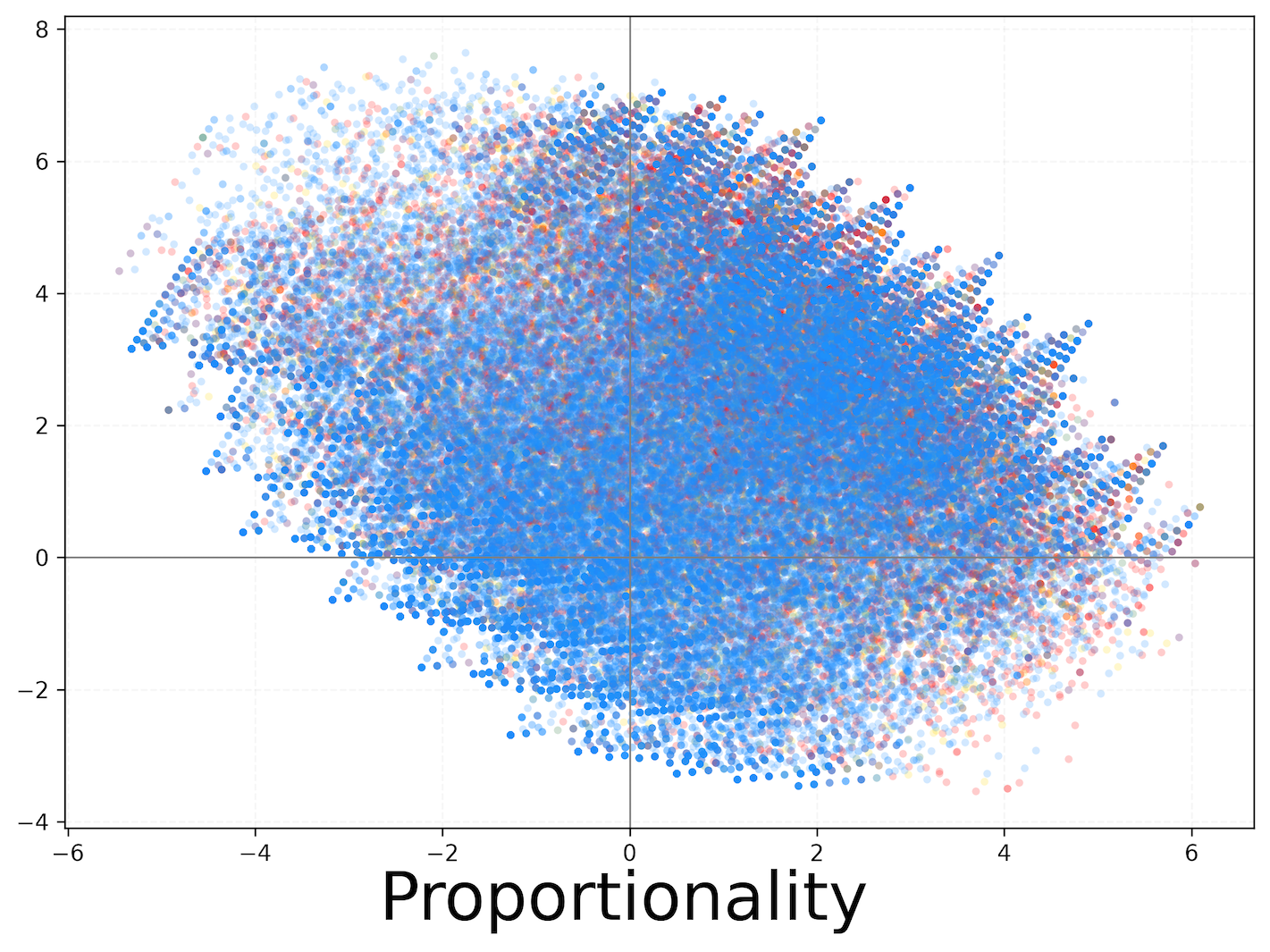}
        \caption{}
    \end{subfigure}


    \begin{subfigure}{0.25\textwidth}
        \centering
        \includegraphics[width=1\linewidth]{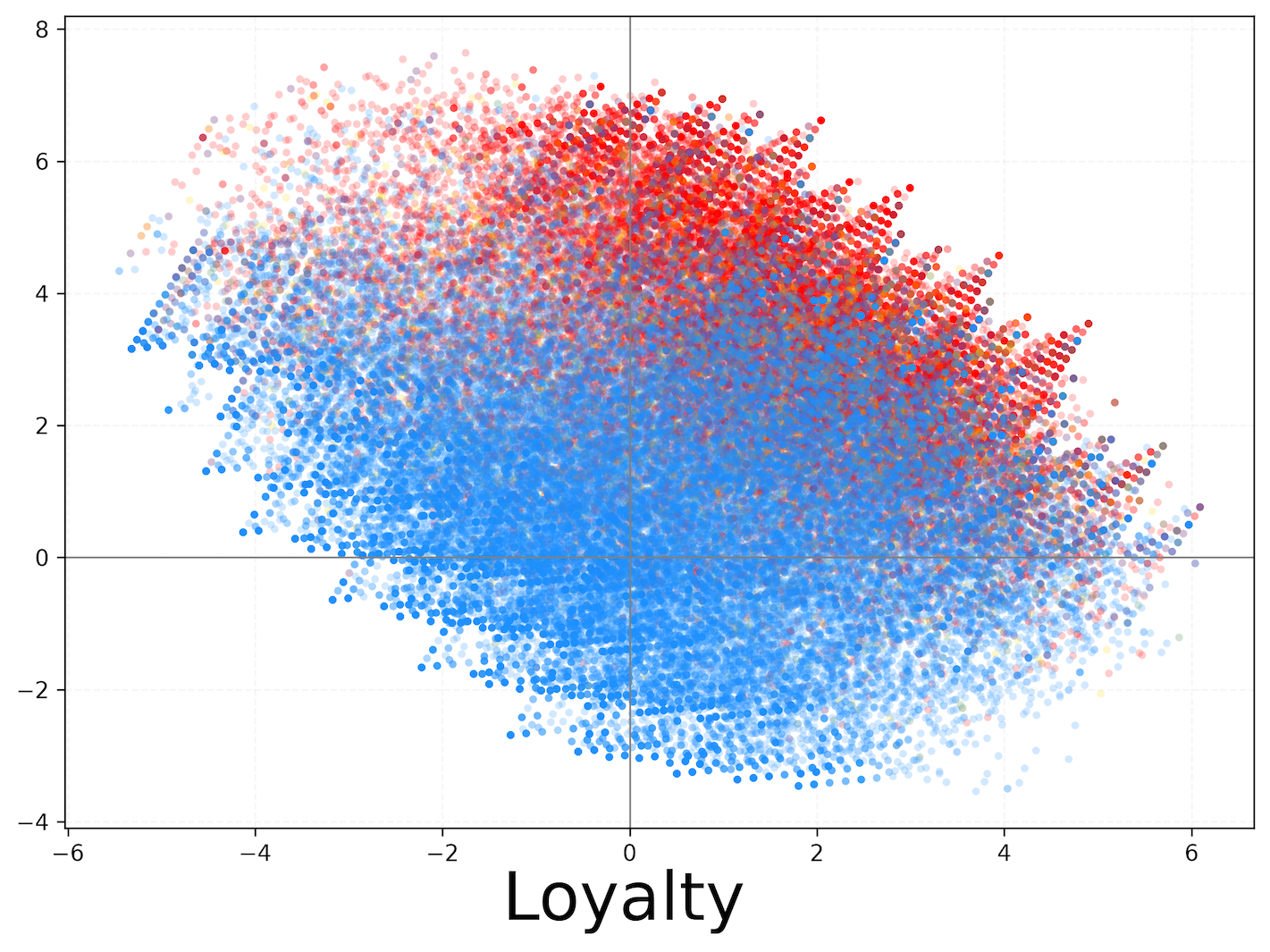}
        \caption{}
    \end{subfigure}\hfill
    \begin{subfigure}{0.25\textwidth}
        \centering
        \includegraphics[width=1\linewidth]{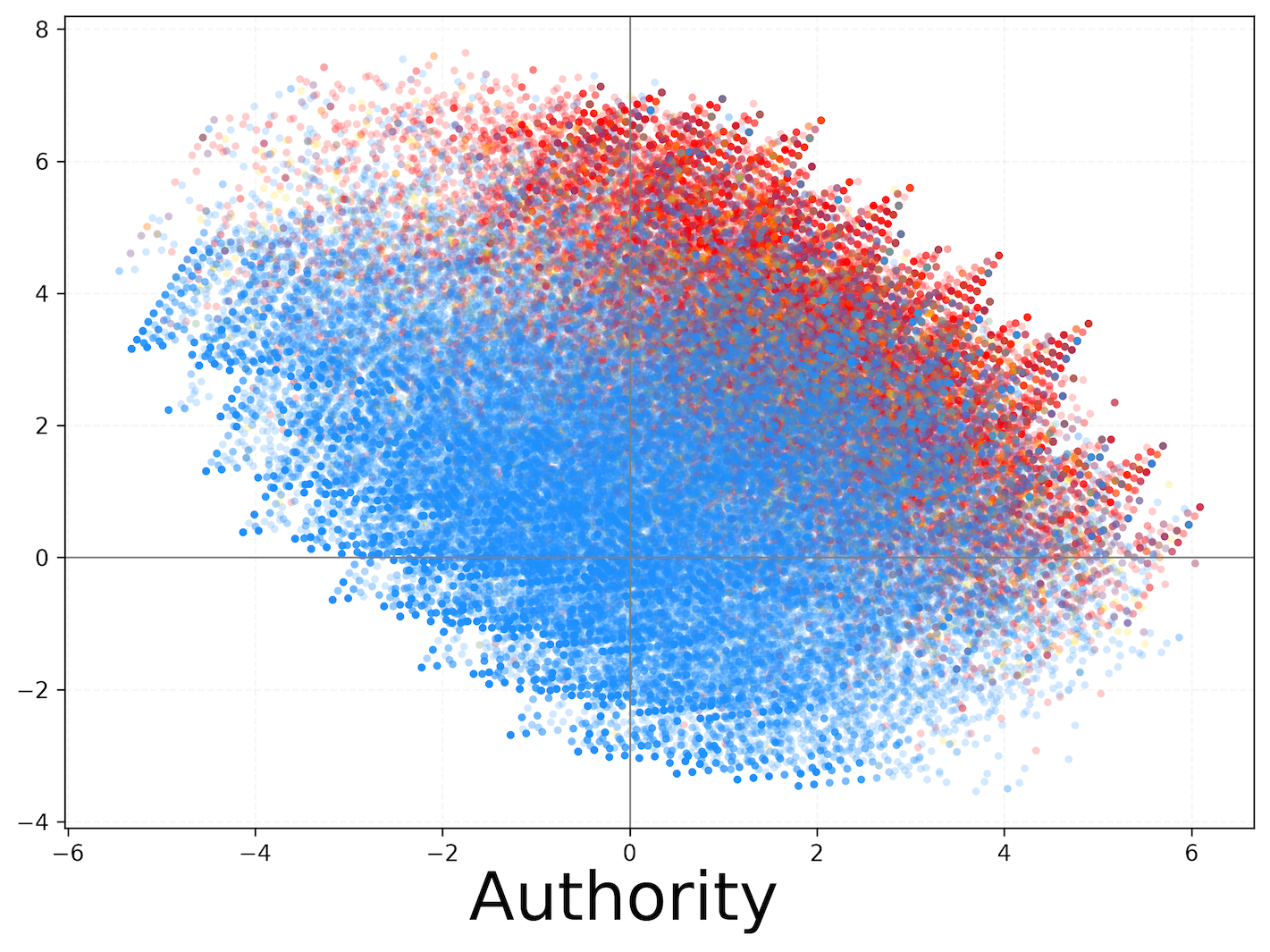}
        \caption{}
    \end{subfigure}\hfill
    \begin{subfigure}{0.25\textwidth}
        \centering
        \includegraphics[width=1\linewidth]{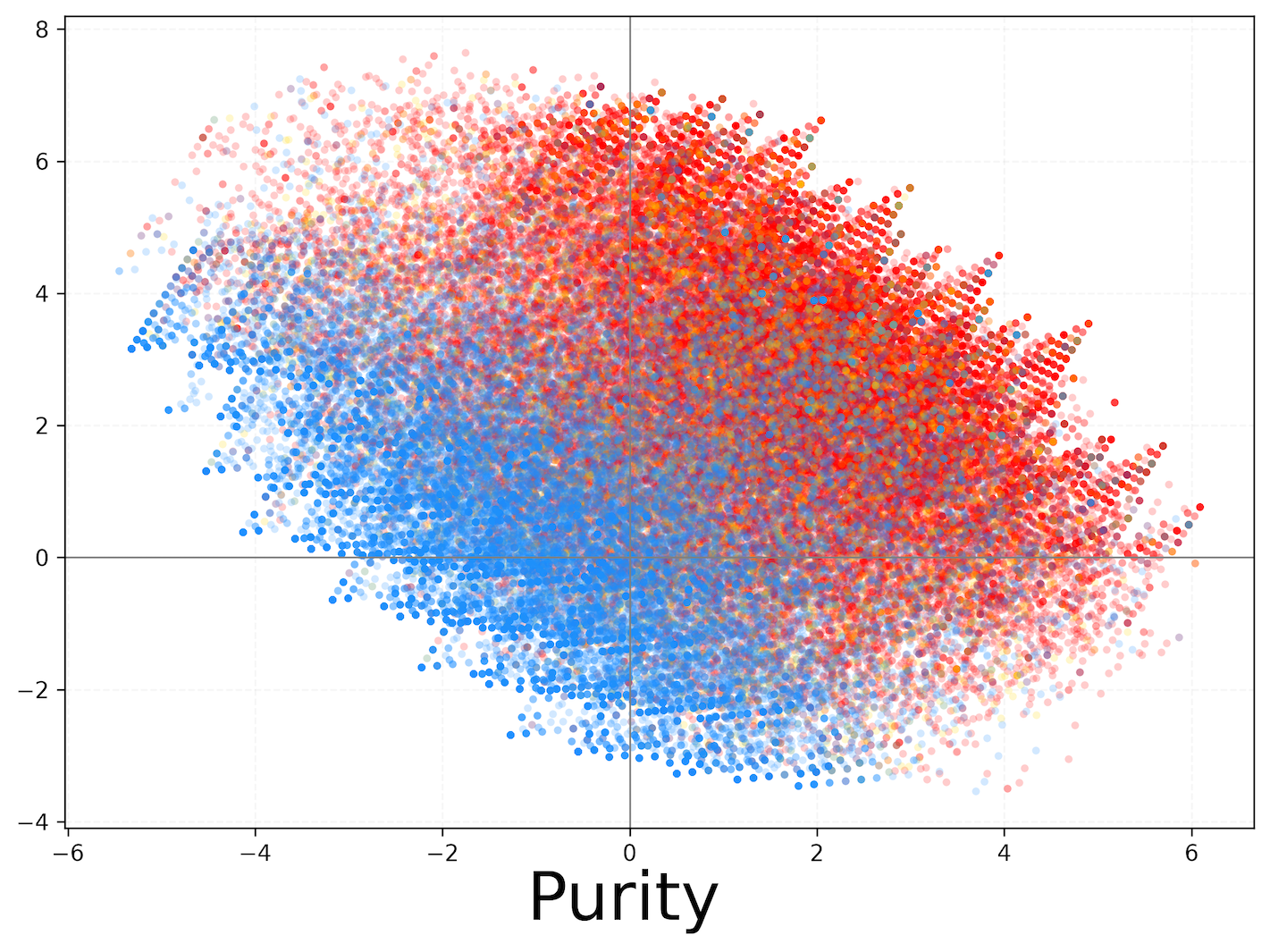}
        \caption{}
    \end{subfigure}

    \caption{\textsf{Llama3.1-8B-Instruct} personas projected onto the IW map, stratified by moral foundation score levels.
 Colors indicate the moral-score group: red for scores $<$ 3, yellow for scores $=$ 3, and blue for scores $>$ 3.}
    \label{fig:LLAMA-iw-mfq}
\end{figure*}

The moral-score maps for \textsf{Qwen3.5-9B} and \textsf{Llama3.1-8B-Instruct} in Figures~\ref{fig:QWEN-iw-mfq} and~\ref{fig:LLAMA-iw-mfq}, compared with the \textsf{gpt-oss-20B} results in Figure~\ref{fig:iw-mfq}, show foundation-specific spatial organization across all three models. The clearest and most consistent spatial differentiation concerns Loyalty, Authority, and Purity, whose score groups tend to occupy different regions of the IW space. By contrast, Care, Equality, and Proportionality show substantially weaker spatial differentiation: their scores tend to remain predominantly on one side of the neutral threshold or, when both score regimes are present, to be more diffusely intermixed rather than concentrated in distinct regions. Thus, the results support the broader distinction between binding and individualizing foundations without implying identical moral-score distributions across models: the individual foundations appear comparatively culturally invariant, whereas the binding foundations exhibit clearer culture-related spatial variation.

The stability results in Tables~\ref{tab:selection_stability}, \ref{tab:QWEN_selection_stability}, and \ref{tab:LLAMA_selection_stability} identify several recurring culture--morality relations. Materialism orientation is consistently selected for Equality and Proportionality, national pride for Loyalty, and religiosity for Purity. Authority also exhibits a substantial shared core composed of child-rearing values, gender equality, national pride, and religiosity; tolerance diversity is additionally stable for \textsf{gpt-oss-20B} and \textsf{Qwen3.5-9B}, but not for \textsf{Llama3.1-8B-Instruct}. These recurring variables indicate cross-model regularities at the level of foundation-specific cultural anchors.  

At the same time, the complete selected sets remain model-dependent. This variation is expected because the MFQ-2-based moral representations used as targets by the greedy procedure are themselves model-specific and differ in both level and shape, as shown by the blue curves in the corresponding figures. Core-set stability is therefore assessed within each model across the 50 independent runs. \textsf{Qwen3.5-9B} generally produces compact and highly stable sets, whereas \textsf{Llama3.1-8B-Instruct} selects broader combinations of cultural variables for several foundations. Care exhibits the largest cross-model difference: \textsf{Qwen3.5-9B} consistently selects religiosity alone, while \textsf{gpt-oss-20B} and \textsf{Llama3.1-8B-Instruct} rely on wider multivariable sets.  

\begin{figure*}[ht!]
    \centering

    \begin{subfigure}{0.3\textwidth}
        \centering
        \includegraphics[width=0.85\linewidth]{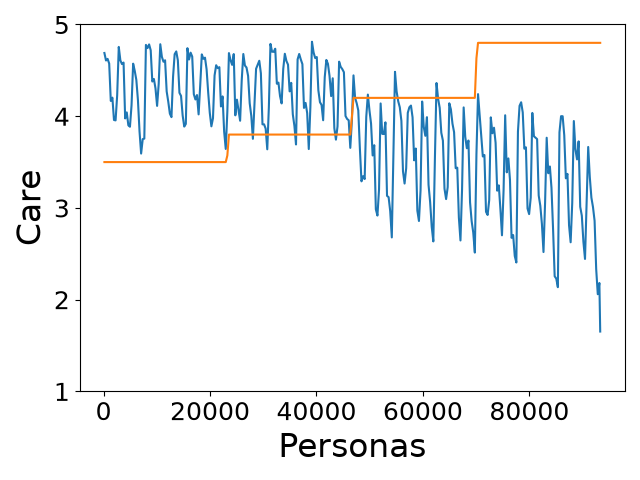}
        \caption{}
    \end{subfigure}\hfill
    \begin{subfigure}{0.3\textwidth}
        \centering
        \includegraphics[width=0.85\linewidth]{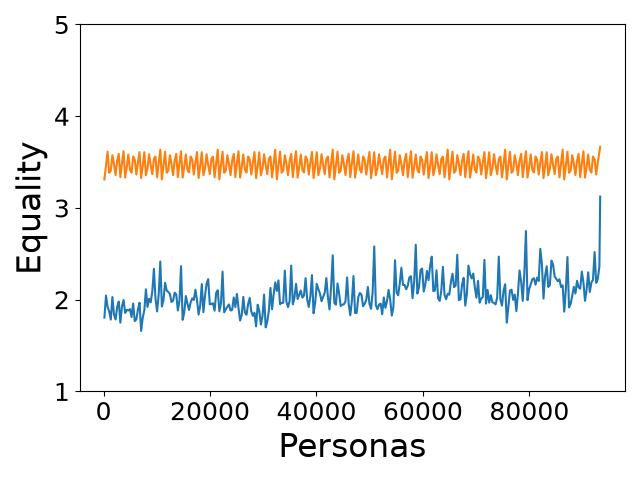}
        \caption{}
    \end{subfigure}\hfill
    \begin{subfigure}{0.3\textwidth}
        \centering
        \includegraphics[width=0.85\linewidth]{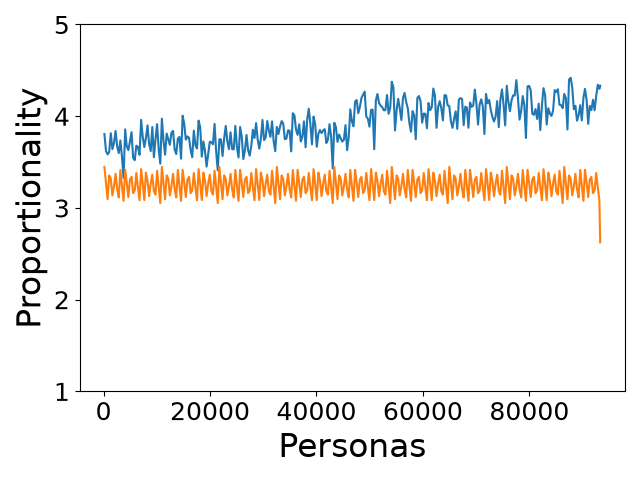}
        \caption{}
    \end{subfigure}


    \begin{subfigure}{0.3\textwidth}
        \centering
        \includegraphics[width=0.85\linewidth]{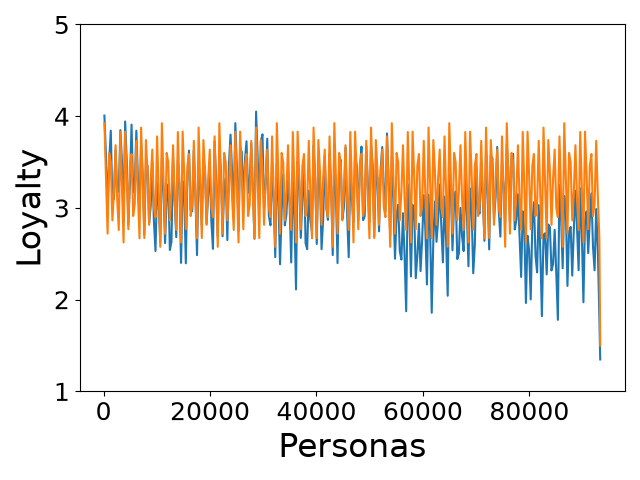}
        \caption{}
    \end{subfigure}\hfill
    \begin{subfigure}{0.3\textwidth}
        \centering
        \includegraphics[width=0.85\linewidth]{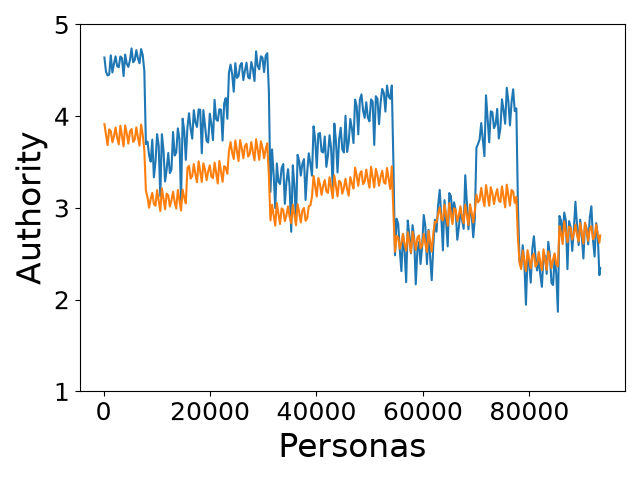}
        \caption{}
    \end{subfigure}\hfill
    \begin{subfigure}{0.3\textwidth}
        \centering
        \includegraphics[width=0.85\linewidth]{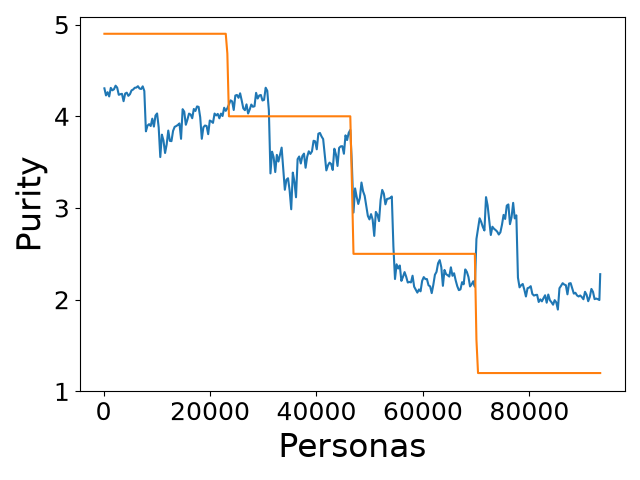}
        \caption{}
    \end{subfigure}

    \caption{\textsf{Qwen3.5-9B} Comparison between MFQ-2–based (blue) and mapping-based (orange) moral foundation scores. Mapping-based scores are obtained using a greedy-selected subset of cultural variables. Personas are ordered along the x-axis; the y-axis reports foundation scores. To avoid cluttering, lines of both score types  are smoothed by averaging scores within consecutive bins of 300 personas (according to the ordering fixed through sorting  by cultural variable then by admissible values).}
    \label{fig:QWEN-mfq2-greedy}
\end{figure*}

\begin{figure*}[ht!]
    \centering

    \begin{subfigure}{0.3\textwidth}
        \centering
        \includegraphics[width=0.85\linewidth]{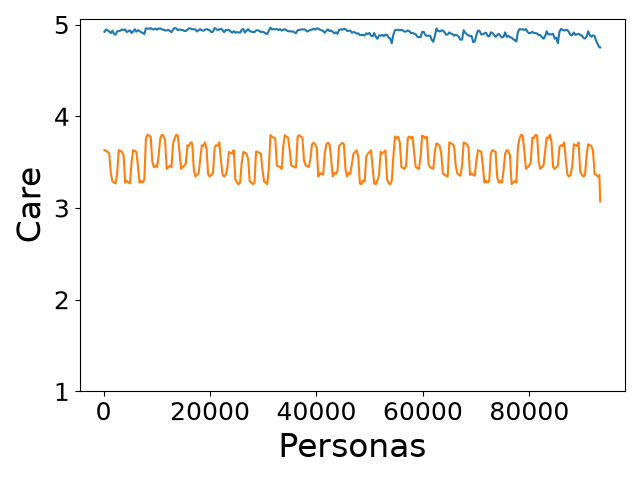}
        \caption{}
    \end{subfigure}\hfill
    \begin{subfigure}{0.3\textwidth}
        \centering
        \includegraphics[width=0.85\linewidth]{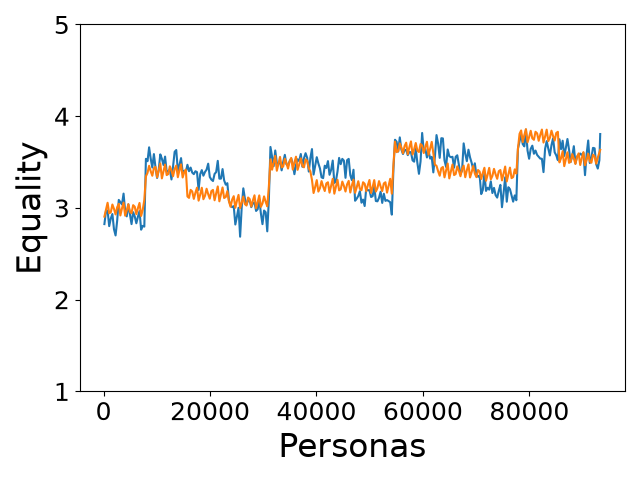}
        \caption{}
    \end{subfigure}\hfill
    \begin{subfigure}{0.3\textwidth}
        \centering
        \includegraphics[width=0.85\linewidth]{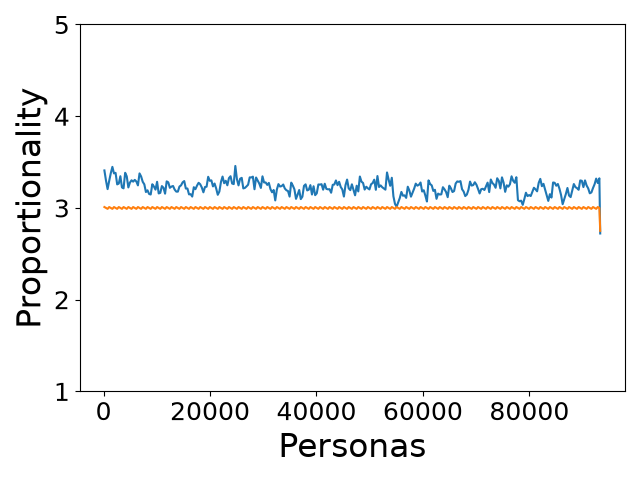}
        \caption{}
    \end{subfigure}


    \begin{subfigure}{0.3\textwidth}
        \centering
        \includegraphics[width=0.85\linewidth]{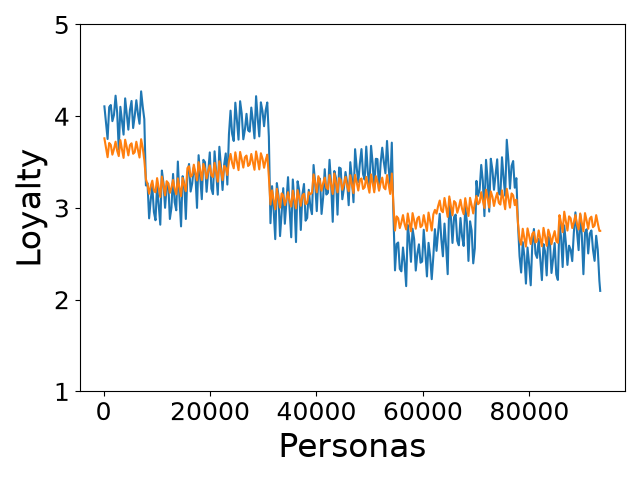}
        \caption{}
    \end{subfigure}\hfill
    \begin{subfigure}{0.3\textwidth}
        \centering
        \includegraphics[width=0.85\linewidth]{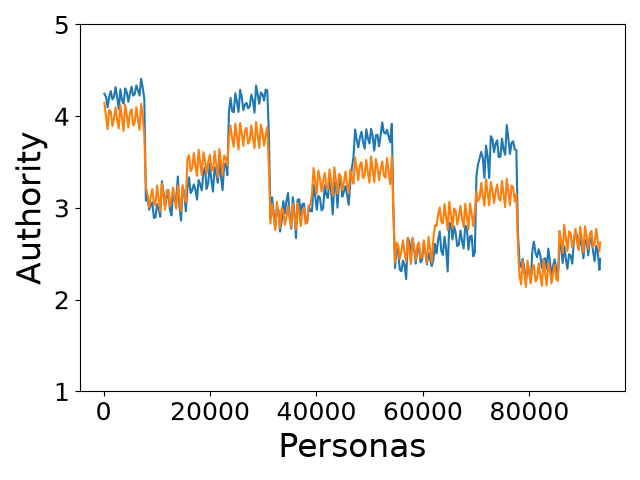}
        \caption{}
    \end{subfigure}\hfill
    \begin{subfigure}{0.3\textwidth}
        \centering
        \includegraphics[width=0.85\linewidth]{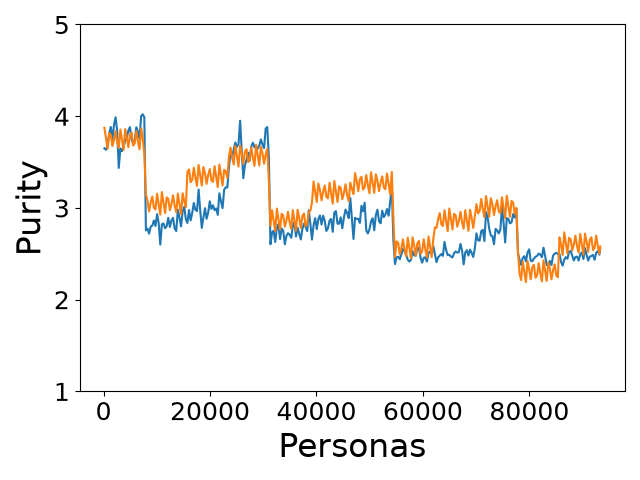}
        \caption{}
    \end{subfigure}

    \caption{Comparison between MFQ-2–based (blue) and mapping-based (orange) moral foundation scores for \textsf{Llama3.1-8B-Instruct}. Mapping-based scores are obtained using a greedy-selected subset of cultural variables. Personas are ordered along the x-axis; the y-axis reports foundation scores. To avoid cluttering, lines of both score types  are smoothed by averaging scores within consecutive bins of 300 personas (according to the ordering fixed through sorting  by cultural variable then by admissible values).}
    \label{fig:LLAMA-mfq2-greedy}
\end{figure*}

Figures~\ref{fig:QWEN-mfq2-greedy} and \ref{fig:LLAMA-mfq2-greedy}, together with the corresponding comparisons based on all ten cultural variables, compare the MFQ-2 representations with the mapping-based scores before and after foundation-specific variable selection. Compared with the mapping-based representation computed from all ten cultural variables, the selected subsets provide closer qualitative tracking for some foundations, with the most recurrent improvements visible for Authority and Purity. However, residual differences in level and shape remain, particularly for Care.

\begin{table*}[h]
    \centering
    \scalebox{0.88}{
    \begin{tabular}{lrrrr}
\toprule
\textbf{Group (Continent \textbar Settlement \textbar Education)} & \# \textbf{personas} & \textbf{$1-\overline{EMD}$} & \textbf{\%($\mathit{EMD}<$0.4)} & \textbf{\%($\mathit{EMD}<$0.2)} \\
\midrule
Europe \textbar Urban \textbar Tertiary & 32716 & 0.803 & 94.4 & 52.8 \\
North America \textbar Urban \textbar Tertiary & 14984 & 0.832 & 97.2 & 69.4 \\
Asia \textbar Urban \textbar Tertiary & 12868 & 0.825 & 97.2 & 66.7 \\
South America \textbar Urban \textbar Tertiary & 4924 & 0.799 & 94.4 & 55.6 \\
North America \textbar Rural \textbar Tertiary & 4586 & 0.817 & 91.7 & 61.1 \\
Europe \textbar Rural \textbar Tertiary & 3364 & 0.828 & 97.2 & 63.9 \\
North America \textbar Rural \textbar Upper to Post Secondary & 2907 & 0.783 & 86.1 & 52.8 \\
Asia \textbar Rural \textbar Tertiary & 1998 & 0.784 & 88.9 & 52.8 \\
Africa \textbar Urban \textbar Tertiary & 1772 & 0.784 & 91.7 & 55.6 \\
Europe \textbar Urban \textbar Upper to Post Secondary & 1523 & 0.780 & 83.3 & 55.6 \\
Europe \textbar Rural \textbar Upper to Post Secondary & 1237 & 0.803 & 97.2 & 55.6 \\
Asia \textbar Urban \textbar Upper to Post Secondary & 1228 & 0.813 & 94.4 & 63.9 \\
North America \textbar Urban \textbar Upper to Post Secondary & 1035 & 0.788 & 91.7 & 55.6 \\
North America \textbar Rural \textbar Lower Secondary & 910 & 0.775 & 83.3 & 61.1 \\
Asia \textbar Rural \textbar Upper to Post Secondary & 859 & 0.808 & 88.9 & 58.3 \\
Asia \textbar Rural \textbar Lower Secondary & 843 & 0.777 & 88.9 & 55.6 \\
South America \textbar Rural \textbar Tertiary & 595 & 0.790 & 91.7 & 61.1 \\
Europe \textbar Rural \textbar Lower Secondary & 524 & 0.805 & 100.0 & 58.3 \\
Oceania \textbar Urban \textbar Tertiary & 513 & 0.796 & 91.7 & 55.6 \\
Asia \textbar Rural \textbar Primary or No Education & 492 & 0.786 & 83.3 & 50.0 \\
Asia \textbar Urban \textbar Lower Secondary & 334 & 0.803 & 94.4 & 61.1 \\
South America \textbar Urban \textbar Upper to Post Secondary & 313 & 0.782 & 94.4 & 44.4 \\
South America \textbar Rural \textbar Upper to Post Secondary & 242 & 0.823 & 91.7 & 69.4 \\
Africa \textbar Urban \textbar Upper to Post Secondary & 223 & 0.794 & 91.7 & 58.3 \\
Europe \textbar Urban \textbar Lower Secondary & 214 & 0.786 & 94.4 & 55.6 \\
Africa \textbar Rural \textbar Tertiary & 211 & 0.787 & 88.9 & 44.4 \\
North America \textbar Rural \textbar Primary or No Education & 192 & 0.774 & 83.3 & 58.3 \\
North America \textbar Urban \textbar Lower Secondary & 137 & 0.762 & 86.1 & 47.2 \\
South America \textbar Rural \textbar Lower Secondary & 135 & 0.774 & 86.1 & 61.1 \\
Europe \textbar Rural \textbar Primary or No Education & 134 & 0.801 & 91.7 & 69.4 \\
Africa \textbar Rural \textbar Upper to Post Secondary & 123 & 0.825 & 94.4 & 66.7 \\
Africa \textbar Rural \textbar Lower Secondary & 112 & 0.809 & 97.2 & 66.7 \\
Africa \textbar Rural \textbar Primary or No Education & 92 & 0.795 & 94.4 & 50.0 \\
South America \textbar Rural \textbar Primary or No Education & 85 & 0.774 & 75.0 & 55.6 \\
South America \textbar Urban \textbar Lower Secondary & 63 & 0.806 & 94.4 & 69.4 \\
Africa \textbar Urban \textbar Lower Secondary & 59 & 0.774 & 83.3 & 55.6 \\
Asia \textbar Urban \textbar Primary or No Education & 55 & 0.791 & 88.9 & 52.8 \\
Oceania \textbar Rural \textbar Tertiary & 32 & 0.840 & 94.4 & 72.2 \\
Europe \textbar Urban \textbar Primary or No Education & 23 & 0.804 & 83.3 & 66.7 \\
North America \textbar Urban \textbar Primary or No Education & 15 & 0.747 & 86.1 & 30.6 \\
Africa \textbar Urban \textbar Primary or No Education & 15 & 0.756 & 88.9 & 36.1 \\
South America \textbar Urban \textbar Primary or No Education & 12 & 0.820 & 97.2 & 66.7 \\
Oceania \textbar Urban \textbar Upper to Post Secondary & 6 & 0.838 & 97.2 & 66.7 \\
Oceania \textbar Rural \textbar Upper to Post Secondary & 4 & 0.774 & 88.9 & 50.0 \\
Oceania \textbar Urban \textbar Primary or No Education & 1 & 0.628 & 55.6 & 27.8 \\
\bottomrule
\end{tabular}
}
    \caption{Complete group-level WVB-Probe alignment for all 45 demographic triples supported by WVB-Probe, sorted by decreasing number of personas.
Score is $1-\overline{EMD}$, where $\overline{EMD}$ is the mean EMD across the 36 probe questions for the group.
\%EMD$<0.4$ and \%EMD$<0.2$ denote the fraction of probe questions whose distance falls below the corresponding threshold.}
    \label{tab:wvb-probe-all-groups}
\end{table*}


\begin{table*}[t]
\centering
\small
\setlength{\tabcolsep}{2pt}
\begin{tabular}{lcp{4cm}p{8cm}}
\toprule
\textbf{Moral Foundation} & $\bar J$ & \textbf{$k$ distribution (\#runs)} & \textbf{Selected variables (support)} \\
\midrule
Care & 0.669 &
$k{=}1$ (10),\ $k{=}6$ (39),\ $k{=}7$ (1) &
child\_rearing\_value (0.80), gender\_equality (0.80), happiness (0.80), materialism\_orientation (0.80), social\_trust (0.80), tolerance\_diversity (0.80), religiosity (0.20), political\_participation (0.02) \\

Equality & 0.738 &
$k{=}1$ (37),\ $k{=}3$ (13) &
materialism\_orientation (1.00), gender\_equality (0.26), tolerance\_diversity (0.26) \\

Proportionality & 1.000 &
$k{=}1$ (50) &
materialism\_orientation (1.00) \\

Loyalty & 1.000 &
$k{=}1$ (50) &
national\_pride (1.00) \\

Authority & 1.000 &
$k{=}5$ (50) &
child\_rearing\_value (1.00), gender\_equality (1.00), national\_pride (1.00), religiosity (1.00), tolerance\_diversity (1.00) \\

Purity & 1.000 &
$k{=}1$ (50) &
religiosity (1.00) \\
\bottomrule
\end{tabular}
\caption{Stability of core-variable selection across 50 independent runs.
$\bar J$ denotes the mean pairwise Jaccard overlap between selected sets.
The $k$ distribution is reported as $k{=}m$ (count), where $m$ is the number of selected variables and the value in parentheses is the number of runs  in which the $m$ variables were selected.}
\label{tab:selection_stability}
\end{table*}

\begin{table*}[h]
\centering
\small
\begin{tabular}{lllrrrrr}
\toprule
\textbf{Moral Foundation} & \textbf{Core variable} & \textbf{Core value} & \textbf{Mean} & \textbf{Median} & \textbf{Q25} & \textbf{Q75} & \textbf{p($\ge$4)} \\
\midrule
Care & child rearing value & independence imagination & 4.639 & 5.000 & 4.670 & 5.000 & 0.894 \\
Care & child rearing value & neutral & 4.529 & 5.000 & 4.000 & 5.000 & 0.850 \\
Care & child rearing value & obedience faith & 4.479 & 5.000 & 4.000 & 5.000 & 0.839 \\
Care & gender equality & traditional & 4.393 & 5.000 & 4.000 & 5.000 & 0.807 \\
Care & gender equality & moderate & 4.567 & 5.000 & 4.170 & 5.000 & 0.869 \\
Care & gender equality & egalitarian & 4.687 & 5.000 & 4.830 & 5.000 & 0.907 \\
Care & happiness & not at all happy & 4.320 & 5.000 & 4.000 & 5.000 & 0.774 \\
Care & happiness & not very happy & 4.479 & 5.000 & 4.000 & 5.000 & 0.835 \\
Care & happiness & rather happy & 4.662 & 5.000 & 4.500 & 5.000 & 0.908 \\
Care & happiness & very happy & 4.736 & 5.000 & 5.000 & 5.000 & 0.927 \\
Care & materialism orientation & materialist & 4.141 & 4.330 & 3.330 & 5.000 & 0.710 \\
Care & materialism orientation & mixed & 4.683 & 5.000 & 4.830 & 5.000 & 0.915 \\
Care & materialism orientation & postmaterialist & 4.823 & 5.000 & 5.000 & 5.000 & 0.958 \\
Care & tolerance diversity & low tolerance & 4.256 & 4.830 & 4.000 & 5.000 & 0.757 \\
Care & tolerance diversity & moderate tolerance & 4.630 & 5.000 & 4.500 & 5.000 & 0.892 \\
Care & tolerance diversity & high tolerance & 4.761 & 5.000 & 5.000 & 5.000 & 0.933 \\
Care & social trust & cannot trust people & 4.449 & 5.000 & 4.000 & 5.000 & 0.824 \\
Care & social trust & most people trusted & 4.649 & 5.000 & 4.670 & 5.000 & 0.898 \\
Equality & materialism orientation & materialist & 1.814 & 1.830 & 1.330 & 2.170 & 0.001 \\
Equality & materialism orientation & mixed & 2.444 & 2.330 & 2.170 & 3.000 & 0.010 \\
Equality & materialism orientation & postmaterialist & 2.758 & 2.830 & 2.330 & 3.170 & 0.044 \\
Proportionality & materialism orientation & materialist & 4.817 & 5.000 & 4.830 & 5.000 & 0.986 \\
Proportionality & materialism orientation & mixed & 4.514 & 4.670 & 4.000 & 5.000 & 0.906 \\
Proportionality & materialism orientation & postmaterialist & 4.194 & 4.000 & 3.830 & 4.830 & 0.711 \\
Loyalty & national pride & not proud at all & 2.767 & 2.830 & 2.500 & 3.000 & 0.013 \\
Loyalty & national pride & not very proud & 3.033 & 3.000 & 2.670 & 3.330 & 0.040 \\
Loyalty & national pride & somewhat proud & 3.825 & 3.830 & 3.500 & 4.170 & 0.468 \\
Loyalty & national pride & very proud & 4.605 & 4.670 & 4.330 & 4.830 & 0.923 \\
Authority & child rearing value & independence imagination & 3.637 & 3.670 & 3.000 & 4.330 & 0.403 \\
Authority & child rearing value & neutral & 3.751 & 3.830 & 3.170 & 4.500 & 0.442 \\
Authority & child rearing value & obedience faith & 4.488 & 4.670 & 4.170 & 4.830 & 0.850 \\
Authority & gender equality & traditional & 4.410 & 4.670 & 4.000 & 5.000 & 0.788 \\
Authority & gender equality & moderate & 3.871 & 4.000 & 3.170 & 4.670 & 0.522 \\
Authority & gender equality & egalitarian & 3.596 & 3.500 & 2.830 & 4.330 & 0.385 \\
Authority & national pride & not proud at all & 3.769 & 3.830 & 3.000 & 4.500 & 0.482 \\
Authority & national pride & not very proud & 3.827 & 4.000 & 3.170 & 4.500 & 0.508 \\
Authority & national pride & somewhat proud & 3.971 & 4.000 & 3.330 & 4.830 & 0.572 \\
Authority & national pride & very proud & 4.268 & 4.500 & 3.670 & 5.000 & 0.698 \\
Authority & religiosity & not at all important & 3.526 & 3.500 & 2.830 & 4.170 & 0.353 \\
Authority & religiosity & not very important & 3.590 & 3.500 & 3.000 & 4.330 & 0.379 \\
Authority & religiosity & quite important & 4.300 & 4.500 & 3.830 & 4.830 & 0.730 \\
Authority & religiosity & very important & 4.419 & 4.670 & 4.000 & 4.830 & 0.798 \\
Authority & tolerance diversity & low tolerance & 4.214 & 4.500 & 3.670 & 4.830 & 0.701 \\
Authority & tolerance diversity & moderate tolerance & 3.915 & 4.000 & 3.170 & 4.670 & 0.540 \\
Authority & tolerance diversity & high tolerance & 3.748 & 3.830 & 3.170 & 4.500 & 0.454 \\
Purity & religiosity & not at all important & 2.552 & 2.500 & 2.000 & 3.000 & 0.024 \\
Purity & religiosity & not very important & 2.688 & 2.670 & 2.170 & 3.170 & 0.043 \\
Purity & religiosity & quite important & 4.086 & 4.170 & 3.830 & 4.500 & 0.709 \\
Purity & religiosity & very important & 4.208 & 4.330 & 4.000 & 4.500 & 0.807 \\
\bottomrule
\end{tabular}
\caption{MFQ-2 scores by core cultural variables and admissible values. We report the mean, median, interquartile range (Q25--Q75), and the fraction of personas with MFQ-2 score $\ge 4$ (column `p($\ge$4)').}
\label{tab:mfq2-core-detailed}
\end{table*}

\clearpage

\begin{table*}[ht!]
\centering
\small
\begin{tabular}{lrrrrrr}
\toprule
Pair & Mean & Median & $P_{90}$ & $P_{95}$ & $d\leq 1$, \# (\%) & $d\leq 2$, \# (\%) \\
\midrule
\textsf{gpt-oss-20B}--\textsf{Qwen3.5-9B} & 1.415 & 1.237 & 2.687 & 3.204 & 36,375 (38.99) & 71,791 (76.94) \\
\textsf{gpt-oss-20B}--\textsf{Llama3.1-8B-Instruct} & 1.775 & 1.558 & 3.349 & 3.896 & 26,861 (28.79) & 59,512 (63.78) \\
\textsf{Qwen3.5-9B}--\textsf{Llama3.1-8B-Instruct} & 1.589 & 1.397 & 3.009 & 3.580 & 30,978 (33.20) & 66,168 (70.91) \\
\bottomrule
\end{tabular}
\caption{Pairwise Euclidean distances between model-specific IW coordinates for matched cultural conditionings. $P_{90}$ and $P_{95}$ denote the 90th and 95th percentiles of the pairwise distance distribution. The $d\leq 1$ and $d\leq 2$ columns report the number and percentage of matched personas whose euclidean distance is at most 1 or 2, respectively.}
\label{tab:iw-pairwise-distances}
\end{table*}

\begin{table*}[ht!]
\centering
\small
\begin{tabular}{lrrr}
\toprule
Metric &
\textsf{gpt-oss-20B} &
\textsf{Qwen3.5-9B} &
\textsf{Llama3.1-8B-Instruct} \\
\midrule
Cells with at least one persona
& 112 & 109 & 110 \\

Cells with itemsets
& 111 & 109 & 109 \\

Itemset--cell pairs
& 735 & 806 & 856 \\

Unique itemsets
& 138 & 156 & 165 \\

Non-singletons in $\geq 2$ cells
& 31 & 33 & 43 \\
\bottomrule
\end{tabular}
\caption{
Summary of closed-itemset mining across models.
All models are evaluated with the same $12\times12$ grid,
minimum support $0.2$, $\rho=0.5$, and a minimum of two
personas per cell.
}
\label{tab:iw-closed-itemsets-global}
\end{table*}

\begin{table*}[ht!]
\centering
\scriptsize
\setlength{\tabcolsep}{2.2pt}
\renewcommand{\arraystretch}{1.15}
\begin{tabular}{@{}p{7cm}llllllllllll@{}}
\toprule
& \multicolumn{4}{c}{\textsf{gpt-oss-20B}}
& \multicolumn{4}{c}{\textsf{Qwen3.5-9B}}
& \multicolumn{4}{c}{\textsf{Llama3.1-8B-Instruct}} \\
\cmidrule(lr){2-5}\cmidrule(lr){6-9}\cmidrule(l){10-13}
Cultural configuration
& \# & avg. & range & avg./cell
& \# & avg. & range & avg./cell
& \# & avg. & range & avg./cell \\
\midrule
\multicolumn{13}{@{}l}{\textit{GPT configurations retained by all three models and shown in Fig. \ref{fig:qwen-llama-gpt-itemsets}}} \\
\addlinespace[2pt]
\makecell[l]{happy = very happy; trust = most people trusted}
& 21 & .735 & [.47,.99] & 353
& 17 & .752 & [.54,1.00] & 287
& 18 & .781 & [.56,.94] & 412 \\
\makecell[l]{happy = not at all happy; trust = cannot trust people}
& 15 & .806 & [.51,1.00] & 370
& 16 & .645 & [.46,.82] & 352
& 23 & .631 & [.49,.90] & 344 \\
\makecell[l]{happy = very happy; pride = very proud; trust = most people trusted}
& 11 & .796 & [.55,1.00] & 131
& 7  & .749 & [.68,.83] & 125
& 10 & .769 & [.60,.90] & 206 \\
\makecell[l]{child rearing = obedience--faith; trust = cannot trust people}
& 6  & .670 & [.47,.84] & 577
& 7  & .630 & [.57,.68] & 493
& 14 & .658 & [.53,.81] & 429 \\
\addlinespace[3pt]
\multicolumn{13}{@{}l}{\textit{Additional configurations retained by all three models}} \\
\addlinespace[2pt]
\makecell[l]{pride = very proud; trust = most people trusted}
& 14 & .790 & [.51,1.00] & 232
& 13 & .796 & [.56,1.00] & 241
& 11 & .831 & [.57,1.00] & 229 \\
\makecell[l]{happy = very happy; pride = very proud}
& 9  & .769 & [.55,.91] & 218
& 8  & .716 & [.56,.79] & 388
& 10 & .809 & [.63,.90] & 309 \\
\makecell[l]{political participation = active; trust = most people trusted}
& 2  & .717 & [.63,.81] & 107
& 11 & .713 & [.57,.87] & 175
& 5  & .691 & [.64,.77] & 129 \\
\addlinespace[3pt]
\multicolumn{13}{@{}l}{\textit{Configurations shared by \textsf{Qwen3.5-9B} and \textsf{Llama3.1-8B-Instruct} but not retained for gpt-oss}} \\
\addlinespace[2pt]
\makecell[l]{trust = cannot trust people; tolerance diversity = low}
& -- & -- & -- & --
& 5  & .690 & [.58,.85] & 182
& 14 & .660 & [.56,.88] & 118 \\
\makecell[l]{religiosity = not at all important; trust = most people trusted}
& -- & -- & -- & --
& 2  & .876 & [.83,.93] & 25
& 7  & .767 & [.72,.83] & 89 \\
\addlinespace[3pt]
\multicolumn{13}{@{}l}{\textit{Model-specific retained configurations}} \\
\addlinespace[2pt]
\makecell[l]{gender equality = egalitarian; trust = most people trusted}
& -- & -- & -- & --
& 3  & .661 & [.64,.68] & 92
& -- & -- & -- & -- \\
\makecell[l]{gender equality = traditional;trust = cannot trust people}
& -- & -- & -- & --
& -- & -- & -- & --
& 5  & .685 & [.58,.79] & 59 \\
\bottomrule
\end{tabular}
\caption{Selected non-singleton closed itemsets across the three models. For each model, \# is the number of grid cells in which the itemset is retained, avg. is its mean within-cell support, range is the minimum--maximum support interval, and avg./cell is the mean number of personas in the covered cells.  
}
\label{tab:iw-selected-itemsets-cross-model}
\end{table*}

\begin{table*}[ht!]
\centering
\small
\begin{tabular}{llrrrrrr}
\toprule
Model &
Set &
\# &
Mean cells &
Med.\ cells &
$\leq 3$ cells (\%) &
Med.\ pers./cell &
Med.\ support \\
\midrule
\textsf{gpt-oss-20B} & Shared by all & 11 & 8.27 & 6.0 & 36.4 & 218.1 & 0.717 \\
\textsf{gpt-oss-20B} & Model-specific & 10 & 3.20 & 2.0 & 80.0 & 136.4 & 0.677 \\
\textsf{Qwen3.5-9B} & Shared by all & 11 & 8.27 & 7.0 & 27.3 & 240.8 & 0.716 \\
\textsf{Qwen3.5-9B} & Model-specific & 7 & 2.29 & 2.0 & 100.0 & 74.5 & 0.653 \\
\textsf{Llama3.1-8B-Instruct} & Shared by all & 11 & 9.36 & 10.0 & 27.3 & 229.4 & 0.723 \\
\textsf{Llama3.1-8B-Instruct} & Model-specific & 21 & 2.48 & 2.0 & 90.5 & 69.5 & 0.641 \\
\bottomrule
\end{tabular}
\caption{
Comparison between non-singleton closed frequent itemsets retained in at least two cells and either shared by all three models or retained by only one model. \emph{Set} distinguishes shared and model-specific itemsets, while \emph{\#} reports the number of itemsets in each set. \emph{Mean cells} and \emph{Med.\ cells} report the mean and median number of grid cells in which an itemset is retained. Since all itemsets cover at least two cells, $\leq 3$ cells reports the percentage retained in either two or three cells. \emph{Med.\ pers./cell} is the median, across itemsets, of the average population of their covered cells and does not indicate the number of personas satisfying the itemset. \emph{Med.\ support} is the median, across itemsets, of their mean within-cell support.
}
\label{tab:iw-shared-vs-model-specific-itemsets}
\end{table*}

\begin{table*}
\centering
\begin{tabular}{lrrr}
\toprule
Aggregation & $1-\overline{EMD}$ & $\%\,\mathit{EMD}<0.4$ & $\%\,\mathit{EMD}<0.2$ \\
\midrule
Unweighted & 0.780 & 90.06 & 51.91 \\
Weighted & 0.795 & 95.17 & 54.68 \\
\bottomrule
\end{tabular}
\caption{WVB-Probe alignment for \textsf{Qwen3.5-9B}, aggregated by demographic groups.  Higher $1-\overline{EMD}$ values indicate closer alignment with the WVS reference distributions.}
\label{tab:qwen3-5-9b-high-run1-wvs-overall}
\end{table*}

\begin{table*}[h]
    \centering
    \scalebox{0.88}{
    \begin{tabular}{lrrrr}
\toprule
\textbf{Group (Continent \textbar Settlement \textbar Education)} & \# \textbf{personas} & \textbf{$1-\overline{EMD}$} & \textbf{\%($\mathit{EMD}<$0.4)} & \textbf{\%($\mathit{EMD}<$0.2)} \\
\midrule
Europe | Urban | Tertiary & 31887 & 0.799 & 100.0 & 55.6 \\
Europe | Urban | Upper to Post Secondary & 16690 & 0.780 & 88.9 & 47.2 \\
Europe | Rural | Upper to Post Secondary & 8908 & 0.799 & 97.2 & 55.6 \\
North America | Rural | Upper to Post Secondary & 7293 & 0.779 & 88.9 & 47.2 \\
North America | Urban | Tertiary & 7293 & 0.800 & 94.4 & 63.9 \\
North America | Urban | Upper to Post Secondary & 5766 & 0.811 & 94.4 & 61.1 \\
Europe | Rural | Tertiary & 4023 & 0.809 & 97.2 & 58.3 \\
North America | Rural | Tertiary & 1529 & 0.812 & 94.4 & 66.7 \\
South America | Urban | Tertiary & 1465 & 0.803 & 94.4 & 55.6 \\
South America | Urban | Upper to Post Secondary & 1380 & 0.800 & 88.9 & 63.9 \\
Asia | Urban | Upper to Post Secondary & 1227 & 0.815 & 100.0 & 58.3 \\
Asia | Urban | Tertiary & 1077 & 0.808 & 100.0 & 66.7 \\
Asia | Rural | Upper to Post Secondary & 478 & 0.790 & 94.4 & 52.8 \\
Europe | Rural | Primary or No Education & 454 & 0.778 & 86.1 & 47.2 \\
Europe | Urban | Lower Secondary & 371 & 0.759 & 94.4 & 25.0 \\
Europe | Rural | Lower Secondary & 347 & 0.794 & 94.4 & 61.1 \\
North America | Rural | Primary or No Education & 331 & 0.756 & 86.1 & 44.4 \\
Africa | Urban | Tertiary & 330 & 0.755 & 86.1 & 38.9 \\
Europe | Urban | Primary or No Education & 316 & 0.791 & 100.0 & 47.2 \\
South America | Rural | Upper to Post Secondary & 309 & 0.805 & 86.1 & 58.3 \\
Africa | Urban | Upper to Post Secondary & 276 & 0.816 & 97.2 & 61.1 \\
Oceania | Urban | Tertiary & 255 & 0.798 & 94.4 & 52.8 \\
North America | Rural | Lower Secondary & 183 & 0.751 & 80.6 & 47.2 \\
North America | Urban | Primary or No Education & 176 & 0.778 & 88.9 & 55.6 \\
North America | Urban | Lower Secondary & 108 & 0.749 & 88.9 & 38.9 \\
Asia | Rural | Primary or No Education & 93 & 0.772 & 86.1 & 52.8 \\
Africa | Rural | Upper to Post Secondary & 84 & 0.821 & 94.4 & 66.7 \\
Oceania | Urban | Upper to Post Secondary & 75 & 0.807 & 100.0 & 52.8 \\
South America | Rural | Tertiary & 71 & 0.771 & 83.3 & 58.3 \\
Asia | Rural | Tertiary & 66 & 0.814 & 100.0 & 55.6 \\
Asia | Urban | Primary or No Education & 62 & 0.772 & 86.1 & 52.8 \\
South America | Rural | Primary or No Education & 52 & 0.755 & 80.6 & 41.7 \\
South America | Urban | Primary or No Education & 52 & 0.823 & 100.0 & 66.7 \\
Asia | Rural | Lower Secondary & 44 & 0.755 & 83.3 & 44.4 \\
South America | Urban | Lower Secondary & 43 & 0.791 & 91.7 & 61.1 \\
South America | Rural | Lower Secondary & 40 & 0.772 & 86.1 & 58.3 \\
Oceania | Rural | Tertiary & 34 & 0.795 & 94.4 & 52.8 \\
Oceania | Rural | Upper to Post Secondary & 32 & 0.794 & 97.2 & 52.8 \\
Africa | Rural | Primary or No Education & 23 & 0.741 & 86.1 & 30.6 \\
Asia | Urban | Lower Secondary & 23 & 0.769 & 88.9 & 38.9 \\
Africa | Urban | Primary or No Education & 17 & 0.780 & 86.1 & 52.8 \\
Africa | Rural | Tertiary & 13 & 0.761 & 88.9 & 38.9 \\
Africa | Urban | Lower Secondary & 7 & 0.716 & 72.2 & 44.4 \\
Africa | Rural | Lower Secondary & 6 & 0.773 & 91.7 & 58.3 \\
Oceania | Urban | Primary or No Education & 1 & 0.582 & 38.9 & 25.0 \\
\bottomrule
\end{tabular}
}
    \caption{Complete group-level WVB-Probe alignment for the 45 demographic triples represented in the \textsf{Qwen3.5-9B} outputs and associated with an available WVB-Probe reference distribution, sorted by decreasing number of personas. The score is $1-\overline{EMD}$, where $\overline{EMD}$ is the mean EMD over the 36 probe questions for the group. $\%\,EMD<0.4$ and $\%\,EMD<0.2$ denote the percentage of probe questions whose distance falls below the corresponding threshold.}
    \label{tab:qwen-wvb-probe-all-groups}
\end{table*}

\begin{table*}
\centering
\begin{tabular}{lrrr}
\toprule
Aggregation & $1-\overline{EMD}$ & $\%\,\mathit{EMD}<0.4$ & $\%\,\mathit{EMD}<0.2$ \\
\midrule
Unweighted & 0.782 & 89.26 & 52.28 \\
Weighted & 0.797 & 93.62 & 56.22 \\
\bottomrule
\end{tabular}
\caption{WVB-Probe alignment for \textsf{Llama3.1-8B-Instruct}, aggregated by demographic groups.  Higher $1-\overline{EMD}$ values indicate closer alignment with the WVS reference distributions.}
\label{tab:llama3-1-8b-instruct-high-run1-wvs-overall}
\end{table*}

\begin{table*}[h]
    \centering
    \scalebox{0.88}{
    \begin{tabular}{lrrrr}
\toprule
\textbf{Group (Continent \textbar Settlement \textbar Education)} & \# \textbf{personas} & \textbf{$1-\overline{EMD}$} & \textbf{\%($\mathit{EMD}<$0.4)} & \textbf{\%($\mathit{EMD}<$0.2)} \\
\midrule

North America | Urban | Tertiary & 20396 & 0.814 & 94.4 & 63.9 \\
Europe | Urban | Tertiary & 18753 & 0.797 & 97.2 & 52.8 \\
North America | Rural | Upper to Post Secondary & 16603 & 0.785 & 91.7 & 55.6 \\
North America | Rural | Tertiary & 5868 & 0.815 & 97.2 & 63.9 \\
Europe | Rural | Upper to Post Secondary & 5303 & 0.792 & 97.2 & 47.2 \\
Europe | Urban | Upper to Post Secondary & 3760 & 0.784 & 88.9 & 52.8 \\
North America | Urban | Upper to Post Secondary & 3667 & 0.767 & 86.1 & 44.4 \\
South America | Urban | Tertiary & 3394 & 0.772 & 91.7 & 52.8 \\
Europe | Rural | Tertiary & 2812 & 0.821 & 94.4 & 61.1 \\
Europe | Rural | Primary or No Education & 1501 & 0.779 & 88.9 & 52.8 \\
Asia | Urban | Tertiary & 1272 & 0.827 & 97.2 & 61.1 \\
North America | Rural | Primary or No Education & 1217 & 0.772 & 86.1 & 50.0 \\
South America | Urban | Upper to Post Secondary & 1214 & 0.771 & 88.9 & 41.7 \\
Oceania | Urban | Tertiary & 1140 & 0.811 & 94.4 & 63.9 \\
North America | Rural | Lower Secondary & 1109 & 0.776 & 86.1 & 47.2 \\
South America | Rural | Upper to Post Secondary & 938 & 0.800 & 88.9 & 52.8 \\
Europe | Rural | Lower Secondary & 883 & 0.833 & 97.2 & 72.2 \\
South America | Rural | Primary or No Education & 726 & 0.771 & 83.3 & 50.0 \\
Asia | Rural | Upper to Post Secondary & 369 & 0.798 & 94.4 & 58.3 \\
Asia | Urban | Upper to Post Secondary & 319 & 0.813 & 94.4 & 61.1 \\
South America | Rural | Lower Secondary & 284 & 0.785 & 80.6 & 55.6 \\
South America | Rural | Tertiary & 264 & 0.747 & 77.8 & 44.4 \\
Europe | Urban | Lower Secondary & 220 & 0.793 & 91.7 & 58.3 \\
Oceania | Rural | Tertiary & 177 & 0.817 & 100.0 & 58.3 \\
Europe | Urban | Primary or No Education & 158 & 0.817 & 100.0 & 52.8 \\
North America | Urban | Lower Secondary & 138 & 0.749 & 88.9 & 44.4 \\
South America | Urban | Lower Secondary & 126 & 0.793 & 94.4 & 50.0 \\
Asia | Rural | Tertiary & 112 & 0.768 & 86.1 & 50.0 \\
South America | Urban | Primary or No Education & 99 & 0.798 & 94.4 & 55.6 \\
Asia | Rural | Primary or No Education & 91 & 0.801 & 88.9 & 61.1 \\
Oceania | Rural | Upper to Post Secondary & 87 & 0.782 & 91.7 & 52.8 \\
Asia | Rural | Lower Secondary & 78 & 0.772 & 86.1 & 52.8 \\
Africa | Urban | Tertiary & 52 & 0.769 & 80.6 & 50.0 \\
North America | Urban | Primary or No Education & 47 & 0.745 & 88.9 & 41.7 \\
Africa | Urban | Upper to Post Secondary & 40 & 0.787 & 91.7 & 58.3 \\
Oceania | Urban | Upper to Post Secondary & 17 & 0.814 & 91.7 & 55.6 \\
Asia | Urban | Lower Secondary & 10 & 0.814 & 88.9 & 66.7 \\
Asia | Urban | Primary or No Education & 7 & 0.771 & 88.9 & 50.0 \\
Africa | Rural | Primary or No Education & 5 & 0.737 & 80.6 & 33.3 \\
Africa | Rural | Upper to Post Secondary & 5 & 0.753 & 80.6 & 38.9 \\
Africa | Urban | Lower Secondary & 3 & 0.766 & 91.7 & 44.4 \\
Africa | Rural | Lower Secondary & 2 & 0.759 & 88.9 & 44.4 \\
Oceania | Urban | Primary or No Education & 2 & 0.761 & 77.8 & 50.0 \\
Africa | Rural | Tertiary & 1 & 0.703 & 72.2 & 36.1 \\
Oceania | Urban | Lower Secondary & 1 & 0.694 & 75.0 & 41.7 \\
\bottomrule
\end{tabular}
}
    \caption{Complete group-level WVB-Probe alignment for the 45 demographic triples represented in the  \textsf{Llama3.1-8B-Instruct} outputs and associated with an available WVB-Probe reference distribution, sorted by decreasing number of personas. The score is $1-\overline{EMD}$, where $\overline{EMD}$ is the mean EMD over the 36 probe questions for the group. $\%\,EMD<0.4$ and $\%\,EMD<0.2$ denote the percentage of probe questions whose distance falls below the corresponding threshold.}
    \label{tab:llama-wvb-probe-all-groups}
\end{table*}

\begin{figure*}[ht!]
    \centering

    \begin{subfigure}{0.3\textwidth}
        \centering
        \includegraphics[width=0.85\linewidth]{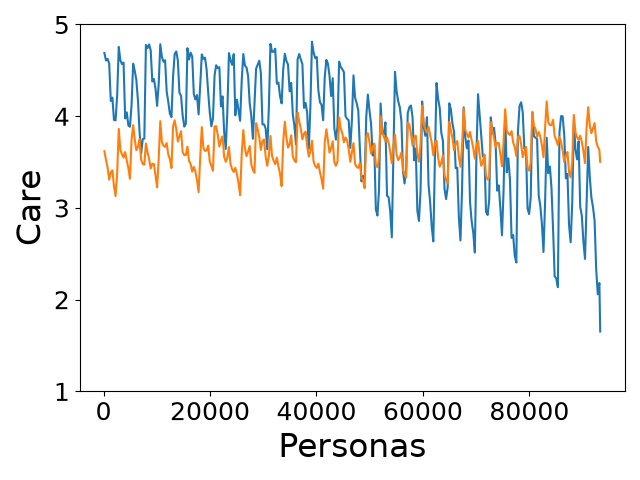}
        \caption{}
    \end{subfigure}\hfill
    \begin{subfigure}{0.3\textwidth}
        \centering
        \includegraphics[width=0.85\linewidth]{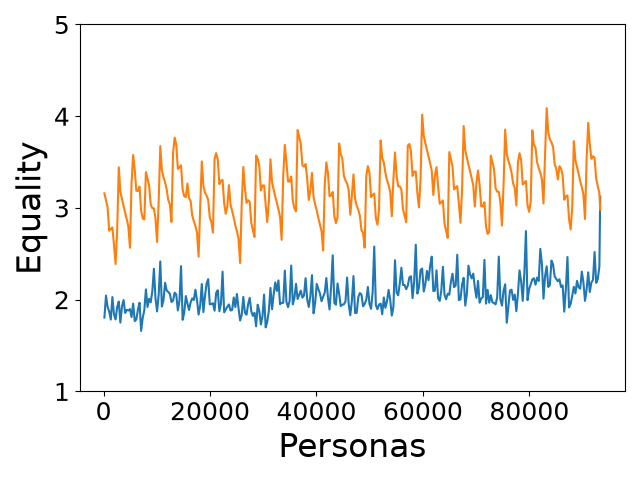}
        \caption{}
    \end{subfigure}\hfill
    \begin{subfigure}{0.3\textwidth}
        \centering
        \includegraphics[width=0.85\linewidth]{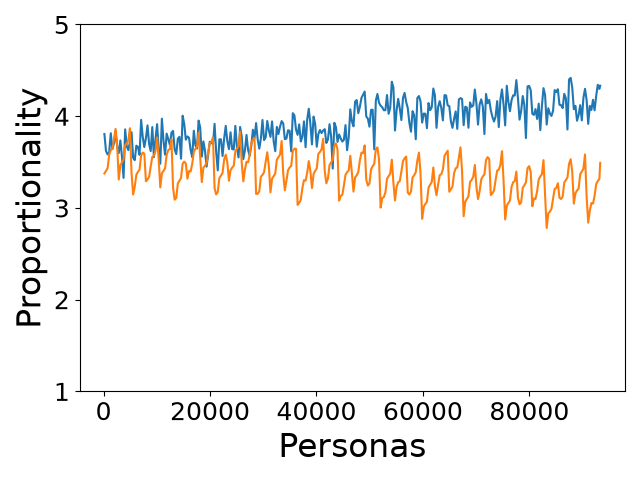}
        \caption{}
    \end{subfigure}


    \begin{subfigure}{0.3\textwidth}
        \centering
        \includegraphics[width=0.85\linewidth]{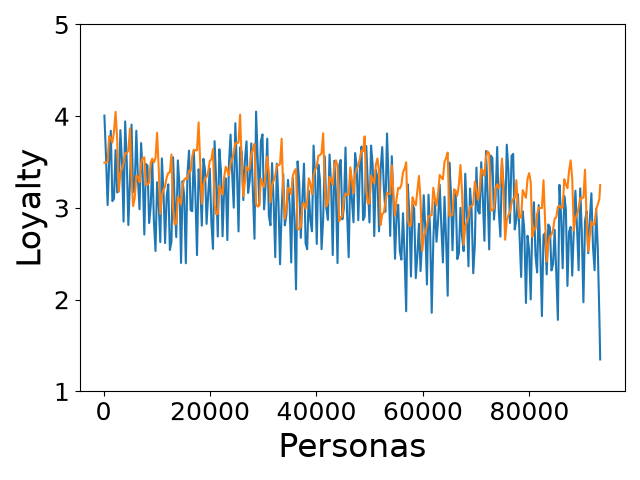}
        \caption{}
    \end{subfigure}\hfill
    \begin{subfigure}{0.3\textwidth}
        \centering
        \includegraphics[width=0.85\linewidth]{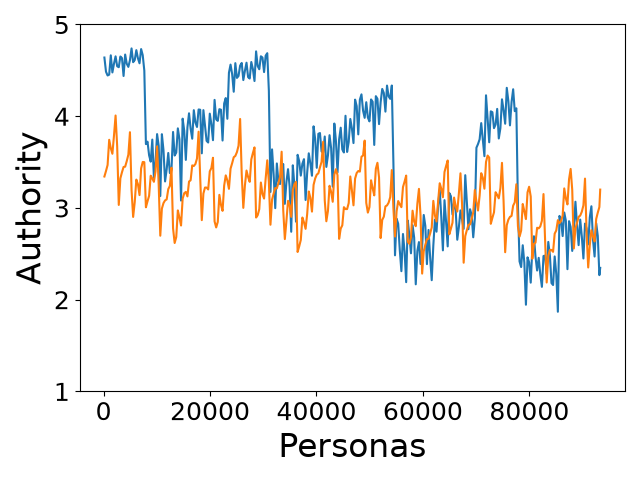}
        \caption{}
    \end{subfigure}\hfill
    \begin{subfigure}{0.3\textwidth}
        \centering
        \includegraphics[width=0.85\linewidth]{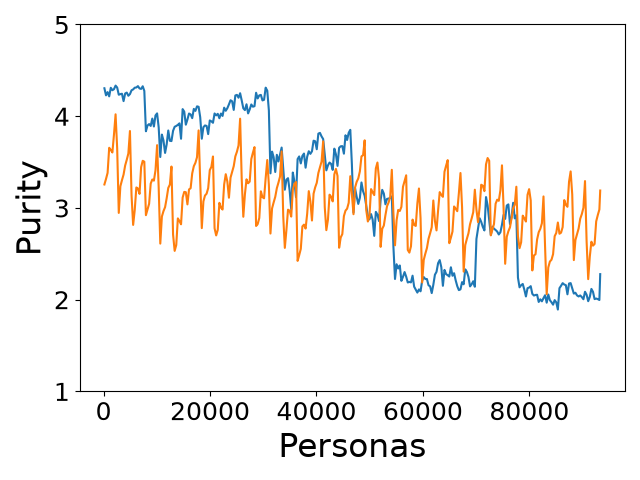}
        \caption{}
    \end{subfigure}

    \caption{\textsf{Qwen3.5-9B} Comparison between MFQ-2–based (blue) and mapping-based moral (orange) foundation scores. Mapping-based scores are obtained using all cultural variables. Personas are ordered along the x-axis; the y-axis reports foundation scores. To avoid cluttering, lines of both score types  are smoothed by averaging scores within consecutive bins of 300 personas (according to the ordering fixed through sorting  by cultural variable then by admissible values).}
    \label{fig:QWEN-mfq2-all10}
\end{figure*}

\begin{table*}[t]
\centering
\small
\setlength{\tabcolsep}{2pt}
\begin{tabular}{lcp{4cm}p{6cm}}
\toprule
\textbf{Moral Foundation} & $\bar J$ & \textbf{$k$ distribution (\#runs)} & \textbf{Selected variables (support)} \\
\midrule

Care & 1.000 &
$k{=}1$ (50) &
religiosity (1.00)\\

Equality & 1.000 &
$k{=}3$ (50) &
materialism\_orientation (1.00), gender\_equality (1.00), national\_pride (1.00) \\

Proportionality & 1.000 &
$k{=}2$ (50) &
materialism\_orientation (1.00), national\_pride (1.00) \\

Loyalty & 1.000 &
$k{=}1$ (50) &
national\_pride (1.00) \\

Authority & 1.000 &
$k{=}5$ (50) &
child\_rearing\_value (1.00), gender\_equality (1.00), national\_pride (1.00), religiosity (1.00), tolerance\_diversity (1.00) \\

Purity & 0.877 &
$k{=}1$ (43), $k{=}2$ (7) &
religiosity (1.00), child\_rearing\_value (0.14) \\
\bottomrule
\end{tabular}
\caption{Stability of core-variable selection across 50 independent runs for \textsf{Qwen3.5-9B}.
$\bar J$ denotes the mean pairwise Jaccard overlap between selected sets.
The $k$ distribution is reported as $k{=}m$ (count), where $m$ is the number of selected variables and the value in parentheses is the number of runs  in which the $m$ variables were selected.}
\label{tab:QWEN_selection_stability}
\end{table*}

\begin{figure*}[ht!]
    \centering

    \begin{subfigure}{0.3\textwidth}
        \centering
        \includegraphics[width=0.85\linewidth]{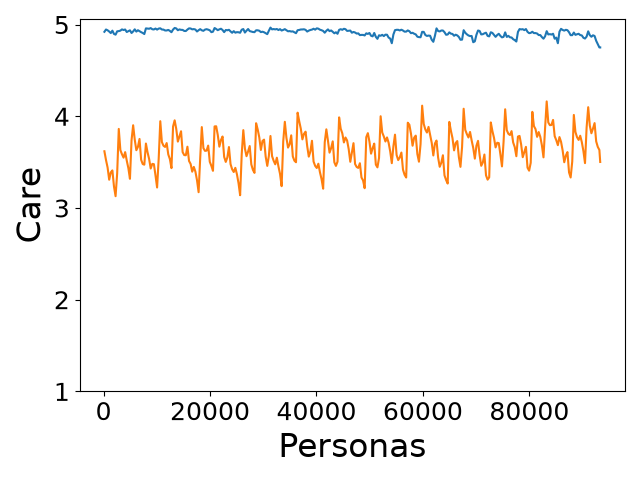}
        \caption{}
    \end{subfigure}\hfill
    \begin{subfigure}{0.3\textwidth}
        \centering
        \includegraphics[width=0.85\linewidth]{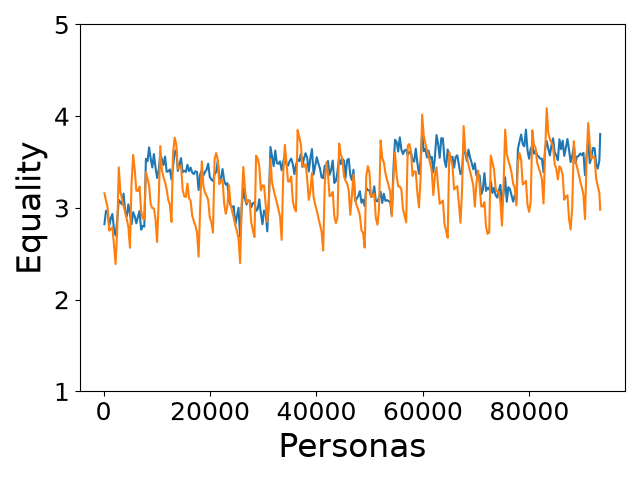}
        \caption{}
    \end{subfigure}\hfill
    \begin{subfigure}{0.3\textwidth}
        \centering
        \includegraphics[width=0.85\linewidth]{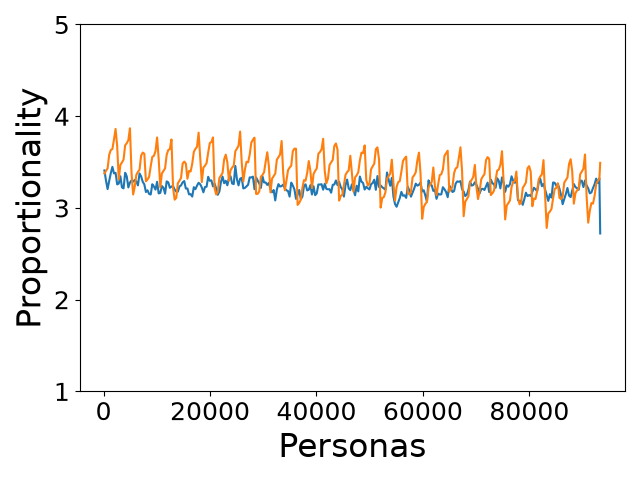}
        \caption{}
    \end{subfigure}


    \begin{subfigure}{0.3\textwidth}
        \centering
        \includegraphics[width=0.85\linewidth]{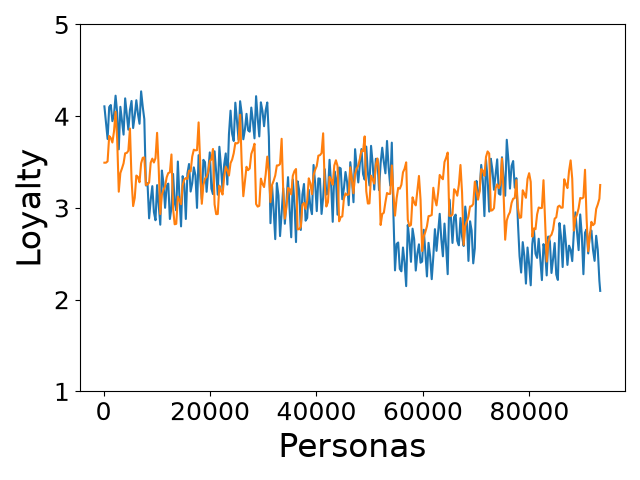}
        \caption{}
    \end{subfigure}\hfill
    \begin{subfigure}{0.3\textwidth}
        \centering
        \includegraphics[width=0.85\linewidth]{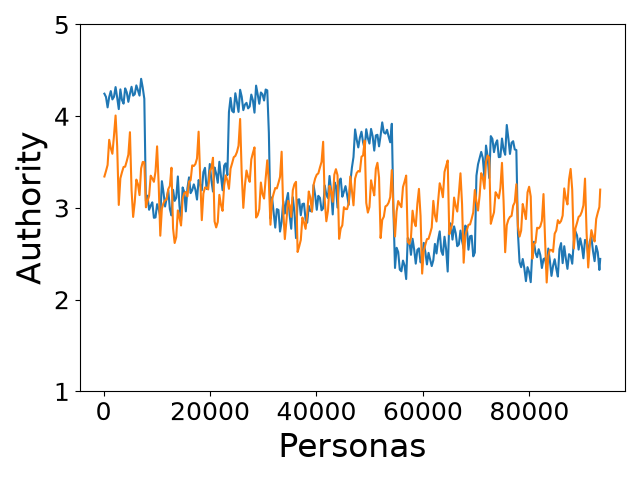}
        \caption{}
    \end{subfigure}\hfill
    \begin{subfigure}{0.3\textwidth}
        \centering
        \includegraphics[width=0.85\linewidth]{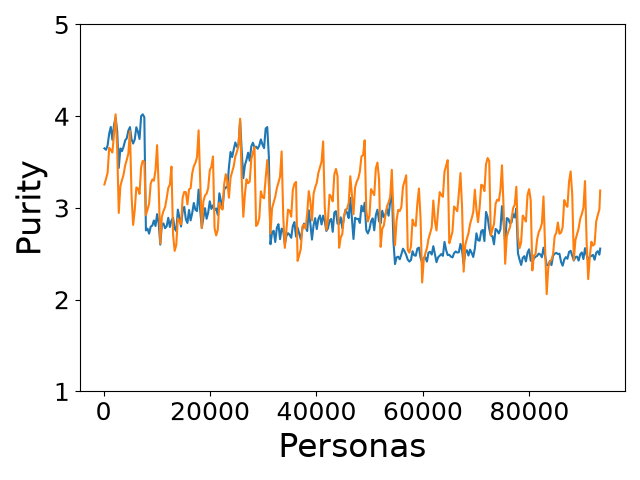}
        \caption{}
    \end{subfigure}

    \caption{ Comparison between MFQ-2–based (blue) and mapping-based moral (orange) foundation scores for \textsf{Llama3.1-8B-Instruct}. Mapping-based scores are obtained using all cultural variables. Personas are ordered along the x-axis; the y-axis reports foundation scores. To avoid cluttering, lines of both score types  are smoothed by averaging scores within consecutive bins of 300 personas (according to the ordering fixed through sorting  by cultural variable then by admissible values).}
    \label{fig:LLAMA-mfq2-all10}
\end{figure*}

\begin{table*}[t]
\centering
\small
\setlength{\tabcolsep}{2pt}
\begin{tabular}{lcp{3cm}p{9cm}}
\toprule
\textbf{Moral Foundation} & $\bar J$ & \textbf{$k$ distribution (\#runs)} & \textbf{Selected variables (support)} \\
\midrule

Care & 0.917 &
$k{=}6$ (12), $k{=}7$ (36), $k{=}8$ (2) &
child\_rearing\_value (1.00), gender\_equality (1.00), happiness (1.00), materialism\_orientation (1.00), social\_trust (1.00), tolerance\_diversity (1.00), national\_pride (0.70), political\_participation (0.10) \\

Equality & 1.000 &
$k{=}6$ (50) &
child\_rearing\_value (1.00), gender\_equality (1.00), materialism\_orientation (1.00), national\_pride (1.00), religiosity (1.00), tolerance\_diversity (1.00)\\

Proportionality & 1.000 &
$k{=}1$ (50) &
materialism\_orientation (1.00) \\

Loyalty & 0.950 &
$k{=}4$ (4), $k{=}6$ (46) &
child\_rearing\_value (1.00), gender\_equality (1.00), national\_pride (1.00), religiosity (1.00), materialism\_orientation (0.92), tolerance\_diversity (0.92) \\

Authority & 0.950 &
$k{=}4$ (46), $k{=}6$ (4) &
child\_rearing\_value (1.00), gender\_equality (1.00), national\_pride (1.00), religiosity (1.00), materialism\_orientation (0.08), tolerance\_diversity (0.08) \\

Purity & 0.922 &
$k{=}5$ (32), $k{=}6$ (18) &
child\_rearing\_value (1.00), gender\_equality (1.00), national\_pride (1.00), religiosity (1.00), tolerance\_diversity (1.00), materialism\_orientation (0.36) \\
\bottomrule
\end{tabular}
\caption{ Stability of core-variable selection across 50 independent runs for \textsf{Llama3.1-8B-Instruct}.
$\bar J$ denotes the mean pairwise Jaccard overlap between selected sets.
The $k$ distribution is reported as $k{=}m$ (count), where $m$ is the number of selected variables and the value in parentheses is the number of runs  in which the $m$ variables were selected.}
\label{tab:LLAMA_selection_stability}
\end{table*}

\end{document}